\documentclass{article}

\usepackage[accepted]{icml2024}

\usepackage{hyperref}

\usepackage{microtype}

\usepackage{algorithm}
\usepackage{algorithmic}
\usepackage{footnote}

\newcommand{\theHalgorithm}{\arabic{algorithm}}

\usepackage{amsthm}

\usepackage{enumitem}
\usepackage{float}

\usepackage{verbatim}
\usepackage{multirow}

\usepackage{xr}
\usepackage{balance}
\usepackage{xspace}

\usepackage{url}
\usepackage{amsmath}
\usepackage{amssymb}

\usepackage{mathtools}

\usepackage{dsfont}
\usepackage{upgreek}
\usepackage{bbm}			

\usepackage{graphicx}
\usepackage{epstopdf}
\graphicspath{ {../../figures/} } 
\usepackage{grffile}
\usepackage{tikz}
\usepackage{float}
\usepackage{svg}

\usepackage{booktabs} 

\usepackage[labelfont=bf]{caption}
\usepackage{subfigure}

\usepackage{mathtools}
\usepackage{relsize}  

\usepackage{makecell}

\usepackage{expl3}

\newcommand{\Sec}[1]		{Sec.\,\ref{#1}}

\newcommand{\Appendix}[1]		{Appendix\,\ref{#1}}		
\newcommand{\Fig}[1]		{Fig.\,\ref{#1}}

\newcommand{\Eq}[1]			{Eq.\,\ref{#1}}
\newcommand{\Expr}[1]			{Expr.\,\ref{#1}}
\newcommand{\Ineq}[1]			{Ineq.\,\ref{#1}}
\newcommand{\Tab}[1]		{Tab.\,\ref{#1}}
\newcommand{\Alg}[1]		{Alg.\,\ref{#1}}
\newcommand{\Theorem}[1]{Theorem\,\ref{#1}}

\newcommand{\Problem}[1]{Problem\,\ref{#1}}	
\newcommand{\vs}   			{vs.\@\xspace}
\newcommand{\ie}   			{i.e.\@\xspace}
\newcommand{\eg}   			{e.g.\@\xspace}
\newcommand{\etc}   		{etc.\xspace}
\newcommand{\wrt}   		{w.r.t.\@\xspace}
\newcommand{\iid}   		{iid\@\xspace}

\newcommand{\one}       {\mathbf{1}}     
\newcommand{\ones}[1]   {\one_{#1}}
\newcommand{\zero}      {\mathbf{0}}
\newcommand{\zeros}[1]   {\zero_{#1}}
\newcommand{\bigO} 		  {\mathcal{O}} 
\newcommand{\ind}       {\mathds{1}}
\newcommand{\Ind}[1]    { \ind{\{#1\}} }

\newcommand{\Exp}    {\mathbb{E}}

\newcommand{\spankern}    {\operatorname{span}}
\newcommand{\Vect}     {\operatorname{vec}}

\newcommand{\real}      {\mathbb{R}}

\newcommand{\Prob}      {\mathbb{P}}

\newcommand{\pdfunc}		{pdf\xspace}
\newcommand{\pdfuncs}		{pdfs\xspace}
\newcommand{\pdf}		    {pdf\xspace}
\newcommand{\pdfs}	 	  {pdfs\xspace}

\newcommand{\Hilbert}      {\mathbb{H}}

\newcommand{\R}      {\mathbb{R}}

\newcommand{\PE}{P\!\!E}  

\newcommand{\KernelFunc} {\textup{K}} 

\newcommand{\fdiv} 	{$\phi$-divergence\xspace}
\newcommand{\fdivs} {$\phi$-divergences\xspace}
\newcommand{\KLdiv} {KL-divergence\xspace}
\newcommand{\PEdiv} {$\chi^2$-divergence\xspace}

\newcommand{\LRE}	  {LRE\xspace}

\newcommand{\RKHS}	  {RKHS\xspace}

\newcommand{\inlinetitle}[2]  {
\noindent\textbf{\emph{#1}{#2}}}

\renewcommand*{\top}{{\mkern-1.5mu\mathsf{T}}}

\newcommand{\Expec}[1]{\mathbb{E}[#1]}

\newcommand{\q}{q}

\newcommand{\ExpecNm}[2]{\mathbb{E}_{p_{#2}(x)}[#1]}

\newcommand{\ExpecAm}[2]{\mathbb{E}_{\q_{#2}(x')}[#1]}

\newcommand{\Hnull}{\operatorname{H}_{\operatorname{null}}} 
\newcommand{\Halt}{\operatorname{H}_{\operatorname{alt}}}
\newcommand{\Hnullv}{\operatorname{H}_{\operatorname{null},v}} 
\newcommand{\Haltv}{\operatorname{H}_{\operatorname{alt},v}} 
\newcommand{\Inull}{\mathbf{I}_{\operatorname{0}}}
\newcommand{\norm}[1]{\left\lVert#1\right\rVert}

\newcommand{\dott}[2]{\langle #1,\,#2 \rangle}
\newcommand{\dotH}[2]{\langle #1,#2 \rangle_{\Hilbert}}

\newcommand{\Gramm}{\mathlarger{\mathcal{K}}}

\newcommand{\abs}[1]{\left|#1\right|}
\newcommand{\Npre}{n}
\newcommand{\Npost}{n'}

\newcommand{\setprev}{X_v}
\newcommand{\setpostv}{{X'_v}}

\newcommand{\X}{{\mathcal{X}}}
\newcommand{\Gdims}{N} 
\newcommand{\G}{G} 

\newcommand{\ThetaG}{\mathbf{\Theta}}

\newcommand{\mydef}{=} 
\newcommand{\id}{I}

\newcommand{\likelihood}{likelihood\xspace}

\newcommand{\FP}{F\!P\xspace}
\newcommand{\TP}{T\!P\xspace}

\newcommand{\FWER}{F\!W\!E\!R\xspace}

\newcommand{\Gs}{G_{\text{S}}}
\newcommand{\Gst}{G_{\text{S\!$\times\!$T}}}
\newcommand{\Cst}{C_{\text{S\!$\times\!$T}}}

\newcommand{\pvalue}{$\pi$-value\xspace}
\newcommand{\pvalues}{$\pi$-values\xspace}

\newcommand{\padded}[1] {\,#1\,}
\newcommand{\void}[1] {\padded{\cdot}}

\usepackage{booktabs,arydshln}
\makeatletter
\def\adl@drawiv#1#2#3{%
        \hskip.5\tabcolsep
        \xleaders#3{#2.5\@tempdimb #1{1}#2.5\@tempdimb}%
                #2\z@ plus1fil minus1fil\relax
        \hskip.5\tabcolsep}
\newcommand{\cmidruledashed}[1]{%
  \noalign{\vskip\aboverulesep
           \global\let\@dashdrawstore\adl@draw
           \global\let\adl@draw\adl@drawiv}
  \cdashline{#1}%
	\noalign{\vskip-\belowrulesep}
	\noalign{\vskip-\belowrulesep}
}
\makeatother

\makeatletter
\newcommand{\pushright}[1]{\ifmeasuring@#1\else\omit\hfill$\displaystyle#1$\fi\ignorespaces}
\newcommand{\pushleft}[1]{\ifmeasuring@#1\else\omit$\displaystyle#1$\hfill\fi\ignorespaces}
\makeatother

\makeatletter
\let\KV@Gin@trim@old\KV@Gin@trim
\define@key{Gin}{trim}{%
  \begingroup
  \edef\x{\endgroup
    \noexpand\setkeys{Gin}{trim@old={#1}}%
  }\x
}
\makeatother
\makeatletter
\let\KV@Gin@viewport@old\KV@Gin@viewport
\define@key{Gin}{viewport}{%
  \begingroup
  \edef\x{\endgroup
    \noexpand\setkeys{Gin}{viewport@old={#1}}%
  }\x
}
\makeatother

\DeclareMathOperator*{\argmax}{\arg\!\max}
\DeclareMathOperator*{\argmin}{\arg\!\min}

\newcommand{\CTST} {CTST\xspace}

\newcommand{\NEW}[1] 				{{\color{blue} #1}}    
\newcommand{\REF}[1] 				{{\color{brown} #1}}   
\usepackage[normalem]{ulem} 
\newcommand{\RMV}[1] 	  {{\color{orange} \sout{#1}}} 
 
\newcommand{\NOTE}[1] 				{{\color{red} #1}}
\newcommand{\VOID}[1] 				{}

\newcommand{\mySqBullet}		{\raisebox{0.25em}{{\scriptsize$_\blacksquare$}}}

\newcounter{marginNoteCounter}
\newcommand{\mN}[1]		{\stepcounter{marginNoteCounter}\,{\footnotesize $^{\text{\color{red}\textbf\themarginNoteCounter}}$}\marginpar{\footnotesize $^\text{\themarginNoteCounter}$\,\NOTE{#1}}}

\usepackage[capitalize,noabbrev]{cleveref}

\theoremstyle{plain}
\newtheorem{theorem}{Theorem}[section]

\theoremstyle{definition}

\theoremstyle{remark}

\icmltitlerunning{Collaborative non-parametric two-sample test over graphs}

\begin{document}

\twocolumn[
\icmltitle{Collaborative non-parametric two-sample testing}




\begin{icmlauthorlist}
\icmlauthor{Alejandro de la Concha}{cb}
\icmlauthor{Nicolas Vayatis}{cb}
\icmlauthor{Argyris Kalogeratos}{cb}
\end{icmlauthorlist}

\icmlaffiliation{cb}{{Université  Paris-Saclay, ENS Paris-Saclay, CNRS,  Centre Borelli, France}}

\icmlcorrespondingauthor{Alejandro de la Concha}{alejandro.de\_la\_concha\_duarte@ens-paris-saclay.fr}
\icmlcorrespondingauthor{Argyris Kalogeratos}{argyris.kalogeratos@ens-paris-saclay.fr}

\icmlkeywords{Machine Learning, ICML}

\vskip 0.3in
]



\printAffiliationsAndNotice{} 

\begin{abstract}
This paper addresses the multiple two-sample test problem in a graph-structured setting, which is a common scenario in fields such as Spatial Statistics and Neuroscience. Each node $v$ in fixed graph deals with a two-sample testing problem between two node-specific probability density functions (\pdfuncs), $p_v$ and $\q_v$. The goal is to identify nodes where the null hypothesis $p_v = \q_v$ should be rejected, under the assumption that connected nodes would yield similar test outcomes. We propose the non-parametric \emph{collaborative two-sample testing} (\CTST) framework that efficiently leverages the graph structure and minimizes the assumptions over $p_v$ and $\q_v$
. Our methodology integrates elements from \fdiv estimation, Kernel Methods, and Multitask Learning. We use synthetic experiments and a real sensor network detecting seismic activity to demonstrate that \CTST outperforms state-of-the-art non-parametric statistical tests that apply at each node independently, hence disregard the geometry of the problem. 
\end{abstract}

\section{Introduction}{\label{sec:introduction}}

Given two probability density functions (\pdfs) $p$ and $\q$, a \emph{Two-sample Test} (TST) assesses if there is significant evidence that the null hypothesis, $\Hnull : p = \q$, is true, versus the alternative $\Halt : p \neq \q$.
TST has been studied in detail in the Machine Learning literature leading to several methods \citep{
Sugiyama2011_LSTST,Gretton2012,Harchaoui2013,lopez2016revisiting,two-sample-test-CB2021}. 
As in most statistical problems, passing from the typical univariate to a multivariate setting is non-trivial
. More precisely, carrying on multiple two-sample test will 
encounter the \emph{Multiple Comparison Problem} (MCP), which refers to the fact that the probability of wrongly rejecting a set of null hypothesis (false positives, or Type-I error), increases artificially with the number of tests. Standard MCP treatments include Bonferroni correction that scales the \pvalues by the number of hypotheses being tested ($\Gdims$) \citep{Dunn1961}, or non-parametric resampling test with a maximum statistic and permutation tests \citep{Westfall1992}.

The \emph{Multiple Two-Sample Testing} (MTST) problem appears in fields such as Spatial Statistics, Neuroscience, or Complex Systems. In these contexts, each test is associated with data sampled from a different `location', and the validity of null hypotheses often depends on the 'proximity' between those locations. For instance, the Hebbian perspective stating ``\emph{Neurons that fire together wire together}'' \cite{hebbian1949} is common ground in Neuroscience, while Tobler's 
\emph{first law of Geography} \cite{Tobler1970} eloquently stating ``\emph{Everything is related to everything else, but near things are more related than distant things}'' is cornerstone in Spatial Statistics. 

\inlinetitle{Multiple two-sample testing on graphs}{.}~%
Motivated by the above application fields, we study the particularly challenging problem of \emph{graph-structured} MTST, where a TST is considered over each node $v \in V=\{1,...,\Gdims\}$ of a given fixed graph $G$, comparing two node-specific \pdfs $p_v$ and $\q_v$. 
Then, the all $\Gdims$ hypotheses 
are simultaneously tested
: 
%
\begin{equation}{\label{eq:statistical_test}}
\begin{aligned}
\!\!\!\!\!\!\!\!\!\!\!\!\!\!\!\big\{\Hnullv :  \ \ p_v=\q_v \ \ \ &\text{\vs} \ \ \  \Haltv : \ \ p_v \neq \q_v\big\}_{v\in V}\\
\end{aligned}
\end{equation}%
%
to determine $R_{\text{MT}} = \{v \in V \padded{|} \Hnullv \text{ is found false}\}$, which contains the nodes with null hypotheses to be rejected with a given level of confidence $1-\pi^*$\footnote{$p$-values appear as \pvalues to distinguish them from \pdf $p$.}. The goal is for $R_{\text{MT}}$ to be as close as possible to the set of hypotheses where really holds $p_v \neq \q_v$, denoted by $\Inull^{_\complement}$ (\ie the set complement of $\Inull$). %
As in any \emph{Multiple Hypothesis Testing} (MT) approach, in this case determining $R_{\text{MT}}$ requires three components:
\begin{enumerate}[topsep=-0.2em,itemsep=-0.35em,leftmargin=1.3em]
    \item A test statistic $S_v$ for $\Hnullv$
, estimated using the data of node $v$, to quantify the dissimilarity of $p_v$ and $\q_v$. 
    \item A \pvalue estimation framework to identify which of the 
		$\{\Hnullv\}_{v\in V}$ to be rejected. 
    \item A Type-I error correction strategy to control the MCP.
\end{enumerate}
In the context of graph-structured MTST, to the best of our knowledge, there exist mostly plug-in methods, in the sense that: they perform (1) and (2) independently for each node; then, for (3) they apply post-hoc Type-I error correction using an aggregation mechanism over the estimated \pvalues, or they avoid the MCP by defining a single test statistic from the multiple test statistics $\{S_v\}_{v \in V}$, and then estimate a \pvalue based on that quantity. The main drawback of these approaches is that individual test statistics that fail to quantify properly the difference between each pair of $p_v$ and $q_v$ may lead to inaccurate conclusions.

Notable graph-structured MT techniques include the \emph{Permutation Cluster Test} (PCT) \citep{Maris2007}, \emph{Threshold-free Cluster Enhancement} (TFCE) \citep{Smith2009}, and the \emph{Structure-Adaptive Benjamini Hochberg Algorithm} (SAHBA) \citep{LiAng2018}, which assume that the null hypotheses to be rejected will be associated with a group of connected nodes. The \pvalues of PCT and TFCE procedures are estimated via a permutation test over a maximum test statistic. In contrast, SAHBA uses a reweighting mechanism of the node-level \pvalues, and relies on the assumption that connected nodes will show similar \pvalues. %

\begin{figure}[t!]
  \centering
\includegraphics[width=\linewidth, trim=0pt 200pt 0pt 40pt,clip]{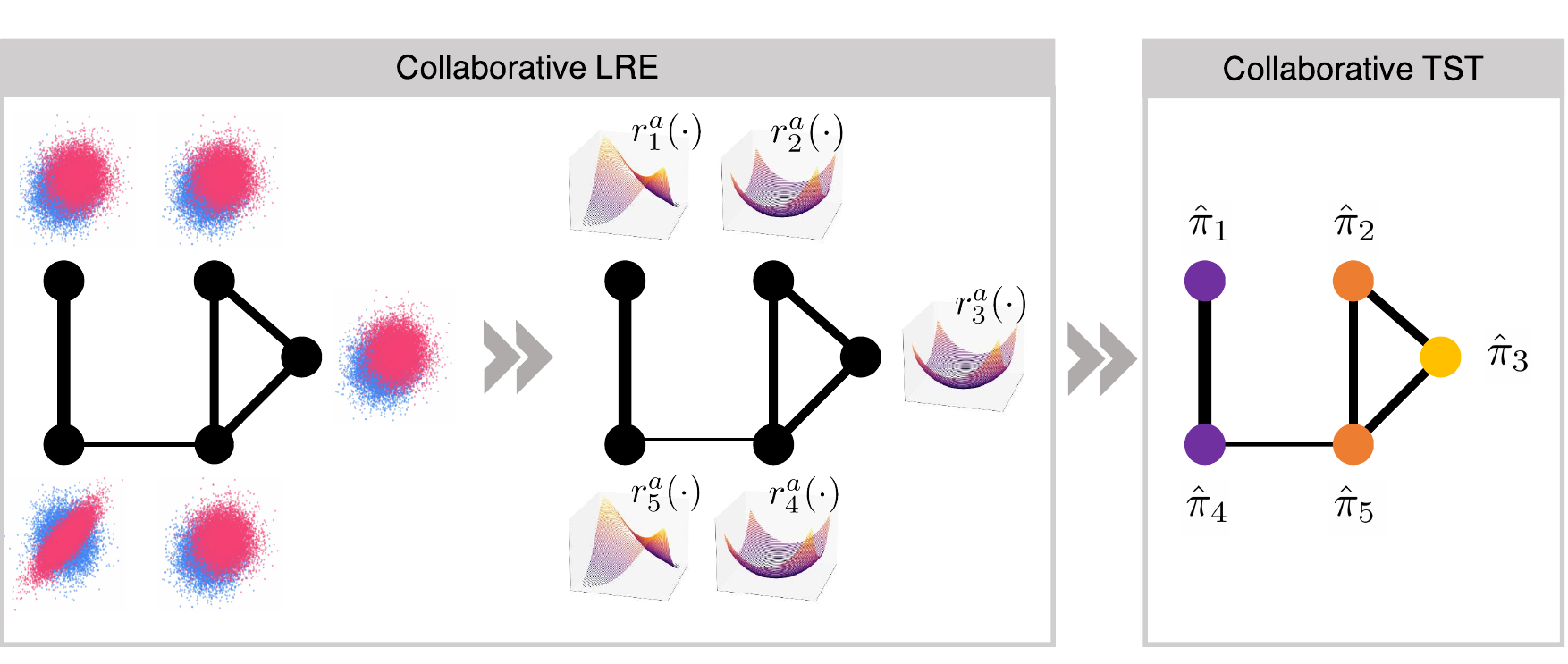
}%
\vspace{-0.5mm}
    \caption{\footnotesize \textbf{Collaborative multiple two-sample testing (CTST) based on collaborative \LRE over a graph.}
\textbf{Left:} 
		Given observations from two	\pdfs, $p_v$ (blue) and $q_v$ (pink) at each node $v$ of a graph, GRULSIF estimates the associated relative likelihood-ratios $\{r_v^{\alpha}\}_v$ in a collaborative manner. 
		In this example, it is easy to see how any given $x\in\X=\R^2$ gets essentially mapped to the graph signal $(r^{\alpha}_1(x),...,r^{\alpha}_\Gdims(x))^{\top}$. 
    \textbf{Right:}
    A visual summary of the CTST testing. 
		The likelihood-ratios computed by GRULSIF are used to estimate node-level $p$-values $\hat{\pi}_v$ 
		that allow us to eventually identify the nodes such that $p_v \neq \q_v$.}
\label{fig:GRULSIF}
\end{figure}

\textbf{Contribution}{.} As a response to the above challenges, in this paper we present the \emph{Collaborative Two-Sample Test} (\CTST): a graph-structured 
TST built upon non-parametric methods and the notion of graph smoothness
. \Fig{fig:GRULSIF} illustrates the approach. Distinct from existing works, \CTST's core novelty is that it not only \emph{estimates jointly and in an associative manner} all node-level test statistics, but it also intertwines that estimation 
with the identification of the hypotheses to be rejected. 
Leveraging techniques from the $\phi$-divergence estimation, Kernel Methods, Multitasking, and more specifically the GRULSIF framework \cite{delaconcha2024collaborative}. 
\CTST adeptly quantifies the 
difference  between $p_v$ and $\q_v$ 
under minimal assumptions. Under the graph smoothness hypothesis, the collaborative estimation enforces the similarity of the test statistics $S_u$ and $S_v$ for connected nodes $u$ and $v$. The induced regularity at the node-level test statistics is exploited by a permutation test that efficiently controls for the \emph{$k$-Family-Wise Error Rate} (FWER), and 
identifies of nodes where $p_v \neq \q_v$. Our experimental study using synthetic data and real seismic data, shows that \CTST compares favorably against state-of-the-art Kernel-based 
techniques that disregard the geometry of the problem.

\section{Preliminaries and problem statement} \label{sec:background}

\subsection{Preliminaries}\label{sec:preliminaries}

\inlinetitle{General notations}{.}~Let $a_i$ be the $i$-th entry of a vector $a$; when the vector is itself indexed by $j$, we refer to its $i$-th entry by $a_{j,i}$. $A_{ij}$ denotes the entry at the $i$-th row and $j$-th column of a matrix $A$, and $A_{i,:}$ is its $i$-th row.  $\Vect(a_1,..,a_n)$ denotes the concatenation of the input vectors $a_1,...,a_{n}$ in a single vector. $\ones{M}$ is a vector with $M$ ones (resp. $\zeros{M}$)
, $\id_{M}$ is a $M \times M$ identity matrix, and $\Ind{\cdot}$ is the indicator function. The Euclidean norm and the dot product are denoted by $\norm{\cdot}$ and $\dott{\cdot}{\cdot}$. When those are endowed to a functional space $\mathcal{F}$, we write $\norm{\cdot}_{\mathcal{F}}$ and $\dott{\cdot}{\cdot}_{\mathcal{F}}$. For an observation $x$ belonging to a $d$-dimensional input space, we write $x \in \mathcal{X} \subset \R^d$.
 
A fixed undirected weighted graph $\G=(V,E,W)$ is defined by the set of $\Gdims$ nodes $V$, and the set of edges $E$. Throughout the rest of the presentation, we suppose that the edges are positive-weighted and undirected, and that the nodes have no self-loops, \ie the entries of its weight matrix $W \in \real^{\Gdims\times \Gdims}$ are such that $W_{uu}=0$, $\forall u \in V$, and $W_{uv}=W_{vu} \geq 0$.  In the rest, composite objects (vectors, matrices, sets, \etc) that refer to all the nodes of a graph, are denoted in bold font. Finally, the notion of \emph{smoothness} is central in this work; the smoothness of a graph function $\vartheta : V \rightarrow \R $ over $G$ is defined as $\sum_{(u,v)\in E} W_{uv} (\vartheta(u)-\vartheta(v))^2$. This notion generalizes for $N$ estimates over the nodes of $G$, hence we use the umbrella term \emph{graph smoothness} to refer to the expected behavior of a studied phenomenon over a graph, which in turn motivates the use of graph regularization techniques.

\inlinetitle{\fdivs and likelihood-ratio}{.}~%
\fdivs are non-negative functions measuring the dissimilarity between two probability measures. For two probability measures with \pdfuncs $p$ and $\q$ with respect to the Lebesgue measure, the \fdiv comparing $p$ and $\q$ is 
defined as:
\begin{equation}\label{eq:fdiv-main}
\mathcal{D}_{\phi}(p \Vert \q) = \int \phi\!\bigg(\frac{\q(x)}{p(x)}\bigg) p(x) dx,\ \ \text{for }x\in\X,
\end{equation}
where $\phi:\real \rightarrow \real$ is a convex and semi-continuous real function such that $\phi(1) = 0$ \citep{Csiszar1967}. 
%
Easy to see, $\mathcal{D}_{\phi}(p \Vert \q) = 0$ iff $p = \q$. Moreover, as the integration in \Eq{eq:fdiv-main} is \wrt $p$, the output is more sensitive to points where $p$ has higher mass, and hence \fdivs may be non-symmetric functions, \ie $\mathcal{D}_{\phi}(p \Vert \q) \not\equiv \mathcal{D}_{\phi}(\q \Vert p)$. 

The quantity $r(x)=\frac{q(x)}{p(x)}$ is called \emph{likelihood-ratio} and is central in the computation of any \fdiv. As we will see in \Sec{sec:estimation_CTST}, we can translate the approximation of the \PEdiv between $p$ and $q$ to a likelihood-ratio estimation (\LRE) problem. In practice, though, $r$ may be an unbounded function, challenging non-parametric methods that may fail to converge. For this reason, a known workaround is to replace $p$ by $p^{\alpha}(x)=(1-\alpha)p(x)+\alpha q(x)$, and use instead the $\alpha$-\emph{relative \likelihood-ratio function} \citep{Yamada2011}: $r^{\alpha}(x) 
= \frac{q(x)}{p^{\alpha}(x)}
\leq \frac{1}{\alpha}$, 
for any $0 \leq \alpha < 1$,  
$x \in \X$.

\subsection{Problem statement}{\label{sec:problem_statement}}

Let a fixed undirected and positive-weighted graph $\G=(V,E,W)$, and suppose each node $v \in V$ has $\Npre + \Npost$ (same for all nodes) \iid observations from two unknown \pdfs, $p_v$ and $\q_v$, respectively. The two data observations subsets taking values in the input space $\mathcal{X} \subset \R^d$ are:
%
\begin{equation}{\label{eq:sample_per_nodes}}
\left\{
    \begin{aligned}
        \mathbf{X} &= \{\mathbf{X}_v\}_{v\in V}= \big\{\{x_{v,i} \,:\, x_{v,i}\,\overset{\text{\iid}}{\sim}\, p_v\}_{i=1}^{\Npre}\big\}_{v \in V};
          \\
					\mathbf{X}' &= \{\mathbf{X}'_v\}_{v\in V}= \big\{\{x'_{v,i} \,:\, x'_{v,i}\,\overset{\text{\iid}}{\sim}\, q_v\}_{i=1}^{\Npost}\big\}_{v \in V}.
    \end{aligned}
    \right.
\end{equation}
The proposed \CTST aims at solving the graph-structured multiple two-sample testing 
problem presented in \Expr{eq:statistical_test}. \Fig{fig:GRULSIF} presents an insightful visualization of the problem. In general terms, \CTST comprises three steps: 
\begin{enumerate}[topsep=0em,itemsep=0.0em,leftmargin=1.3em]
    \item[1.] \emph{Collaborative estimation}: 
		Joint estimation of the node-level relative likelihood-ratios, $\mathbf{r}^{\alpha} \mydef (r^{\alpha}_1,...,r^{\alpha}_{\Gdims})$,
	using the available data 
(\Eq{eq:sample_per_nodes}). 
	The vector-valued function $\mathbf{r}^{\alpha}$ is then used to approximate for each node $v$ the 
\PEdiv\!$(p_v\Vert\q_v)$.
\item[2.] \emph{Node-level test statistics}: 
The \fdivs' properties (see \Sec{sec:preliminaries}) make them good candidates 
for node-level test statistics. To deal with their non-symmetricity, at each node the pair of node-level test statistics $\{S_v\}_{v \in V}$, $\{S'_v\}_{v \in V}$ 
are used, which corresponds to both the \PEdiv\!$(p_v\Vert\q_v)$ and \PEdiv\!$(\q_v\Vert p_v)$.

\item[3.] \emph{\pvalue estimation}: A permutation test is used for the estimation of two sets of node-level \pvalues, $\{\pi_v\}_{v \in V}$ and $\{\pi'_v\}_{v \in V}$
. These sets of \pvalues to identify the set of null hypotheses to be rejected ($R_{\text{\CTST}}$). The permutation test guarantees weak control of FWER. 
\end{enumerate}

\section{The proposed collaborative non-parametric two-sample test (\CTST)}\label{sec:CTST}

The foundation of the \CTST method is the collaborative likelihood-ratio estimation (\LRE)
in a graph-structured setting. Conveniently for our purpose, this problem has been formally introduced in \cite{delaconcha2024collaborative}, and the \emph{Graph-based Relative Unconstrained Least Squares Importance Fitting} (GRULSIF) method has been proposed, which we employ in this work. Before presenting the principal components of \CTST, we mention below basic notions regarding that non-parametric estimation. 

\inlinetitle{Reproducing Kernel Hilbert Spaces}{.}~%
%
Given an input space $\mathcal{X} \subset \R^d$, 
we aim to estimate $r^{\alpha}(x)$ \wrt a Reproducing Kernel Hilbert Space (RKHS) 
$\Hilbert$ containing as elements 
functions $f:\mathcal{X} \rightarrow \R$. 
$\Hilbert$ is equipped with the inner product $\dotH{\cdot}{\cdot}:\Hilbert \times \Hilbert \rightarrow \R$, which will be reproduced by a Mercer Kernel; \ie~by a continuous symmetric real function, which is the positive semi-definite kernel function $\KernelFunc(\cdot,\cdot) : \mathcal{X} \times \mathcal{X} \rightarrow \R$. 
Then, the space $\Hilbert$ 
enjoys the so-called RKHS reproducing property: $\dotH{\KernelFunc(x,\cdot)}{f} = f(x)$, for any $f \in \Hilbert$; and also satisfies that $\Hilbert = \overline{\operatorname{span}}(\{\KernelFunc(x,\cdot) : \forall x \in \mathcal{X}\})$, where $\overline{\operatorname{span}}$ refers to the closure of all the linear combinations of the elements $\KernelFunc(x,\cdot) \in \Hilbert$, $\forall x \in \mathcal{X}$. 
Finally, the earlier seen concept of smoothness can be generalized in the RKHS: for $\vartheta(u),\vartheta(v)\in \Hilbert$, this is $
\sum_{(u,v)\in E} W_{uv} \norm{\vartheta(u)-\vartheta(v)}^2_{\Hilbert}$.

\subsection{Step 1: Collaborative likelihood-ratio estimation}{\label{sec:estimation_CTST}}

The graph-based framework for \LRE \citep{delaconcha2024collaborative} that we employ for this 
step, focuses on the \PEdiv. By setting $\phi(r(x)))=\frac{(r(x)-1)^2}{2}$ in \Eq{eq:fdiv-main}, one recovers the \PEdiv \citep{Pearson1900}, $\mathcal{D}_{\phi}(p \Vert \q) = \PE(p \Vert q)$, which can be expressed as: 
\begin{subequations}{\label{eq:div_ratio}}
\begin{align}
\!\!\!\!\!\!\!\PE(p \Vert q) &:=
\int \frac{(r(x)-1)^2}{2}p(x)dx \label{eq:div_ratio_original}\\
& \geq \sup_{f \in \mathcal{F}} \int\!\! f(x) q(x) dx - \!\!\int \frac{f^2(x)}{2}  p(x) dx - \frac{1}{2},\!\!\label{eq:variational_representation}
\end{align}
\end{subequations}
%
where $\mathcal{F}$ is a functional space. \Ineq{eq:variational_representation} is known as the variational representation of the \PEdiv, and it is a consequence of  Lemma\,1 in \cite{Nguyen2007} that gives the conditions where such lower-bound is attained, \ie 
when computing a \fdiv amounts to solving an optimization problem in a functional space. In the case of \PEdiv, the function $f$ appearing in \Ineq{eq:variational_representation} approximates the likelihood-ratio  
$r$.

As reasoned in \Sec{sec:preliminaries}, instead of estimating $r$, it is suggested to work with the relative likelighood-ratio 
$r^{\alpha}$, hence to estimate $\PE^{\alpha}(p  \Vert q) := \PE(p^{\alpha} \Vert q)$.
Finally, we can express the variational representation of \Ineq{eq:variational_representation} in expectation, using the data observations described in \Expr{eq:sample_per_nodes}:
%
\begin{align}\label{eq:variational_representation_exp}
\Expec{\PE(p^\alpha \Vert q)} & \geq \ \sup_{f \in \mathcal{F}}\ \Exp_{q(x')}[f(x')]  - \frac{(1-\alpha)}{2}\Exp_{p(x)}[f^2(x)] \nonumber\\
& -  \frac{\alpha}{2}\Exp_{\q(x')}[f^2(x')] - \frac{1}{2}.
\end{align}%
The choice of the functional space  $\mathcal{F}$ is key for defininig a learning algorithm that can be implemented in practice, and at the same time enjoying desired theoretical properties such as stability and consistency. In our approach, we opt for a \RKHS whose geometry can enhance 
the graph smoothness hypothesis. This derives from the fact that two functions $f_u,f_v \in \Hilbert$ close in the \RKHS, will exhibit similarity when evaluated at the same point  $x \in \mathcal{X}$, as elucidated below: 
\begin{equation}{\label{eq:similarity_ratios}}
\!\!\!\!\!\!\!\!\abs{f_u(x)\!-\!f_v(x)}=\abs{\dotH{\KernelFunc(x,\cdot)}{f_u\!-\!f_v}} \leq C \norm{f_u\!-\!f_v}_{\Hilbert}\!,\!\!\!\!\!\!
\end{equation}
where $0<C<\infty$ is a constant so that $\sup_{x \in \X} \KernelFunc(x,x) \leq C$. The first inequality is a consequence of the reproducing property of $\Hilbert$, and the second, is a consequence of the Cauchy-Schwarz inequality. Thus, enforcing graph smoothness, \ie $\norm{f_u-f_v}_{\Hilbert}$ to be small for two adjacent nodes $u$ and $v$, is expected to lead to similar \PEdiv estimates.

\inlinetitle{Optimization problem}{.}~%
The aim is to learn the vector-valued function $\mathbf{r}^{\alpha} = (r^{\alpha}_1,...,r^{\alpha}_\Gdims)$ via $\mathbf{f} = (f_1,...,f_\Gdims) \in \Hilbert^{\Gdims}$, where $\Hilbert$ is a scalar \RKHS. The cost function to optimize is:
\begin{equation}
{\label{eq:multitasking}}
\begin{aligned}
 & \!\!\!\!\!\!\!\!\!\!\!\!\!\!\min_{\mathbf{f} \in \Hilbert^{\Gdims}} \frac{1}{N} \sum_{v \in V} \!\Big( \frac{(1-\alpha)}{2} \ExpecNm{f_v^2(x)}{v}  + \frac{\alpha}{2} \ExpecAm{f_v^2(x')}{v}  \\
&  \!\!- \ExpecAm{f_v(x')}{v} \Big) + \frac{\lambda}{4} \sum_{u,v \in V} \!\!\! W_{uv} \norm{f_u-f_v}^2_{\Hilbert} 
+ \frac{\lambda \gamma}{2} \sum_{v \in V} \norm{f_v}^2_{\Hilbert}.\!\!\!\!\!\!\!\!\!
\end{aligned}
\end{equation}
The first term corresponds to the negative 
variational representation of the \PEdiv at each node (\ie the non-constant terms of \Expr{eq:variational_representation_exp}). 
The second term evaluates the graph smoothness of the estimates. The last one is a penalty term that reduces the risk of overfitting \citep{Sheldon2008}.

Provided a dictionary $D_{\hat{L}}$ of $\hat{L}$ basis functions, such that the finite dimensional space $\mathbf{F}={\spankern(\{\varphi(x) : x \in D_{\hat{L}}\})}$ approximates $\Hilbert$, 
it was further proposed to use Nyström approximation to replace the feature map $\varphi(x)$ by its orthogonal projection into the  space $\mathbf{F}$. By determining a set of so-called \emph{anchor points} in $\Hilbert$, $\varphi(x_1),...,\varphi(x_{\hat{L}})$, and via the associated kernel matrix,  
$\Gramm_{\hat{L}} \in \R^{\hat{L}\times\hat{L}}$, $[\Gramm_{\hat{L}}]_{ij}=\KernelFunc(x_i,x_j)$, the new feature map derives: 
\begin{equation}{\label{eq:psi_definition}}
    \psi(\cdot)=\Gramm_{\hat{L}}^{-\frac{1}{2}} \left(\KernelFunc(\cdot,x_1),..,\KernelFunc(\cdot,x_{\hat{L}})\right)^{\top}.%
\end{equation}
It was shown that, writing \Problem{eq:multitasking} in terms of the empirical expectations and by involving the Nyström approximation, 
its solution $\hat{\mathbf{f}}=(\hat{f}_1,...,\hat{f}_{\Gdims}) \in \Hilbert^N$, 
takes the form: 
\begin{equation}{\label{eq:approx_f_v}}
    \hat{f}_v(\cdot)= \psi(\cdot)^{\top} \hat{\theta}_v,  
\end{equation}
where $\hat{\theta}_v  \in \R^{L}$. By defining 
$\ThetaG=\Vect(\theta_1^{\top},...,\theta_{\Gdims}^{\top})^{\top} \in \R^{\Gdims\hat{L}}$ that vectorizes all the node parameters, \Problem{eq:multitasking} is rewritten as a quadratic problem over $\ThetaG$: 
%
\begin{align}{\label{eq:Nyström_problem}}
\min_{\ThetaG \in \real^{\Gdims \hat{L} }}  & \frac{1}{\Gdims} \sum_{v \in V }  \left( \frac{1-\alpha}{2} \theta_v ^\top  H_{\psi,v} \theta_v + \frac{\alpha}{2} \theta_v ^\top  H'_{\psi,v} \theta_v  -  h'_{\psi,v} \theta_v \right) \nonumber\\
& + \frac{\lambda}{4} \sum_{u,v \in V} \! W_{uv} \norm{\theta_v-\theta_u}^2
+ \frac{\lambda \gamma}{2} \sum_{v \in V } \norm{\theta_v}^2\!,
\end{align}
%
\vspace{-1.2em}
%
\begin{align}{\label{eq:updatehs}}
\text{where}\ \ H_{\psi,v} &= \frac{1}{\Npre_v}\sum_{x \in \mathbf{X}_v} \! \psi(x) \psi(x)^\top, \ \ 
h_{\psi,v}'= \frac{1}{\Npost_v}\sum_{x \in \mathbf{X}'_v} \! \psi(x), \nonumber\\ 
\!\!\!\!H_{\psi,v}' &= \frac{1}{\Npost_v}\sum_{x \in \mathbf{X}'_v} \! \psi(x)\psi(x) ^\top.
\end{align}%
%
Notice that $H_{\psi,v},H_{\psi,v}' \in \R^{\hat{L} \times \hat{L}}$ and $h_{\psi,v}' \in \R^{\hat{L}}$. 

\inlinetitle{Implementation}{.}~%
We follow the implementation of \cite{delaconcha2024collaborative}, which proposed to solve \Problem{eq:Nyström_problem} with the Cyclic Block Coordinate Descent (CBCD) \citep{Beck2013,Li2018}. If $n=n'$, then the final computational cost 
is $\bigO(\Gdims \hat{L}^3+  n \Gdims \hat{L}^2+
\Gdims \hat{L}^2 \log^2(\Gdims \hat{L}))$, where $\hat{L} \ll \Gdims n$, which makes it scalable to real-life graphs. Other important implementation elements of GRULSIF, which we do not detail here, are the selection of the anchor points and the choice of the hyperparameters; the latter refers to the parameters of the kernel function $\KernelFunc(\cdot,\cdot)$ and the regularization constants $\lambda$, $\gamma$. Since the regularization parameter $\alpha$ requires special attention, we provide several enlightening experiments for the studied \CTST task in \Appendix{appendix:experimentsdetailsCM2ST}.

\subsection{Step 2: Node-level test statistics}\label{sec:node-level-statistics}

After the parameter vector $\hat{\ThetaG}$ has been estimated, we can approximate the 
\PEdiv\!$(p^{\alpha}_v \Vert \q_v)$ by: 
%
\begin{align}{\label{eq:PE_div_GRULSIF}}
   \!\!\PE^{\alpha}(p_v \Vert q_v) &:=  \PE(p_v^{\alpha}\Vert q_v) \nonumber\\
   & \approx h'^{\top}_{\psi,v} \hat{\theta}_v - \frac{1-\alpha}{2} \hat{\theta}_v ^\top  H_{\psi,v} \hat{\theta}_v - \frac{\alpha}{2}  \hat{\theta}_v ^\top  H'_{\psi,v} \hat{\theta}_v  - \frac{1}{2} \nonumber\\
   & =: \hat{\PE}_v^{\alpha}(\mathbf{X}_v \Vert \mathbf{X}'_v). 
\end{align}
%
To address the issue of the non-symmetricity of divergence (see below \Eq{eq:fdiv-main}), we identify  the set of hypotheses to be rejected ($R_{\textup{\CTST}}$) by considering both the comparisons $\PE^{\alpha}(p \Vert \q)$ and $\PE^{\alpha}(\q \Vert p)$ to derive two sets of test statistics: 
\begin{equation}\label{eq:test-statistics}
\begin{aligned}
   \{S_v\}_{v \in V} &=  \{ \hat{\PE}_v^{\alpha}(\mathbf{X}_v \Vert \mathbf{X}'_v) \sim \PE^{\alpha}(p_v \Vert \q_v)\}_{v \in V}; \\ 
	\{S'_v\}_{v \in V}&=  \{ \hat{\PE}_v^{\alpha}(\mathbf{X}'_v \Vert \mathbf{X}_v)\sim \PE^{\alpha}(\q_v \Vert p_v)\}_{v \in V}. 
\end{aligned}
\end{equation}
It has been shown that $\hat{\PE}_v^{\alpha}(\mathbf{X}_v \Vert \mathbf{X}'_v)$ is an asymptotic unbiased estimator of $\PE^{\alpha}(p_v \Vert q_v)$, and that the graph smoothness hypothesis and the collaborative \LRE becomes more relevant as the estimation problem becomes more challenging \cite{delaconcha2024collaborative}, \eg the fewer are the available observations per node. 
Notice, that the graph smoothness hypothesis would be totally satisfied if $\forall v \in V$, $p_v=\q_v$, since all the relative likelihood-ratios will be equal to $1$, hence $\norm{r^{\alpha}_u -r^{\alpha}_v}_{\Hilbert}=0$ for all connected nodes, $u$, $v$.  

In this work, we exploit these properties of the collaborative 
\LRE to propose a permutation test to control for FWER under the global hypothesis that for all nodes $p_v=q_v$ ($\Hnull$), but still sensitive enough to 
distinguish the nodes that experience a change of measure.

\subsection{Step 3: \pvalue estimation}{\label{sec:p_values_estimation}}

%
%
Our MT strategy 
applies a threshold $\eta^*$ to each estimated node-level \pvalue $\{\hat{\pi}_v\}_{v \in V}$, hence considers the set of rejected hypotheses $R_{\text{MT}}=\{v \in V \padded{|} \hat{\pi}_v < \eta^*\}$. We denote by $\TP=\#\{v \padded{|} v \in \mathbf{I}_0 \cap R_{\text{MT}} \}$ the number of true positives, and by $\FP=\#\{v \padded{|} v \in R_{\text{MT}} \setminus \mathbf{I}_0 \}$ the number of false positives. 
We address the MCP by weak control of the \emph{$k$-Family-Wise Error Rate} (FWER for $k=1$), which is to control the probability to occur at least one false rejection of the individual node-level hypotheses: 
\begin{equation}{\label{eq:FWER_control}}
    \Prob(\FP \geq 1 \padded{|} \mathbf{I}_0) = \Prob(\{ \exists v \in \mathbf{I}_0: \hat{\pi}_v < \eta^*\}) \leq \pi^*,
\end{equation}
where $\pi^*$ is a user-defined rate (\eg $0.01$ or $0.05$). 
Henceforth, we consider the following null hypothesis: 
\begin{equation}\label{eq:Hnull}
\Hnull : p_v = \q_v, \forall v\in V.
\end{equation}%
Unlike \emph{strong FWER control} that refers to any subset $\mathbf{I}_0 \subset V$, \emph{weak FWER control} is less demanding as it deals only with the case where $\mathbf{I}_0=V$. 

Weak control in MTST is particularly relevant when studying the behavior of complex systems under two different experimental conditions. When there is no statistically significant difference between both conditions, then all nodes are expected to satisfy the null hypothesis (\Expr{eq:Hnull}). Neuroscience offers a good example of this situation: several sensors are used to monitor brain activity, and MTST aims to detect clusters of firing neurons to a given stimulus. Classical methods, such as PCT \citep{Maris2007} and TFCE \citep{Smith2009}, account for the inherent graph structure of the brain function and perform weak FWER control. By 
this, they  mitigate the risk of claiming a false difference between two experimental conditions, while still maintaining the sensitivity necessary to detect true neural activity patterns.

The graph regularization introduced by the collaborative \fdiv estimation
of Step $1$, leads to robust estimators against outliers in the node-level test statistics. This is particularly relevant under $\Hnull$ and we want to avoid false positives, thus it is natural to exploit this feature and design a \pvalue estimation procedure with weak FWER control. Bear in mind that the flexibility of non-parametric \LRE allows a certain level of heterogeneity of the \pdfs in $\{p_v\}_{v \in V}$ and $\{\q_v\}_{v \in V}$, as long as the relative likelihood-ratios of the pairs ($(p_v,q_v)$ and $(p_u,q_u)$, $(u,v)\in E$) in adjacent nodes can be approximated by functions that are close in the shared \RKHS. This feature complicates the distribution of the test statistic under $\Hnull$ and, consequently, the derivation of an explicit formula for FWER control. Moreover, for the intended applications it is important to account for correlations between the node-level estimates. 

We propose the use of a permutation test, which is a non-parametric strategy to address the above challenges without  restricting our framework \cite{Westfall1992}. %
The designed permutation test is over the vectors $X_{:,j}=(x_{1,j},...,x_{\Gdims,j})^{\top}$ (resp. $X'_{:,j}=(x'_{1,j},...,x'_{\Gdims,j})^{\top}$), each one carrying the observations having a given sample index $j$ for all nodes. The permutation test infers the distribution of the maximum test statistic $S_{G}=\max_{v \in V} S_v$, 
and uses it to determine $R_{\text{\CTST}}$, achieving this way weak FWER control at the level of $\pi^*$'s. 

The complete \CTST algorithm is provided in \Alg{alg:TS-GRULSIF}.
{\newcommand{\algorithmicindent}{2.0em}
\begin{algorithm}[!t]
\scriptsize
   \caption{\textbf{--} Collaborative  two-sample tests over a graph (\CTST) 
   }\label{alg:TS-GRULSIF}
\begin{algorithmic}[1]
  \STATE {\bfseries Input:} $\mathbf{X}, \mathbf{X}'$: two samples with observations over the graph $\G=(V,E,W)$; \\
 \STATE \hspace*{\algorithmicindent}  $\alpha \in [0,1)$: parameter of the relative \likelihood-ratio;\\
 \STATE  \hspace*{\algorithmicindent} 
 $n_{\text{perm}}$: the number of random permutations 
for \pvalue computation;\\
 \STATE  \hspace*{\algorithmicindent} $\pi^*$: the FWER rate for the test required by the user. 
 \STATE {\bfseries Output:} $\{\hat{\pi}_v\}_{\{v \in V\}}$, $\{\hat{\pi}'_v\}_{\{v \in V\}}$: a pair of \pvalues 
for each node;\\
 \STATE \hspace*{\algorithmicindent} $R_{\textup{\CTST}}$: nodes where the null hypothesis $\Hnullv: p_v= \q_v$ is rejected.
\vspace{1.3mm}
\hrule
\vspace{1.3mm}

\raisebox{0.25em}{{\scriptsize$_\blacksquare$}}~\textbf{Produce the required elements to define $\hat{\PE}{}^{\alpha}_1(\mathbf{X},\mathbf{X}')$ and $\hat{\PE}{}^{\alpha}_2(\mathbf{X}',\mathbf{X})$}

\STATE Compute the anchor points associated with the kernel $\KernelFunc : \mathcal{X} \times \mathcal{X} \rightarrow \R$ \hfill\! \hfill (see $\star$)
\STATE Select the hyperparameters $\sigma^*_1,\lambda^*_1,\gamma^*_1$, $\sigma^*_2,\lambda^*_2,\gamma^*_2$ \hfill (see $\star$)

\raisebox{0.25em}{{\scriptsize$_\blacksquare$}}~\textbf{Compute the node-level test-statistics on the observed data}

\STATE Estimate $\hat{\ThetaG}_1(\mathbf{X},\mathbf{X}') = \text{GRULSIF}(\mathbf{X},\mathbf{X}',\alpha,\sigma^*_1,D_1,\gamma^*_1,\lambda^*_1)$\\
\phantom{Estimate} $\hat{\ThetaG}_2(\mathbf{X}',\mathbf{X}) =\text{GRULSIF}(\mathbf{X}',\mathbf{X},\alpha,\sigma^*_2,D_2,\gamma^*_2,\lambda^*_2)$

\STATE Compute $S_v=\{\hat{\PE}{}^{\alpha}_v(\setprev,\setpostv)\}_{v\in V}$ and $S'_v=\{\hat{\PE}{}^{\alpha}_v(\setpostv,\setprev)\}_{v\in V}$ \\\hspace*{\algorithmicindent} using 
$\hat{\ThetaG}_1(\mathbf{X},\mathbf{X}')$, $\hat{\ThetaG}_2(\mathbf{X}',\mathbf{X})$\hfill (see \Expr{eq:PE_div_GRULSIF})

\raisebox{0.25em}{{\scriptsize$_\blacksquare$}}~\textbf{Permutation test}

\FOR{$ i \in \{1,...,n_{\text{perm}}\}$}

\STATE  
Generate a random permutation $\tau$ of the set 
such that\\ \hspace*{\algorithmicindent}$\mathbf{X}^{(\tau)}=\{X_{:,\tau(1)},...,X_{:,\tau(n+n')}\}$\\  

\STATE Assign the first $\Npre$ elements of $\mathbf{X}^{(\tau)}$ to the set $\dot{\mathbf{X}}$ and the rest $\Npost$ to the 
$\dot{\mathbf{X}}'$

\STATE Compute $\hat{\ThetaG}_{1}(\dot{\mathbf{X}},\dot{\mathbf{X}}')$ and $\hat{\ThetaG}_{2}(\dot{\mathbf{X}}',\dot{\mathbf{X}})$ 

\STATE Compute $\{\hat{\PE}{}^{\alpha}_v(\dot{\mathbf{X}}_v,\dot{\mathbf{X}}_v')\}_{v \in V}$ and $\{\hat{\PE}{}^{\alpha}_v(\dot{\mathbf{X}}_v',\dot{\mathbf{X}}_v)\}_{v \in V}$ 
\\ \hspace*{\algorithmicindent}using $\hat{\ThetaG}_{1}(\dot{\mathbf{X}},\dot{\mathbf{X}}')$, $\hat{\ThetaG}_{2}(\dot{\mathbf{X}}',\dot{\mathbf{X}})$
\STATE Compute the test statistic $s_1^i=\max_v \{\hat{\PE}{}^{\alpha}_v(\dot{\mathbf{X}}_v,\dot{\mathbf{X}}_v')\}_{v \in V}$ \\\hspace*{\algorithmicindent}and $s_2^i=\max_v \{\hat{\PE}{}^{\alpha}_v(\dot{\mathbf{X}}_v',\dot{\mathbf{X}}_v)\}_{v \in V}$

\ENDFOR

\FOR{$v \in \{1,...,\Gdims\}$} 
\STATE 
$\hat{\pi}_v= \frac{1}{n_{\text{perm}}} \sum_{i=1}^{n_{\text{perm}}} \Ind{S_v \leq s_1^i}$  

\STATE 
$\hat{\pi}'_v= \frac{1}{n_{\text{perm}}} \sum_{i=1}^{n_{\text{perm}}} \Ind{S'_v \leq s_2^i}$ 
\ENDFOR
\STATE 
\raisebox{0.25em}

\raisebox{0.25em}{{\scriptsize$_\blacksquare$}}~\textbf{Identify the nodes where the null hypothesis should be rejected}

\STATE Define the set  $R_{\textup{\CTST}}=\{v \in V \padded{|} \hat{\pi}_v \leq \frac{\pi^*}{2} \ \ \text{or} \ \ \hat{\pi}'_v \leq \frac{\pi^*}{2} \}$

\STATE \textbf{return} $\{\hat{\pi}_v\}_{\{v \in V\}}$, $\{\hat{\pi}'_v\}_{\{v \in V\}}$, $R_{\textup{\CTST}}$
\end{algorithmic}
\hrule\leavevmode \\
$\star$ Steps $7$ \& $8$ are treated using the implementation of \cite{delaconcha2024collaborative}.
\end{algorithm}
\Theorem{th:2sample} validates that \CTST is a MT procedure with weak FWER control, provided a user-defined rate $\pi^*$. The technical details of the proof are provided in \Appendix{appendix:2sample}.

\begin{theorem}{\label{th:2sample}} 
Consider \Problem{eq:statistical_test} and 
assume the observations $\mathbf{X}=\{\setprev\}_{v\in V}$
are \iid for each node $v\in V$, same for and $\mathbf{X}'= \{ \setpostv \}_{v\in V}$ (see \Eq{eq:sample_per_nodes}). 
Let $\dot{\mathbf{X}}$, $\dot{\mathbf{X}}'$ the permuted datasets as described in \Alg{alg:TS-GRULSIF} and $\pi^*$ a user-defined rate. Let 
 $F(\,\cdot \padded{|} \mathbf{X} \cup \mathbf{X}')$ denote the probability distribution of $S(\dot{\mathbf{X}} \Vert\dot{\mathbf{X}}')=\max_{v \in V} \hat{\PE}{}^{\alpha}_v(\dot{\setprev} \Vert \dot{\setpostv})$ given $\mathbf{X} \cup \mathbf{X}'$ and let
    $\hat{q}(\mathbf{X} \cup \mathbf{X}')=\sup\{s \in \R \padded{|} F(s\padded{|} \mathbf{X} \cup \mathbf{X}') \leq 1-\frac{\pi^*}{2}\}$ 
be the point determining the upper $((1-\frac{\pi^*}{2})\cdot 100)$-percentile. Then, if $\Hnull$ is true, that is $p_v=q_v$, $\forall v \in V$, then it holds:%
\begin{equation}
    \Prob(S >  \hat{q}'(\mathbf{X} \cup \mathbf{X}')) \leq \frac{\pi^*}{2}.
\end{equation}
Moreover, when $S'=\max_{v \in V} \hat{\PE}{}^{\alpha}_v(\dot{\setpostv} \Vert  \dot{\setprev} )$ is used as a test statistic, then, under $\Hnull$ we have:
\begin{equation}
    \Prob(S >  \hat{q}(\mathbf{X} \cup \mathbf{X}')) \ \ \text{or} \ \ S' >  \hat{q}'(\mathbf{X} \cup \mathbf{X}'))  \leq \pi^*,
\end{equation}
which implies $\FWER(R_{\textup{\CTST}})=\Prob(\FP\geq 1|\Hnull) \leq \pi^*$.
\end{theorem}

\subsection{A \CTST variant ignoring the graph structure}\label{sec:CTST-POOL}

We can derive a reduced \CTST variant relying on POOL, which is a GRULSIF variant that makes use of the same estimation framework, but neutralizes the graph component (\ie by setting $W = \zeros{\Gdims\times\Gdims}$ in \Eq{eq:Nyström_problem}) \cite{delaconcha2024collaborative}. %
This POOL-\CTST variant can be relevant 
when there is no graph underlying the MT problem. Note that POOL can be seen as a variant of RULSIF \citep{Yamada2011,Yamada2013_RULSIFjournal}, while differing to the fact that: i) its joint hyperparameter selection for all nodes based on the mean score $\frac{1}{\Gdims} \sum_{v \in V} \left( \frac{1-\alpha}{2} \theta_v ^\top  H_{\psi,v} \theta_v + \frac{\alpha}{2}  \theta_v ^\top  H'_{\psi,v} \theta_v -h'^{\top}_{\psi,v} \theta_v  \right)$, compared to RULSIF's independent hyperparemeter selection for each node; ii) it uses the Nyström dimensionality reduction technique over all the full data observations, while RULSIF uses a simple uniform random subsampling at each node or all the data \citep{Sugiyama2012}.


\section{Experiments}{\label{sec:experiments}}

\CTST is put in action in the context of graph-structured MT
, in synthetic and real scenarios. 
The goal is to show the gains of combining a non-parametric graph-based collaborative estimation of node-level test statistics, with the weak FWER control based on permutation test with a maximum statistic. Note that \CTST is not a direct competitor to methods such as SAHBA, PCT, or TFCE (see \Sec{sec:introduction}). In fact, those can be seen as complementary approaches to \CTST, as they could post-process \CTST's output. 
Studying how to combine these approaches is beyond the scope of this paper. %

To make fair comparisons and keep the flexibility of non-parametric methods, we restrict our attention to estimation approaches built upon Kernel Methods. We compare against \LRE-based non-parametric algorithms
, where the test statistics correspond to \fdiv estimates, and against kernel-based methods built upon MMD \citep{Gretton2012}, which is the state-of-the-art in non-parametric statistical testing. \Tab{tab:M2ST_competitors} shows all the compared methods and that only \CTST integrates a graph structure. 

Each non-parametric method requires fixing the regularization constants and the hyperparameters of the kernel function; we focus on Gaussian kernels with width parameter $\sigma$. \LRE-based methods use cross-validation to fix the hyperparameters, 
while from several works addressing this issue for MMD, we compare against the original MMD-MEDIAN version that is based on the median heuristic \citep{Gretton2012}, and the MMD-MAX method proposed in \cite{Sutherland2017} that aims at a score associated with the power of the two-sample test. Details on hyperparameter selection are provided in \Appendix{appendix:experimentsdetailsCM2ST}. %

For the competitors, we follow the traditional MT approach: we first estimate node-level test statistics independently for each node, and then we control for the MCP using a non-parametric resampling test with a maximum statistic \citep{Westfall1992}, which achieves weak FWER control. The node-level test statistics $\{S_v\}_{v \in V}$ coincide with the notion of dissimilarity measured by each method (\fdiv or MMD), and the distribution of $S_{G}=\max_{v \in V} S_v$ under 
the $\Hnull$  is estimated via a permutation test (see \Alg{alg:TS-GRULSIF}). 
We address the non-symmetricity of the \fdiv-based methods same as we did in \Sec{sec:node-level-statistics} for \CTST, by comparing both $p,\q$ and $\q ,p$. Then, given a user-provided threshold rate $\pi^*$, we identify the sets of rejected hypotheses $R_{\phi\text{-div}}=\{v \in V \padded{|} \hat{\pi}_v < \frac{\pi^*}{2}  \ \text{or}  \ \hat{\pi}'_v< \frac{\pi^*}{2} \}$ and $R_{\text{MMD}}=\{v \in V \padded{|} \hat{\pi}_v< \pi^*\}$.

Each instance of the four designed fully synthetic scenarios is generated by first generating a random graph and then by defining the scheme of the occurring change over a subset of the nodes.\\
\noindent$\mySqBullet$~\emph{Synth.Ia\&b} use a \emph{Stochastic Block Model} (SBM) with $4$ clusters, with $25$ nodes each (intra-cluster edge probability: $0.5$; inter-cluster edge probability: $0.01$). Then, a cluster-based scheme sets the same behavior (change of measure or not) for all the nodes of each cluster, $C_1,...,C_4$.\\
\noindent$\mySqBullet$~\emph{Synth.IIa\&b} use a \emph{Grid} graph (GRID) with $100$ nodes forming a $10\!\times\!10$ regular tiling. In this case, an ego-network-based scheme is employed, which picks a node $u$ at random, with probability proportional to its node degree, and then considers that only the nodes in $u$'s $2$-hop ego-network, denoted simply as $C(u)$, shall experience a change of measure.
%
%

\subsection{Synthetic experiments}{\label{sec:exp_setup_M2ST}}

Synthetic scenarios provide by design the set $\Inull^{_\complement}= V \backslash \Inull$, which is the indexes $v$'s where $p_v \neq q_v$, 
hence allow the comparison of the power of the different MT frameworks. The scenarios detailed in \Tab{tab:descr_synthetic_scenarios} are similar to those in \cite{delaconcha2024collaborative} to satisfy the graph smoothness hypothesis (connected nodes have similar behavior), and to pose various challenges to the \LRE addressed by GRULSIF. On the top of each of those scenarios, we build a two-sample test comparing $p_v$ \vs $\q_v$. The two \pdfs may differ in terms of mean, shape, covariance, \etc (see the node-level hypotheses in \Tab{tab:descr_synthetic_scenarios}). Moreover, there can be more than one type of change in the same scenario. %
\begin{table}[t]
\caption{\textbf{List of competitors.} All the methods that are included in our experimental evaluation study for the graph-structured multiple two-sample test problem. l.-r. indicates the method that estimates the non-regularized likelihood-ratio ($\alpha=0$).}{\label{tab:M2ST_competitors}} 
\footnotesize
\centering
\vspace{-2mm}
\makebox[\linewidth][c]{%
\scalebox{.735}{
\begin{tabular}{l c r c l c}
     \toprule
\textbf{Method} & \textbf{Reference} & \textbf{Estimate} & \textbf{Similarity measure}  & \textbf{Graph}\\
\midrule
KLIEP   & \cite{Sugiyama2007} & l.-r. & \KLdiv  & No\\
LSTT   & \cite{Sugiyama2011_LSTST} & l.-r. & \PEdiv  & No\\
RULSIF  & \cite{Yamada2013_RULSIFjournal}   & relative l.-r. & \PEdiv  & No\\
MMD  & \cite{Gretton2012}   & MMD & MMD  & No\\
\cmidruledashed{1-6}\\
POOL    & this work (\Sec{sec:CTST-POOL}) & relative l.-r. & \PEdiv  & No\\
\CTST & this work & relative l.-r. & \PEdiv  & Yes\\
     \bottomrule
\end{tabular}
}
}
\end{table}%
\begin{table}[t]
\vspace{-0.4em}
\caption{\textbf{Synthetic experiments.} The scenarios are defined by the graph structure they employ and the node-level distributions ($p_v$ and $q_v$) generating the data observations at each node. 
\scalebox{.85}{$\bullet$} denotes cases where distributions or their parameters remain unchanged.}
{\label{tab:descr_synthetic_scenarios}}
\newcommand{\algnopar}{}
\footnotesize
\centering
\vspace{-2mm}
\makebox[\linewidth][c]{%
\scalebox{.55}{
\begin{tabular}{c l r c l}
		\cmidrule[0.8pt]{3-5} 
      & & \multicolumn{3}{c}{\textbf{Node-level hypotheses}} \\
		\cmidrule[0.8pt](r{1mm}){1-2}\cmidrule[0.8pt]{3-5}
     \textbf{Experiment} &\textbf{Location} & $p_v$ & \textbf{\vs} & $q_v$ \\
		\cmidrule[0.5pt](r{1mm}){1-2}\cmidrule[0.5pt]{3-5}
		\multicolumn{1}{c}{\multirow{3}{*}{\raisebox{-2pt}{$\stackrel{\mbox{\textbf{Synth.Ib}}}{\stackrel{\mbox{SBM}}{\mbox{\tiny $4$ clusters}}}$}}} & $ v \in C_1$ & N$(\mu\!=\!0, \, \sigma\!=\!1)$  &\vs& $\text{Uniform}(-\sqrt{3}, \, \sqrt{3})$ \\
		& $v \in C_2 \cup C_3$ & N$(\mu\!=\!0, \, \sigma\!=\!1)$  &\vs&$\bullet$ \\
    & $v \in C_4$ & N$(\mu\!=\!0, \, \sigma\!=\!1)$  &\vs  & N$(\mu\!=\!1, \, \sigma\!=\!\bullet)$\\
		\cmidrule[0.5pt](r{1mm}){1-2}\cmidrule[0.5pt]{3-5}
		\multicolumn{1}{c}{\multirow{3}{*}{%
		\raisebox{-2pt}{$\stackrel{\mbox{\textbf{Synth.Ib}}}{\stackrel{\mbox{SBM}}{\mbox{\tiny $4$ clusters}}}$}}} 
		& $v \in C_1\cup C_2$ & N$(\mu\!=\!(0,0)^{\top}, \, \Sigma_{1,2}\!=\!-\frac{4}{5})$ & \vs &$\bullet$\\
    & $v \in C_3$ & N$(\mu\!=\!(0,0)^{\top}, \, \Sigma_{1,2}\!=\!\phantom{-} \frac{4}{5})$  &\vs&  N$(\mu\!=\!\bullet, \, \Sigma_{1,2}\!=\!\,0)$\\
    & $v \in C_4$ & N$(\mu\!=\!(0,0)^{\top}, \, \Sigma_{1,2}\!=\! \,\phantom{-}0)$   &\vs& N$(\mu\!=\!(1,1)^{\top}, \, \Sigma_{1,2}\!=\! \, \bullet)$ \\
		\cmidrule[0.5pt](r{1mm}){1-2}\cmidrule[0.5pt]{3-5}
		\multirow{2}{*}{\raisebox{-2pt}{$\stackrel{\mbox{\textbf{Synth.IIa}}}{\mbox{$\stackrel{\mbox{GRID}}{\mbox{\tiny $10\!\times\!10$}}$}}$}}   & $v \in C(u)$ &  N$(\mu\!=\!\zeros{3}, \, \Sigma_{i,i}\!=\!1, \, \Sigma_{1,2}\!=\!{\textstyle\frac{4}{5}}, \,\Sigma_{3,1}\!=\!0)$ 
		&\vs&	N$(\mu\!=\!\bullet, \, \Sigma_{i,i}\!=\!\bullet, \, \Sigma_{1,2}\!=\!0, \,\Sigma_{3,1}\!=\!\bullet)$  \\
		& $v \notin C(u)$ & N$(\mu\!=\!\zeros{3}, \, \Sigma_{i,i}\!=\!1, \, \Sigma_{1,2}\!=\!{\textstyle\frac{4}{5}}, \,\Sigma_{3,1}\!=\!0)$ &\vs&$\bullet$ \\
		\cmidrule[0.5pt](r{1mm}){1-2}\cmidrule[0.5pt]{3-5}
		\multirow{7}{*}{$\stackrel{\mbox{\textbf{Synth.IIb}}}{\mbox{$\stackrel{\mbox{GRID}}{\mbox{\tiny $10\!\times\!10$}}$}}$} & $v \in C(u)$ & N$(\mu\!=\!(0,0)^{\top}, \, \Sigma \!=\! 10 \id_{2})$ &\vs& Gaussian Mixture $\big(\raisebox{-2pt}{$\stackrel{\mbox{\tiny with equal}}{\mbox{\tiny proportion}}$}\big)$	\\
		&&&& \ \ N$(\mu_1\!=\!(\phantom{-}0,\phantom{-}0)^{\top}, \Sigma\!=\!5 \id_{2})$\\
		&&&& \ \ N$(\mu_2\!=\!(\phantom{-}0,\phantom{-}5)^{\top}, \Sigma\!=\!5 \id_{2})$\\
		&&&& \ \ N$(\mu_3\!=\!(\phantom{-}0,-5)^{\top}, \Sigma\!=\!5 \id_{2})$\\
		&&&& \ \ N$(\mu_4\!=\!(\phantom{-}5,\phantom{-}0)^{\top}, \Sigma\!=\!5 \id_{2})$\\
		&&&& \ \ N$(\mu_5\!=\!(-5,\phantom{-}0)^{\top}, \Sigma\!=\!5 \id_{2})$ \\
		& $v \notin C(u)$ & \ \ N$(\mu\d!=\!(0,0)^{\top}, \, \Sigma \!=\! 10 \id_{2})$ &\vs & $\bullet$\\
		\cmidrule[0.8pt](r{1mm}){1-2}\cmidrule[0.8pt]{3-5}
\end{tabular}
}}
\end{table}
\begin{table}
\caption{\textbf{Results on synthetic scenarios.}
Non-parametric methods applied on multiple two-sample testing over a known graph. 
Keeping the graph fixed, the AFROC and ROC curves were computed over $1000$+$1000$ experiment instances generated over $\Hnull$ 
and $\Halt$ of \Problem{eq:statistical_test}, respectively. Higher AUC values are better.
} 
\label{tab:synthetic_experiments}
\newcommand{\algnopar}{}
\label{tab:experiments}
\footnotesize
\centering
\vspace{-1mm}
\makebox[\linewidth][c]{%
\scalebox{.66}{
\begin{tabular}{c l   c c   c c   c c }
     \cmidrule[0.8pt]{3-8}
     &&  \multicolumn{2}{c}{$n=n'=50$}  & \multicolumn{2}{c}{$n=n'=100$}  &   \multicolumn{2}{c}{$n=n'=250$}   \\
    \toprule
     \multirow{2}{*}{\textbf{Experiment}}  & \multirow{2}{*}{\textbf{\  Method\ \ \ \ \ \ \ \ }}
		& \textbf{AFROC} & \textbf{ROC}  & \textbf{AFROC} & \textbf{ROC}  &  \textbf{AFROC} & \textbf{ROC}  \\
       &  & \textbf{AUC}  & \textbf{AUC}  & \textbf{AUC}  & \textbf{AUC}  & \textbf{AUC}  & \textbf{AUC} \\
		\midrule
    \multirow{7}{*}{\textbf{Synth.Ia}}  & \CTST $\alpha$$=$$0.1$  & \textbf{0.50} & \textbf{0.93} & \textbf{0.66} & \textbf{0.99} & \textbf{0.99} & \textbf{1.00} \\
     & POOL $\alpha$$=$$0.1$  & 0.28 & 0.84 & 0.49 & 0.93 & 0.64 & 0.99 \\
     & RULSIF $\alpha$$=$$0.1$  & 0.18 & 0.88 & 0.47 & 0.76 & 0.76 & \textbf{1.00} \\
     & LSTT \algnopar  &  0.07 & 0.84 & 0.38  & 0.91 &  0.23 & 0.76 \\
      & KLIEP \algnopar & 0.00 & 0.74 & 0.34 & 0.89 & 0.55 & \textbf{1.00} \\
     & MMD-MEDIAN \, & 0.33 & 0.82 & 0.50 & 0.89 & 0.54 & 0.97   \\   
     & MMD-MAX\hspace{1.5em}  & 0.33 & 0.82 & 0.50 & 0.88 & 0.54 & 0.97 \\
    \midrule
 \multirow{7}{*}{\textbf{Synth.Ib}}  & \CTST $\alpha$$=$$0.1$  & \textbf{1.00} & \textbf{1.00} & \textbf{1.00} & \textbf{1.00} & \textbf{1.00} & \textbf{1.00} \\
     & POOL $\alpha$$=$$0.1$  & 0.72 & \textbf{1.00} & 0.99 & \textbf{1.00} & \textbf{1.00} & \textbf{1.00} \\
     & RULSIF $\alpha$$=$$0.1$  & 0.44 & 0.97  & 0.88 & 0.88 & 0.94 & 0.95  \\
     & LSTT \algnopar  & 0.36  & 0.94 & 0.77 & 0.90 & 0.96 & 0.96 \\
      & KLIEP \algnopar  & 0.33 & 0.90 & 0.79 & 0.94 & 0.92 & 0.93 \\
     & MMD-MEDIAN\,  &  0.48 &  0.96 & 0.52 & 0.99 & 0.96 & \textbf{1.00} \\  
     & MMD-MAX\hspace{1.5em}  & 0.48 & 0.96 & 0.52 & 0.99 & 0.96 & \textbf{1.00} \\
      \midrule
       \multirow{7}{*}{\textbf{Synth.IIa}}  & \CTST $\alpha$$=$$0.1$  & \textbf{0.94} & \textbf{1.00} & \textbf{1.00} & \textbf{1.00} & \textbf{1.00} & \textbf{1.00} \\
     & POOL $\alpha$$=$$0.1$  & 0.18 & 0.98 & 0.22 & 0.84 & \textbf{1.00} & \textbf{1.00} \\
     & RULSIF $\alpha$$=$$0.1$  & 0.01 & 0.82 & 0.30 & 0.99 & 0.52 & 0.61 \\
     & LSTT \algnopar  &  0.00 & 0.81 & 0.23 &  0.83 & 0.97 & \textbf{1.00}  \\
      & KLIEP \algnopar  & 0.00 & 0.80 & 0.29 &  0.91 & 0.67 & 0.73 \\
     & MMD-MEDIAN\,  & 0.00  &  0.81 & 0.01 & 0.95 & 0.43 & \textbf{1.00} \\   
     & MMD-MAX\hspace{1.5em} & 0.00 & 0.82 & 0.01 & 0.95 & 0.39 & \textbf{1.00}  \\
    \midrule
       \multirow{7}{*}{\textbf{Synth.IIb}}  & \CTST $\alpha$$=$$0.1$  & \textbf{0.30} &  \textbf{0.92}  &  \textbf{0.65} & \textbf{0.98}  & \textbf{0.98} & \textbf{1.00}  \\
     & POOL $\alpha$$=$$0.1$  &  0.02 & 0.84  &  0.12 & 0.95 & 0.78 & \textbf{1.00} \\
     & RULSIF $\alpha$$=$$0.1$  & 0.01 & 0.80 &  0.06 &  0.92 & 0.75 & \textbf{1.00} \\
     & LSTT \algnopar  &  0.00 & 0.78  &  0.04 & 0.91 & 0.66 & \textbf{1.00} \\
      & KLIEP \algnopar  & 0.00 & 0.79 &  0.03 &  0.85 & 0.63 & \textbf{1.00} \\
     & MMD-MEDIAN\,  &  0.00 & 0.78  & 0.05 & 0.92 & 0.60 & \textbf{1.00} \\ 
     & MMD-MAX\hspace{1.5em} & 0.00 &  0.78 & 0.05 & 0.92 & 0.52 & \textbf{1.00} \\
      \bottomrule
\end{tabular}
}
}
\end{table}

We measure the performance of a MT procedure 
along two 
axes: First, the efficiency of its FWER control, \ie the probability to occure one or more false positives under the $\Hnull$ of \Eq{eq:Hnull}. Second, how informative the estimated node-level \pvalues are, \ie whether the low \pvalues are associated with nodes in $\Inull^{_\complement}$. From a practitioner's perspective, when comparing a complex system across two different time-stamps or experimental conditions, methods that are robust to false positives (avoid asserting a non-existent statistically significant difference) are preferred. Second, we measure how accurately the MT procedure identifies the nodes responsible for an observed deviation. This quality is summarized by the Alternative Free-response Receiver-Operating Characteristic (AFROC) curve \citep{Chakraborty1990}. The detailed estimation we used for the AFROC curves is provided in \Appendix{appendix:experimentsdetailsCM2ST}. 

The AUC of the AFROC curves is reported in \Tab{tab:experiments}. The higher the value of the AUC the better, indicating that a method achieved the required FWER level of $\pi^* = 0.05$ and is still able to identify the nodes in $\Inull^{_\complement}$. The AFROC curves we designed ignore the false positives at nodes $\{v \padded{|} v \in R_{\text{MT}} \setminus \mathbf{I}_0 \}$ . For this reason, we report also the AUC of the ROC curves. The interpretation should take AFROC-AUC as the most important criterion, and ROC-AUC rather as a tiebreaker for approaches with similar AFROC-AUC.

\inlinetitle{Findings}{.}~\Tab{tab:experiments} shows that \CTST 
is a clearly more efficient test 
compared to the rest of the 
methods that disregard 
the geometry of the problem. 
The role of the graph becomes more relevant as the observations are fewer, and when the difference between $p_v$ and $q_v$ is more subtle. This effect is more evident when comparing \CTST to the no-graph variant POOL (\Sec{sec:CTST-POOL})
. An additional advantage of \CTST over POOL is that it is robust and consistent 
when varying the regularization parameter $\alpha$ (see \Appendix{appendix:experimentsdetailsCM2ST}). 

\subsection{Two-sample testing on real seismic data}\label{sec:real-data_CTST}

We use seismic data as a practical example showcasing \CTST's potential in performing spatial statistical analyses. We remark, though, that this should not be interpreted as an attempt to outperform existing state-of-the-art methods in that field. 
Geological hazard monitoring systems comprises of several stations strategically positioned across a territory to monitor ground noise and shaking through a number of sensors. 
When a seismic event occurs, it travels through the earth, and this is captured by the monitoring sensors. Stations closer to the epicenter of a seism 
tend to show higher response to the event, exhibiting faster reactions and more pronounced differences in their pre- and post-event data. In this context, a graph-structured multiple two-sample test can be used for assessing the significance of a seismic event, and for identifying the stations and time periods during which each of them got activated.

\textbf{Data preprocessing:} We analyze two seismic events occurred in New Zealand: Seism\,A is of magnitude $5.5$ in Richter scale, occurred on May 31, 2021
\footnote{\scriptsize The data are publicly available by the GeoNet project \cite{GeoNet1970}: \\\phantom{xxxx.}\url{https://www.geonet.org.nz/earthquake/2021p405872}}; Seism\,B is a weaker seism of magnitude $2.6$, occurred on Oct 2, 2023
\footnote{\scriptsize \url{https://www.geonet.org.nz/earthquake/2023p741652}}. The stations are equipped with strong-motion accelerometers that provide $3$d signals corresponding to the shaking across three perpendicular directions. 
To compare the situation 
before and after an event, 
we analyze the waveforms from $50$ seconds before to $50$ seconds after the event, 
at $100$ Hz frequency. 
The preprocessing details are in \Appendix{appendix:experimentsdetailsCM2ST}. 

\textbf{Graph structure:} We build in two steps a graph representation that accounts for both spatial and temporal similarities between the seismic stations and their signals. The first step is to build an unweighted \emph{spatial graph} $\Gs=(V,E,W)$
 considering as nodes the stations whose all accelerometers have available data at the analyzed time period. %
Edges are drawn from each station to its geographical $3$-nearest-neighbors. Subsequently, we integrate the temporal dimension by building a \emph{multiplex graph} $\Gst$ over $G$. We segment the signal before and after the seismic event 
in $10$ time-windows, each containing the same amount of observations. $\Gst$ indexes the nodes of $\Gs$ by the time-window
, $V \times \mathcal{T}$, where $\mathcal{T}=\{1,...,10\}$. Two nodes in $\Gst$, $(u,t)$ and $(v,t')$, are connected: i) if $t=t'$ and $(u,v) \in E$, \ie they refer to the same time-window and the nodes $u$ and $v$ are connected in the spatial graph $\Gs$, ii) or if $u=v$ and $|t'-t|=1$, \ie
each node $v \in V$ is connected to its `copies' in the two adjacent time-windows.


The data observations are indexed by the node and the time-window they belong, so the \pdfs $\{p_{(v,t)}\}_{(v,t) \in V \times T}$ and $\{\q_{(v,t)}\}_{(v,t) \in V \times T}$.  After the preprocessing, we obtain two  samples for each pair $(v,t)$,
$X_{(v,t)}=\{x_{((v,t),i)}\}_{i=1}^{100} \sim p_{(v,t)}$ and $X'_{(v,t)}=\{x'_{((v,t),i)}\}_{i=1}^{100} \sim \q_{(v,t)}$.
We denote by $t=0$ the beginning of the sample, that is $50$ seconds before the seism. Then, the set $X_{(v,1)}$ refers to the first $5$ seconds of preprocessed observations after $t=0$ and $X'_{(v,1)}$ the $5$ seconds of preprocessed observations after the event. %
Under this configuration, a two-sample test aims to identify the pairs $(v,t)$ where $p_{(v,t)} \neq \q_{(v,t)}$. 
For each method, a figure shows the map of the computed \pvalues, where the nodes $(v,t)$ whose \pvalues is smaller than $0.05$ are highlighted. We only report the largest cluster of $\Cst$ of $\Gst$, containing such nodes.

\inlinetitle{Findings}{.}~\Fig{fig:NEW_ZEALAND_2021_intext} shows a visual of the \CTST result for Seism\,A, while the visualization of the rest of the results are in \Appendix{sec:results_seismic_maps}. All the tested methods 
detect correctly an occurring seismic event and identify the most sensitive nodes as those closer to the epicenter. However, the methods that do not account the graph structure, which here encodes the expected spatial and temporal similarities between stations, lead to results where detections seem not informative. For example, looking at the associated \pvalues, the effect of a seism 
takes longer to fade out even when it ceases to be visible in the signals. Contrary, \CTST recovers most of the nodes closer to the epicenter and follows better the evolution of the seismic event. These findings are compatible with the results found in the synthetic experiments, as the AFROC-AUC and ROC-AUC measures show that \CTST is more robust to false alarms, and that it recovers the nodes of interest with a higher confidence when the assumption of graph smoothness of the likelihood-ratios is satisfied.

\begin{figure}[t!] 
\centering
\begin{minipage}{0.15\textwidth} 
\includegraphics[width=\linewidth,trim=160pt 80pt 140pt 0pt,clip]{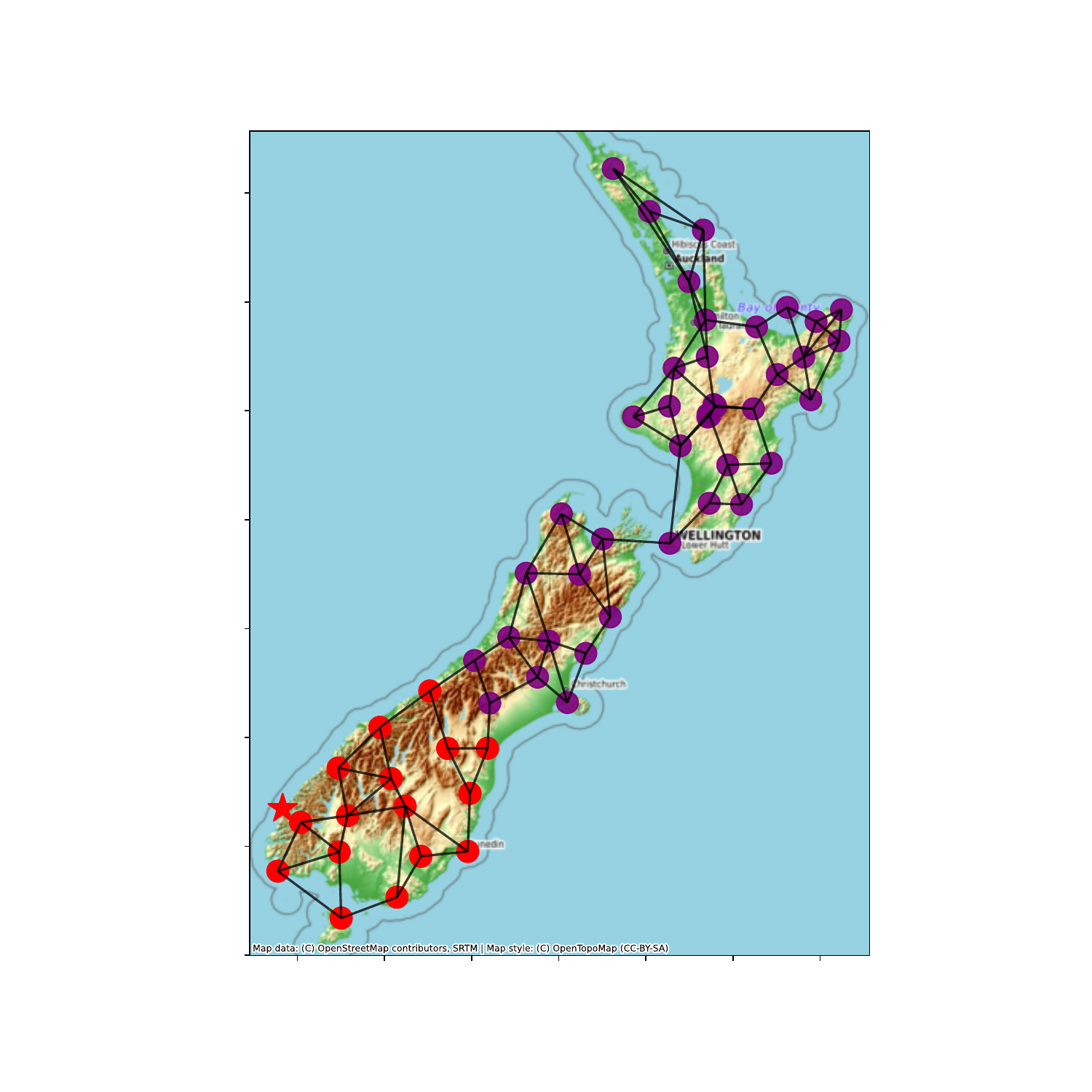}\\
\includegraphics[width=\linewidth,trim=0pt 550pt 285pt 0pt,clip]{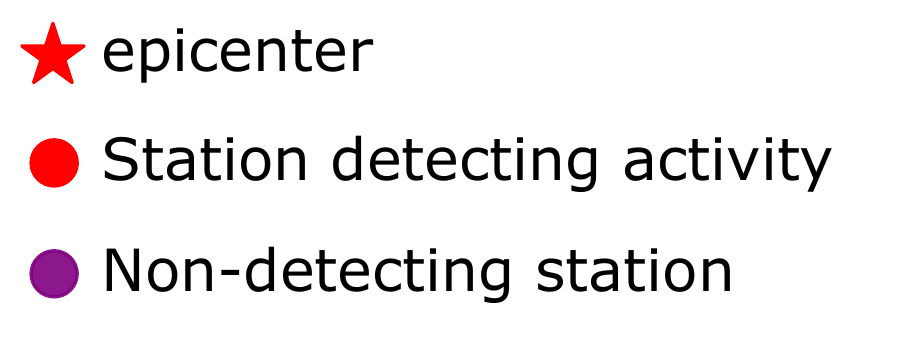}
\end{minipage}%
\begin{minipage}{0.35\textwidth}
	\includegraphics[width=\linewidth,trim=250pt 250pt 215pt 250pt,clip]{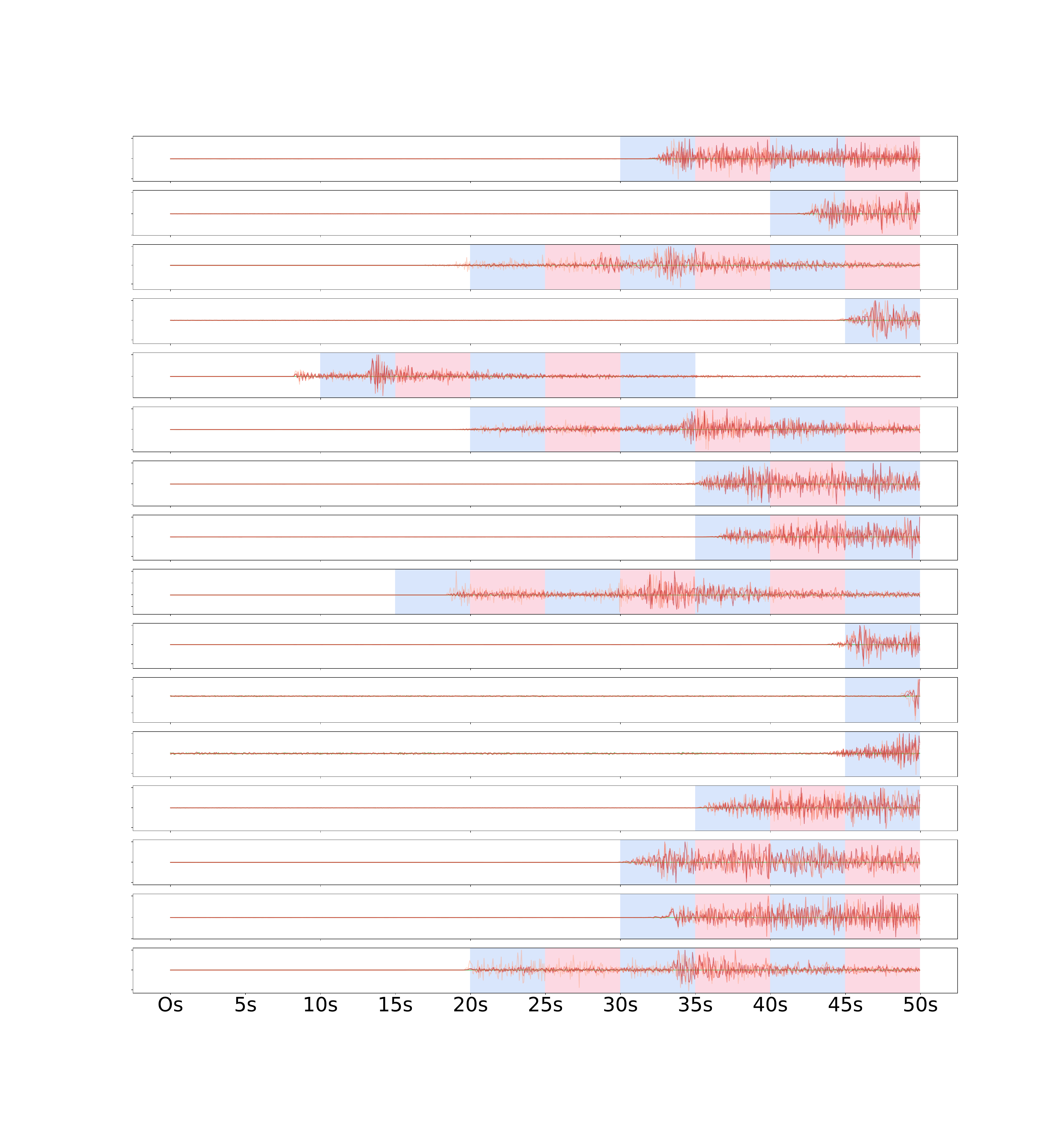}
\end{minipage}
\caption{\textbf{Seism\,A.}~Result of the proposed graph-based \CTST (with $\alpha=0.1$). \textbf{Left:} The location of the stations on the map of New Zealand connected in a $3$NN graph. \textbf{Right:} The post-event signals associated with the stations (in proximity order to the epicenter) detecting activity in at least one time-window (colored).}\label{fig:NEW_ZEALAND_2021_intext}
\end{figure}

\section{Conclusions and further work}

In this paper, we introduced a novel graph-structured non-parametric test designed for multiple two-sample testing over the nodes of a graph. Its appeal is that it integrates advances in collaborative likelihood-ratio estimation to compute jointly node-level test statistics and  identify null hypotheses to be rejected, under a graph smoothness hypothesis. This approach is flexible  
and capable of dealing with complex scenarios in which the data at every node can be multivariate, the nature of the difference between the compared \pdfs is unknown and it is allowed a certain amount of heterogeneity among the tests of the nodes. %
Synthetic and real experiments show that our methods compare favorably against state-of-the-art non-parametric approaches that do not account for the similarity between tests. As future work, it would be interesting to extend the use of this approach to more applications, and design strategies for stricter Type-I error control. 

\bibliographystyle{icml2024}
{\small
\bibliography{references}
}

\newpage
\appendix
\onecolumn

\section{Regarding \CTST and FWER control}{\label{appendix:2sample}}

Bellow, we provide the proof for \Theorem{th:2sample}, which validates that \CTST achieves weak FWER control.

\inlinetitle{Proof of \Theorem{th:2sample}}{.}~%
%
We start with the assumption that the observations of $\mathbf{X}$ come from the joint \pdfunc $\mathbf{p}$, whose marginals are the node \pdfuncs $\{p_v\}_{v \in V}$. Same for those observations of $\mathbf{X}'$ collected from the joint \pdfunc $\mathbf{\q}$ whose marginals are the node \pdfuncs $\{\q_v\}_{v \in V}$. We assume the observations at a specific node $v$,  namely $\{x_{v,i}\}_{\forall v \in V, i = 1,...,n}$ and $\{x'_{v,i}\}_{\forall v \in V, i = 1,...,n'}$, are \iid over the variation of index $i$. Let us define the set of vectors  as  $\mathbf{Z} \mydef \{Z_1,Z_2,...,Z_{\Npre+\Npost}\}$, where $Z_i=X_{:,i}=\{x_{v,i}\}_{v \in V}$ for $i \in \{1,...,\Npre\}$  and $Z_{\Npre+j}=X'_{:,j}=\{x'_{v,j}\}_{v \in V}$ for $j \in \{1,...,\Npost\}$. Then, under $\Hnull$ and the hypothesis of statistical independence, we have that the probability distribution 
$\mathbf{p}^z$ is exchangeable, where exchangeability means that  for any permutation $\tau$ on $\{1,...,\Npre+\Npost\}$, the permuted set of vectors $\mathbf{Z}_{\tau}=\{Z_{\tau(1)},Z_{\tau(2)},...,Z_{\tau(\Npre+\Npost)}\}$ follow the same law $\mathbf{p}^z$.

Given a permutation $\tau$  we assign the first $\Npre$ elements of $\mathbf{Z}_{\tau}$ to the set $\dot{\mathbf{X}}$ and the remaining $\Npost$ to the set $\dot{\mathbf{X}}'$. Denote by $F(\,\cdot\padded{|}\mathbf{X} \cup \mathbf{X}')$ the distribution of the scores $S=\max_{v \in V} \hat{\PE}{}^{\alpha}_v(\dot{\mathbf{X}}_v \Vert \dot{\mathbf{X}}'_v)$ conditioned on $\mathbf{X} \cup \mathbf{X}'$, and let
$\hat{q}(\mathbf{X} \cup \mathbf{X}') = \sup\{s \in \R \padded{|} \ F(s\padded{|}\mathbf{X} \cup \mathbf{X}') \leq 1-\frac{\pi^*}{2}\}$. Then, under $\Hnull$, the echangeability property implies:
\begin{equation}
\begin{aligned}
    \Prob( S > \hat{q}(\mathbf{X} \cup \mathbf{X}') ) \ =  \mathbb{E}_{\mathbf{X} \cup \mathbf{X}'} \Big[\Prob\left(S > \hat{q}(\mathbf{X} \cup \mathbf{X}') \padded{|} \mathbf{X} \cup \mathbf{X}' \right) \Big]  \leq \mathbb{E}_{\mathbf{X}\cup \mathbf{X}'} \Big[1- F(\hat{q}(\mathbf{X}\cup \mathbf{X}') \padded{|}\mathbf{X} \cup \mathbf{X}') \Big] 
    \leq \frac{\pi^*}{2}.
\end{aligned}
\end{equation}
In a similar manner, we can verify that for $S'= \max_{v \in V} \hat{\PE}{}^{\alpha}_v(\dot{\mathbf{X}}'_v \Vert \dot{\mathbf{X}}_v)$:
\begin{equation}
\Prob( S' > \hat{q}'(\mathbf{X} \cup \mathbf{X}')) \leq 
 \frac{\pi^*}{2}.
\end{equation}%
By putting together both inequalities, we can conclude: 
\begin{equation}
 \Prob( S > \hat{q}(\mathbf{X} \cup \mathbf{X}') \ \ \text{or} \ \ S' > \hat{q}'(\mathbf{X} \cup \mathbf{X}') ) \leq  \Prob\left( S > \hat{q}(\mathbf{X} \cup \mathbf{X}') \right) + \Prob\left(S' > \hat{q}'(\mathbf{X} \cup \mathbf{X}') \right) \leq 
 \pi^*.
\end{equation}%
And weak control over FWER comes from:
\begin{equation}
\begin{aligned}
    \FWER(R_{\text{CMT}}) & = \Prob( \{ \exists v: S_v >  \hat{q}(\mathbf{X} \cup \mathbf{X}')  \ \text{or} \ S'_v >  \hat{q}'(\mathbf{X} \cup \mathbf{X}')  \} | \Hnull) \\  & \leq
    \Prob( \{ \exists v: S_v >  \hat{q}(\mathbf{X} \cup \mathbf{X}')  \}| \Hnull) + \Prob( \{ \exists v: S'_v >  \hat{q}'(\mathbf{X} \cup \mathbf{X}') \} |\Hnull) \\ &  =  \Prob( S > \hat{q}(\mathbf{X} \cup \mathbf{X}')\padded{|} \Hnull) 
    + \Prob( S' > \hat{q}'(\mathbf{X} \cup \mathbf{X}') \padded{|} \Hnull) \leq \pi^*.
\end{aligned}
\end{equation}%
\hfill$\blacksquare$

\section{Further details about the experiments}{\label{appendix:experimentsdetailsCM2ST}}

In this section, we give more details on the implementation of the experimental setting described in the main text. Mainly: 
\begin{enumerate}[topsep=0em,itemsep=0.1em]
    \item More details on how the hyperparameters of GRULSIF and the other methods were chosen. 
    \item Elements to complement the results on the synthetic scenarios. This includes the way the AFROC and the ROC curves were estimated, and a detailed discussion about on the role of the regularization parameter $\alpha$ used by \CTST and POOL. 
    \item Further details on the real-world example, including the preprocessing pipeline and the figures comparing the different multiple hypothesis testing settings.
\end{enumerate}

\subsection{Details regarding hyperparameters selection}


For  RULSIF and ULSIF algorithms, we follow \citep{Sugiyama2011} and \citep{Yamada2011}, and the hyperapemeters are selected independently for each of the nodes. We run a leave-one-out cross-validation procedure over the parameter associated with the Gaussian kernel and the penalization term $\gamma$. The parameter $\sigma$ is selected from the grid $\{0.6 \sigma_{\text{median}},0.8\sigma_{\text{median}},1\sigma_{\text{median}},1.2\sigma_{\text{median}},1.4\sigma_{\text{median}} \}$  where $\sigma_{\text{median}}$ is the parameter $\sigma$ found via the median heuristic over the observations in $X'_v$. On the other hand, the penalization parameter $\gamma$ is optimized from the grid $\{1e^{-5},1e^{-3},0.1,10\}$.  The procedure for KLIEP is similar, but we use instead a $5$-fold cross-validation procedure.

For MMD median and MMD max, we identify the hyperparameters independently for each of the nodes, we follow the guidelines given in \citep{Gretton2012,Sutherland2017}, respectively. 

Finally, for \CTST and the POOL algorithms, we apply  $5$-fold cross-validation to select the hyperparameters $\sigma$, $\gamma$, and $\lambda$ using the implementation of \cite{delaconcha2024collaborative}. Since the POOL approach ignores the graph structure, we fix $\lambda=1$, and the penalization term related with the norm of each functional $f_v$ will depend only on the parameter $\gamma$. In order to select the width $\sigma$ for the Gaussian kernel, we first compute $\{\sigma_{v}\}_{v \in V}$ for each node via the median heuristic applied to the observations of $X_v$ (such quantities are available when generating the dictionary), and we define $\sigma_{\text{min}}= \argmin \{\sigma_{v}\}_{v \in V}$, $\sigma_{\text{median}}= \text{median}\{\sigma_{v}\}_{v \in V}$ and $\sigma_{\text{max}}= \argmax \{\sigma_{v}\}_{v \in V}$; we then chose the final parameter from the set $\{\sigma_{\text{min}},\frac{1}{2}(\sigma_{\text{min}}+\sigma_{\text{median}}),\sigma_{\text{median}},\frac{1}{2}(\sigma_{\text{max}}+\sigma_{\text{median}}),\sigma_{\text{max}}\}$. $\gamma$ is selected from the set $\{1e^{-5},1e^{-3},0.1,1\}$. Finally, we define the average node degree $\bar{d}$, 
and we identify the optimal $\lambda^*$ from the set $\{1e^{-3}\cdot\frac{1}{\bar{d}},1e^{-2}\cdot\frac{1}{\bar{d}},0.1\cdot\frac{1}{\bar{d}},1\cdot\frac{1}{\bar{d}},10\cdot\frac{1}{\bar{d}}\}$. 

\subsection{Details regarding synthetic scenarios}

\subsubsection{AFROC and ROC curves}

The \emph{Alternative Free-response Receiver Operating Characteristic} (AFROC) curve is an important tool in the context of multiple hypothesis testing, especially in fields where the practitioner seeks a decision to a global problem while requiring correct localization for true positive events \cite{Chakraborty1990}. In our context, AFROC allow us to quantify to which extent the compared methods achieve \emph{Family-wise False Positive Rate} (FWER) control under the null hypothesis that all nodes $p_v=\q_v$ (see \Sec{sec:p_values_estimation} and the $\Hnull$ in \Eq{eq:sample_per_nodes}), while still being sensitive enough to identify those nodes where $p_v \neq \q_v$ ($\Halt$). 
For each of the synthetic experiments described in \Tab{tab:experiments}, the given input graph $G$ according to the scenario being studied is kept fixed (see \Tab{tab:descr_synthetic_scenarios}), and then the axis of the AFROC curves for the experiments are estimated as follows: 
\begin{enumerate}[topsep=0em,itemsep=0.1em]
\item Generate $1000$ synthetic experiment instances, where for all nodes $p_v=q_v$ and the graph is fixed (Null-instances).
\item Generate $1000$ synthetic experiment instances that satisfy the associated schema (\Tab{tab:descr_synthetic_scenarios}-\ref{tab:experiments}) (Alternative- instances). 
\item For each of the Null-instances and Alternative-instances compute the node-level tests statistics associated to the MTST method used. We refer to the output of this step as processed-Null-instances and processed-Alternative-instances. 
\item Threshold the processed-Null-instances and processed-Alternative-instances at the full range of possible threshold values $thd$ (bigger than $0$ value for the methods being tested), and compute the FWER and the true positive rate (TPR): 
\begin{itemize}[itemsep=0.1em]
\item \textbf{FWER (x-axis)} For each threshold level, compute the fraction of processed-Null-instances where there was a least one node whose value was bigger than the fixed $thd$. 

\item \textbf{TPR (y-axis)} For each threshold value, for each of the Alternative-instances compute the fraction of nodes where $p_v \neq q_v$ whose associated test statistic was bigger than $thd$. The reported TPR is the average TPR estimated over all the Alternative-instances.
\end{itemize}
\item Finally, we compute the AUC from the resulting curve limited to values of FWER in $[0.00, 0.05]$, which are the values of interest for a test of significance level $0.05$.
\end{enumerate}
The higher the value of the AUC of the AFROC curve, the more efficient the analyzed algorithm. We divide the result by $0.05$ in order to scale the result and keep the same interpretation as for a classical AUC result. 

Notice that AFROC ignores the nodes in the Alternative-instances where $p_v = q_v$ whose associated \pvalue is small (false rejections), thus the Null hypothesis is incorrectly rejected. To quantify how well a method differentiates the nodes that should be rejected, we estimate as well the usual ROC curves from the processed-Alternative instances and compute the associated AUC. The interpretation of the results should take AFROC-AUC as the most important criterion, and ROC-AUC rather as a tiebreaker for approaches with similar AFROC-AUC.

Finally, recall that, in a given study, the graph is not a random variable but it is rather a given fixed element, which justifies why we do not vary this element in the analysis above.

\subsubsection{The role of $\alpha$}

In this section, we discuss the role of parameter $\alpha$ in the graph-structured MTST problem. We retain the same set of experiments described in \Sec{sec:experiments} to compare the role of $\alpha$ in \CTST that integrates the graph structure, as well as in the POOL variant that does not consider the graph. The comparison relies on the AFROC-AUC and ROC-AUC measures. Results are summarized in \Tab{tab:experiments_alphas_CTST}.

As explained in the main text, the role of $\alpha$ is to upper-bound the relative likelihood-ratios $r_v^{\alpha}$, thereby preventing convergence issues in terms of sample size and numerical instability. In previous works, such as those in \cite{Yamada2011,delaconcha2023online,delaconcha2024collaborative}, the role of $\alpha$ has been made explicit as a component that controls the speed of convergence of the \LRE based on the Pearson's $\chi^2$-divergence. The conclusion drawn by those papers is consistent: a higher value of $\alpha$ will lead to a faster convergence rate.  Nevertheless, a high level of $\alpha$ will hinder to quantify the difference between $p_v$ and $\q_v$ via the quantity $\PE^{\alpha}(p_v \Vert \q_v)$. In the limit case, that is $\alpha=1$, $\PE^{\alpha}(p_v \Vert \q_v)=0$, meaning these measures fail to differentiate  $p_v$ and $\q_v$ regardless of the form those \pdfs. Therefore, there exists a trade-off: the stability associated with high values of $\alpha$ versus the sensibility of $\PE^{\alpha}(p_v \Vert \q_v)$ in distinguishing between $p_v$ and $\q_v$. This trade-off becomes more relevant when $\PE^{\alpha}(p_v \Vert \q_v)$ is to be used as a  test statistic to carry out hypothesis testing and detection tasks. 

\inlinetitle{Findings}{.}~\Tab{tab:experiments_alphas_CTST} compares \CTST and POOL with $\alpha \in \{0.01, 0.1, 0.5\}$. The first notable observation is that \CTST outperforms consistently POOL regardless of the value of $\alpha$ being used. This finding highlights the predominant role of the graph component over that of $\alpha$, particularly when $\alpha$ is set in a range of meaningful values. The second observation is that POOL's performance appears more sensitive to the values of $\alpha$, it shows lower stability, especially when there are fewer observations. In contrast, \CTST is more robust with respect to this parameter. This can be attributed to the graph-based regularization term that enforces the 
relative likelihood-ratios estimates to be close in the \RKHS, which translates to point-wise similarity as well (see \Eq{eq:similarity_ratios}).

\Tab{tab:experiments_alphas_CTST} does not provide a clear guideline for choosing the optimal parameter $\alpha$ for CM2ST. In the main text, we fix $\alpha=0.1$ because it yielded the 
best results for POOL, and we generally recommend using a value of $\alpha < 0.5$ when deploying \CTST. 

\begin{table*}
\caption{\textbf{Results on synthetic scenarios with variable regularization parameter $\alpha$.} Non-parametric methods applied on multiple two-sample testing over a known graph. Keeping the graph fixed, the AFROC and ROC curves were computed over $1000$+$1000$ experiment instances generated over $\Hnull$ and $\Halt$ of \Problem{eq:statistical_test}, respectively. Higher AUC values are better.
} 
\label{tab:synthetic_experiments_alphas_CTST}
\newcommand{\algnopar}{}
\label{tab:experiments_alphas_CTST}
\footnotesize
\centering
\vspace{-2mm}
\makebox[\linewidth][c]{%
\scalebox{.85}{
\begin{tabular}{c l   c c   c c   c c }
     \cmidrule[0.8pt]{3-8}
      &  &  \multicolumn{2}{c}{$n=n'=50$}  & \multicolumn{2}{c}{$n=n'=100$}  &   \multicolumn{2}{c}{$n=n'=250$}   \\
    \toprule
     \multirow{2}{*}{\textbf{Experiment}}  & \multirow{2}{*}{\textbf{\  Method\ \ \ \ \ \ \ \ }}  & \textbf{AFROC} & \textbf{ROC}  & \textbf{AFROC} & \textbf{ROC}  &  \textbf{AFROC} & \textbf{ROC}  \\
    &  & \textbf{AUC}  & \textbf{AUC}  & \textbf{AUC}  & \textbf{AUC}  & \textbf{AUC}  & \textbf{AUC} \\
		\midrule
    \multirow{6}{*}{\textbf{Synth.Ia}}  & \CTST\; $\alpha$$=$$0.01$  & \textbf{0.57} &  0.90 & \textbf{0.76} &  0.96 & \textbf{1.00} & \textbf{1.00}  \\
     & POOL $\alpha$$=$$0.01$  &  0.13 & 0.81 & 0.24   &  0.94  & 0.87 & \textbf{1.00} \\
     & \CTST\; $\alpha$$=$$0.1$  & 0.50 & \textbf{0.93} & 0.66 & \textbf{0.99} & 0.99 & \textbf{1.00} \\
     & POOL $\alpha$$=$$0.1$  & 0.28 & 0.84 & 0.49 & 0.93 & 0.64 & 0.99 \\
      & \CTST\; $\alpha$$=$$0.5$  & \textbf{0.57} & 0.92 & 0.72 & 0.97 & 0.98 & \textbf{1.00} \\
     & POOL $\alpha$$=$$0.5$  & 0.27  & 0.87 & 0.53 & 0.96 & 0.88 & \textbf{1.00}  \\
    \midrule
 \multirow{6}{*}{\textbf{Synth.Ib}}  & \CTST\; $\alpha$$=$$0.01$  & 0.99 & \textbf{1.00} & \textbf{1.00} & \textbf{1.00} & \textbf{1.00} & \textbf{1.00} \\
     & POOL $\alpha$$=$$0.01$  & 0.53  & \textbf{1.00} & 0.91 & \textbf{1.00} & \textbf{1.00} & \textbf{1.00} \\
    & \CTST\; $\alpha$$=$$0.1$  & \textbf{1.00} & \textbf{1.00} & \textbf{1.00} & \textbf{1.00} & \textbf{1.00} & \textbf{1.00} \\
     & POOL $\alpha$$=$$0.1$  & 0.72 & \textbf{1.00} & 0.99 & \textbf{1.00} & \textbf{1.00} & \textbf{1.00} \\
     & \CTST\; $\alpha$$=$$0.5$  & 0.99 & \textbf{1.00} &  \textbf{1.00} & \textbf{1.00} & \textbf{1.00} & \textbf{1.00} \\
     & POOL $\alpha$$=$$0.5$  & 0.41 & 0.99 & 0.85 & \textbf{1.00} & \textbf{1.00} & \textbf{1.00} \\
      \midrule
       \multirow{6}{*}{\textbf{Synth.IIa}}  & \CTST\; $\alpha$$=$$0.01$  & \textbf{0.99} & \textbf{1.00} & \textbf{1.00} & \textbf{1.00} & \textbf{1.00} & \textbf{1.00} \\
     & POOL $\alpha$$=$$0.01$  & 0.14  & 0.96 &  0.72 & \textbf{1.00} & \textbf{1.00} & \textbf{1.00} \\
    & \CTST\; $\alpha$$=$$0.1$  & 0.94 & \textbf{1.00} & \textbf{1.00} & \textbf{1.00} & \textbf{1.00} & \textbf{1.00} \\
     & POOL $\alpha$$=$$0.1$  & 0.18 & 0.98 & 0.84 &  0.98 & \textbf{1.00} & \textbf{1.00} \\
      & \CTST\; $\alpha$$=$$0.5$  & 0.98  & \textbf{1.00} & \textbf{1.00} & \textbf{1.00} & \textbf{1.00} & \textbf{1.00} \\
     & POOL $\alpha$$=$$0.5$  & 0.04 & 0.89 & 0.43 &  0.99 & \textbf{1.00} &  \textbf{1.00} \\
    \midrule
       \multirow{6}{*}{\textbf{Synth.IIb}}  & \CTST\;$\alpha$$=$$0.01$  & 0.18 &  \textbf{0.94}  &  0.42  & \textbf{0.99} & \textbf{1.00} & \textbf{1.00}  \\
     & POOL $\alpha$$=$$0.01$  & 0.02 & 0.83 & 0.00  & 0.73 & 0.43 & 0.99 \\
     & \CTST\; $\alpha$$=$$0.1$  & \textbf{0.30} &  0.92  &  \textbf{0.65} & 0.98  & 0.98 & \textbf{1.00}  \\
     & POOL $\alpha$$=$$0.1$  &  0.02 & 0.84  &  0.12 & 0.95 & 0.78 & \textbf{1.00} \\
     & \CTST\; $\alpha$$=$$0.5$  & 0.06 & 0.89 & 0.52  & \textbf{0.99} & 0.97 &  \textbf{1.00} \\
     & POOL $\alpha$$=$$0.5$  &  0.04 & 0.84 &  0.07 & 0.91 & 0.60 & 0.99 \\
      \bottomrule
\end{tabular}
}
}
\end{table*}

\subsubsection{Further details regarding the application of two-sample testing on real seismic data}\label{sec:results_seismic_maps}

In this section, we provide more details on the preprocessing pipeline to derive the results described in \Sec{sec:real-data_CTST}, and the additional figures showing the performance of the different methods. 

\inlinetitle{Data preprocessing}{.}~As mentioned in the main text, we analyze waveforms that correspond to two seismic events that occurred in New Zealand. Seism\,A is of magnitude $5.5$, while Seism\,B is magnitude 2.6. 
These seismic events are part of the publicly available dataset provided by GeoNet. We used the Python package ObsPy to access the data \cite{ObsPy2010}. 

To study the evolution of seismic activity associated to these events, we retrieve the waveforms from $50$ seconds before to $50$ seconds after the event. These waveforms correspond to the measurements provided by strong-motion accelerometers that monitor shaking in three perpendicular directions. Therefore, here the input space is $\mathcal{X} \subseteq \R^3$. In each of the scenarios, we limit our attention to stations that had recorded observations for all the three directions during all the analyzed time period. 

There are three main characteristic that are required to implement \CTST in practice: %
\begin{enumerate}[topsep=0em,itemsep=0.1em]
    \item The relative likelihood-ratios $\{r^{\alpha}_v\}_{v \in V}$ are expected to be approximated by the same \RKHS.
    \item The FWER control of \CTST (see \Theorem{th:2sample}) requires that the observations $\mathbf{X}=\{\setprev\}_{v\in V} = \{x_{v,1},...,x_{v,\Npre}\}_{v\in V}$
are \iid for each node $v$, and the same for and $\mathbf{X}'= \{ \setpostv \}_{v\in V} =\{x'_{v,1},...,x'_{v,\Npost}\}_{v\in V}$.
\item The vector-valued function $\mathbf{r}^{\alpha}=(r_1,...,r_{\Gdims})$ is expected to be smooth with respect to the graph $G$, \ie $\norm{r_u-r_v}_{\Hilbert} < \epsilon$ for connected nodes. 
\end{enumerate}
The preprocessing aims to transform and prepare the data so they satisfy these conditions.

We follow the preprocessing pipeline described in \cite{Chen2021} with the toolbox for Seismology ObsPy. The preprocessing is performed independently for each station and independently for each direction. We start by steps that are considered to be standard in seismology: we remove the linear trend and we apply a $2$-$16$ bandpass filter. To reduce the temporal dependency, we compute a root mean square amplitude envelope, then we fit an autoregressive model of order $1$, and we keep the residuals from this model. The output is standardized so that it has zero mean and unit variance. To make the data comparable between stations, we divide the output by its maximum value. 

Our objective is to provide a visualization that captures the evolution of the seismic event using the measurements available at each station. To this end, we use a graph-structured MTST to identify the specific moments and locations (stations) where the seismic activity appeared to be statistically significant. In this context, $v \in V$ denotes that station $v$ belongs to the set of stations $V$. To define the statistical test, we need to identify the samples $\mathbf{X}$ and $\mathbf{X}'$ (see \Eq{eq:sample_per_nodes}) that should be compared across the spatial and temporal dimensions. We denote by $\tau$ the time-stamp of the seismic event, then we consider the preprocessed observations in two time frames: 
$[\tau - 50,...,\tau)$ and $[\tau ,...,\tau+50)$, \ie from $50$ seconds before $\tau$ to $50$ seconds after $\tau$. These periods are segmented into $10$ time-windows ($\mathcal{T}=\{1,...,10\}$, each of $5$ seconds duration) made of $100$ prepossessed observations in each of them. According to our notation, $X_{v,1}$ is the first $100$ observations at station $v$ after $\tau-50$, while $X'_{v,1}$ denotes the first $100$ observations post-event ($\tau$). Following the same logic, $X_{v,2}$ has the observations after $\tau-45$ at station $v$, while $X'_{v,1}$ denotes the first $100$ observations after $\tau+5$. This segmentation yields two samples for each location-time pair $(v,t) \in V \times \mathcal{T}$, $X_{(v,t)}=\{x_{((v,t),i)}\}_{i=1}^{100} \sim p_{(v,t)}$ and $X'_{(v,t)}=\{x'_{((v,t),i)}\}_{i=1}^{100} \sim \q_{(v,t)}$. Then, the MTST compares the \pdfs $\{p_{(v,t)}\}_{(v,t) \in V \times \mathcal{T}}$ and $\{\q_{(v,t)}\}_{(v,t) \in V \times \mathcal{T}}$. Alternatives can be implemented for defining different observations to consider from $\{p_{(v,t)}\}_{(v,t) \in V \times \mathcal{T}}$ to be used to compare with the post-event alternative.

The sets $\mathbf{X}=\{X_{(v,t)}\}_{v \in V, t \in \mathcal{T}}$ and $\mathbf{X}'=\{X'_{(v,t)}\}_{v \in V, t \in \mathcal{T}}$ represent the observations available at the graph $\Gst$ whose nodes represent a position in space and in time. As in the general graph-structure hypothesis testing problem, $\Gst$ encodes the expected similarity between the results of the test. To encode the fact that close stations are expected to affect each other (recall  the \emph{first law of Geography} from \Sec{sec:introduction}), we generate an unweighted spatial graph $\Gs=(V,E,W)$ where the nodes represent the geographical positions of the seismic stations and the edges are computed in order to form a $3$-nearest neighbors graph. To account for the temporal component similarity expected from the propagation of the seismic waves through the earth, we build an unweighted multiplex network $\Gst=(V_{T},E_{T},W_{T})$ on top of $\Gs$. The set of nodes is then the pair $(v,t) \in V_{T}:=V \times \mathcal{T}$, where $V$ denotes the set of nodes of $\Gs$. 
Two nodes in $\Gst$, $(u,t)$ and $(v,t')$, are connected: i) if $t=t'$ and $(u,v) \in E$, \ie they refer to the same time-window and the nodes $u$ and $v$ are connected in the spatial graph $\Gs$, ii) or if $u=v$ and $|t'-t|=1$, \ie
each node $v \in V$ is connected to its `copies' in the two adjacent time-windows.

The implementation details of the statistical methods being compared are the same as in the synthetic scenarios, which include the hyperparameters selection related to the estimation of the non-parametric test statistics and the way the permutation test is run.

\inlinetitle{Findings}{.}~\Fig{fig:NEW_ZEALAND_2021_1}-\ref{fig:NEW_ZEALAND_2021_4} and \Fig{fig:NEW_ZEALAND_2023_1}-\ref{fig:NEW_ZEALAND_2023_5} illustrate the output of the graph-structured MTST applied to the waveforms related to each for the two seismic events, Seism\,A and Seism\,B. The figures highlight the biggest connected component $\Cst$ made of pairs $(v,t) \in V_{T}$ that were identified as statistically significant by the method being used. In this application, we called a pair to be statistically significant if its \pvalue is smaller than $0.05$. 

We try to 
show in the figures both dimensions of the test. 
The graph on the left highlights in red the stations $v \in V$ which were elements of the biggest connected component $C_{T}$ for at least one time-window, That is, there exist $t \in \mathcal{T}=\{1,...,10\}$ such that $(v,t) \in  C_{T}$. The epicenter is marked by a red star. The time-series at the left show both preprocessed data samples $X_{(v,t)}=\{x_{((v,t),i)}\}_{i=1}^{100} \sim p_{(v,t)}$ (green time-series) and $X'_{(v,t)}=\{x'_{((v,t),i)}\}_{i=1}^{100} \sim \q_{(v,t)}$ (red time-series) for the highlighted stations. The periods that were considered statistically significant are delineated by blue/pink colors (we use two colors to differentiate adjacent time-windows where the test rejected the $\Hnullv$ hypothesis).  

The first thing to notice is that all methods identified the stations that were closer to the epicenter as locations where the there was statistically significant evidence of a change. 
Algorithms that neglect the graph component tend to identify a larger number of time-windows. Upon closer inspection, we can see that many of the identified time-windows appear to be false positives, lacking in global relevance or consistency with the expected evolution of the sesmic activity. %
Intuitively, for a short time-period around an even (here we analyze 100 seconds overall), we expect the event to alter the behavior of the measurements during consecutive time-windows, and this effect will vanish with time. This pattern is not evident in methods that disregard the spatial and temporal similarity. In contrast, \CTST identifies correctly nodes close to the epicenter, and captures the evolution of the seismic activity in a more consistent way. The results are consistent with those obtained from synthetic experiments, where \CTST demonstrated superior performance in terms of the AFROC-AUC. This performance also indicates the effective weak FWER control and higher sensitivity in pinpointing the nodes where $p_v \neq \q_v$.

From the practitioners' perspective, MT is usually an initial exploratory tool, where tests identified as statistically significant are further inspected with further analyzis. In this sense, false positives may translate to a high cost, since they may lead to the allocation of resources towards the wrong direction. 
Thus, the accuracy of identifying nodes where $p_v \neq q_v$ is not just a statistical concern, but also a practical one, directly impacting the efficiency and effectiveness of subsequent research efforts.

\begin{figure} 
\caption{\textbf{Seism\,A 
in New Zealand (1 of 4).}}\label{fig:NEW_ZEALAND_2021_1}
\vspace{1em}
\centering
\vspace{-3mm}
\begin{minipage}{0.25\textwidth} 
\centering
\textbf{\footnotesize \CTST $\alpha=0.1$}\\
\vspace{0.5em}
\includegraphics[width=\linewidth,trim=160pt 80pt 140pt 70pt,clip]{figures/Grulsif_New_Zealand2021p405872_network.pdf}\\
\includegraphics[width=\linewidth,trim=0pt 550pt 285pt 0pt,clip]{figures/seisms_legend.pdf}
\end{minipage}
\hspace{1em}
\begin{minipage}{0.50\textwidth}
  \includegraphics[width=\linewidth,trim=250pt 150pt 210pt 230pt,clip]{figures/Grulsif_New_Zealand2021p405872_waveforms.pdf}
\end{minipage}\\
\vspace{2em}
\begin{minipage}{0.25\textwidth} 
\centering
\textbf{\footnotesize POOL $\alpha=0.1$}\\
\vspace{0.5em}
\includegraphics[width=\linewidth,trim=160pt 80pt 140pt 70pt,clip]{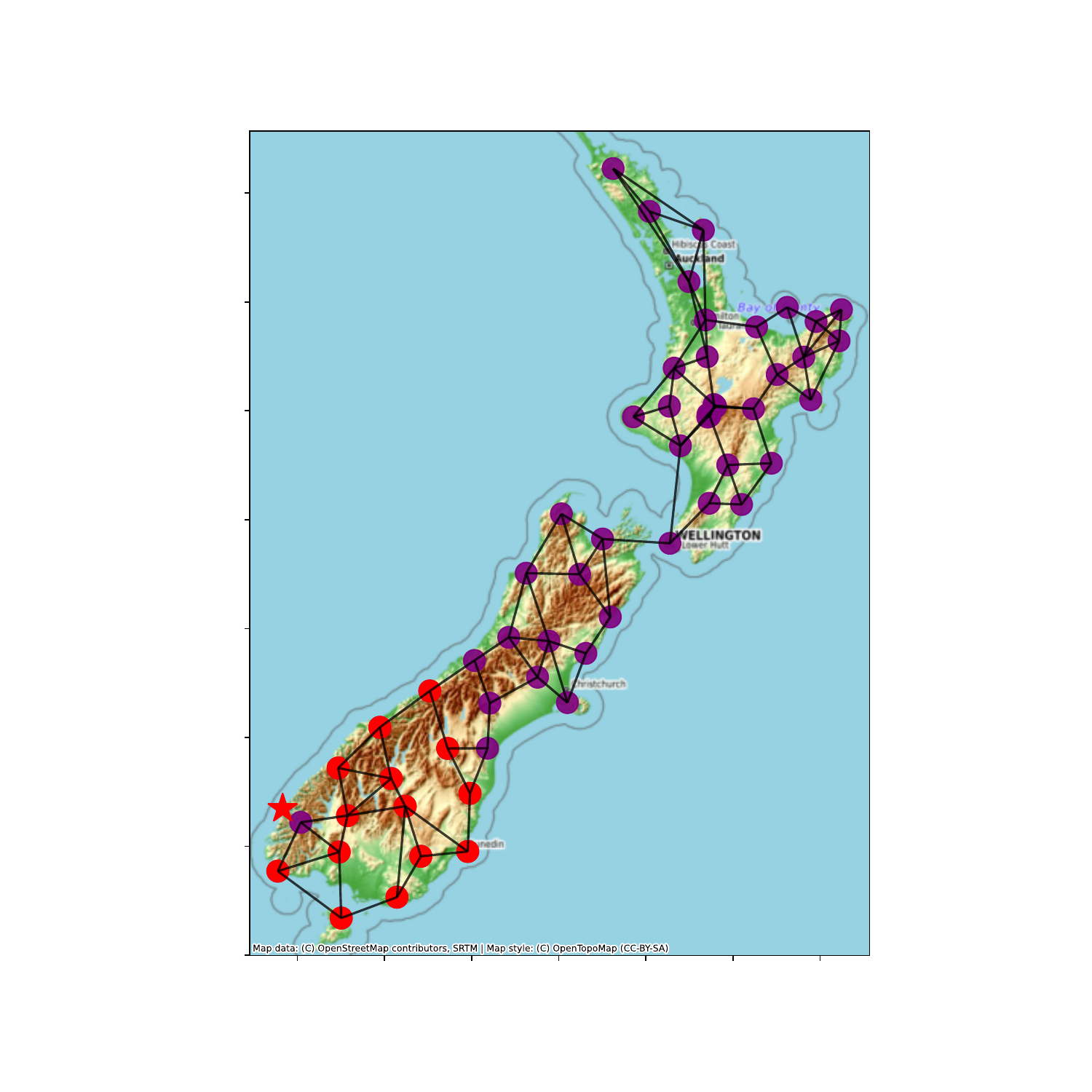}\\
\includegraphics[width=\linewidth,trim=0pt 550pt 285pt 0pt,clip]{figures/seisms_legend.pdf}
\end{minipage}
\hspace{1em}
\begin{minipage}{0.50\textwidth}
\includegraphics[width=\linewidth,trim=250pt 150pt 210pt 230pt,clip]{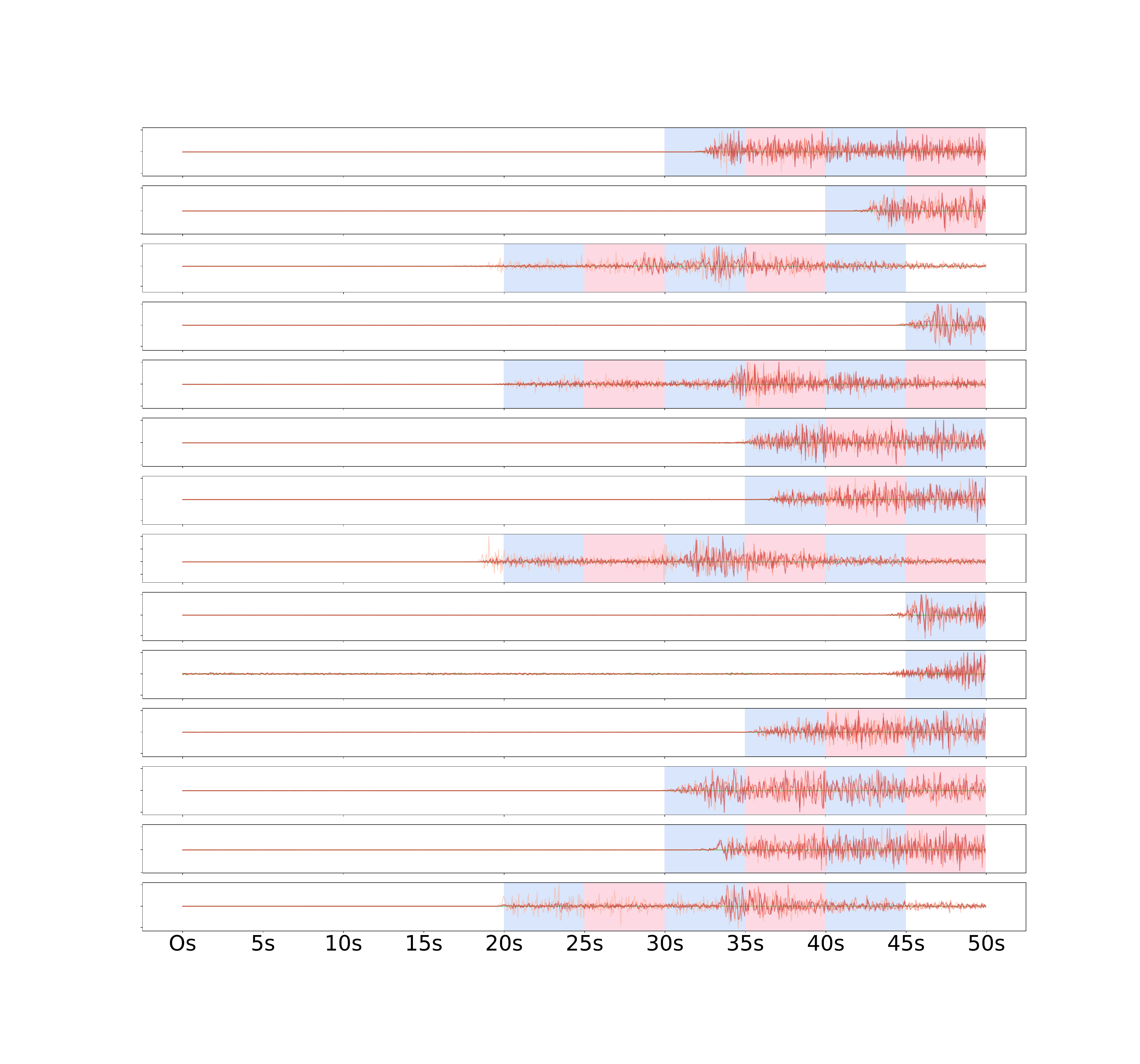}
\end{minipage}
\end{figure}

\begin{figure} 
\caption{\textbf{Seism\,A 
in New Zealand  (2 of 4).}}\label{fig:NEW_ZEALAND_2021_2}
\vspace{1em}
\centering
\vspace{-3mm}
\begin{minipage}{0.25\textwidth} 
\centering
\textbf{\footnotesize RULSIF $\alpha=0.1$}\\
\vspace{0.5em}
\includegraphics[width=\linewidth,trim=160pt 80pt 140pt 70pt,clip]{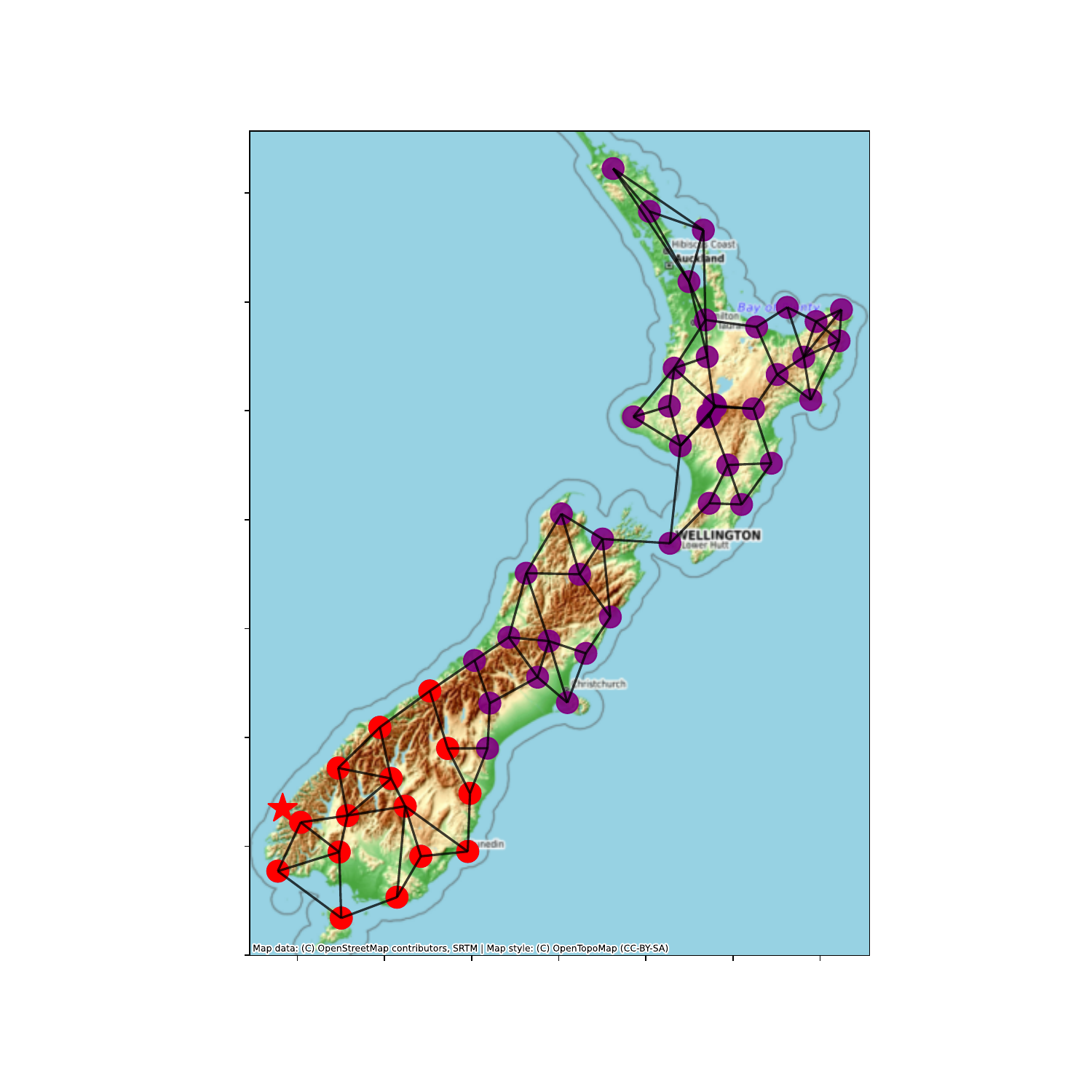}\\
\includegraphics[width=\linewidth,trim=0pt 550pt 285pt 0pt,clip]{figures/seisms_legend.pdf}
\end{minipage}
\hspace{1em}
\begin{minipage}{0.50\textwidth}
  \includegraphics[width=\linewidth,trim=250pt 150pt 210pt 230pt,clip]{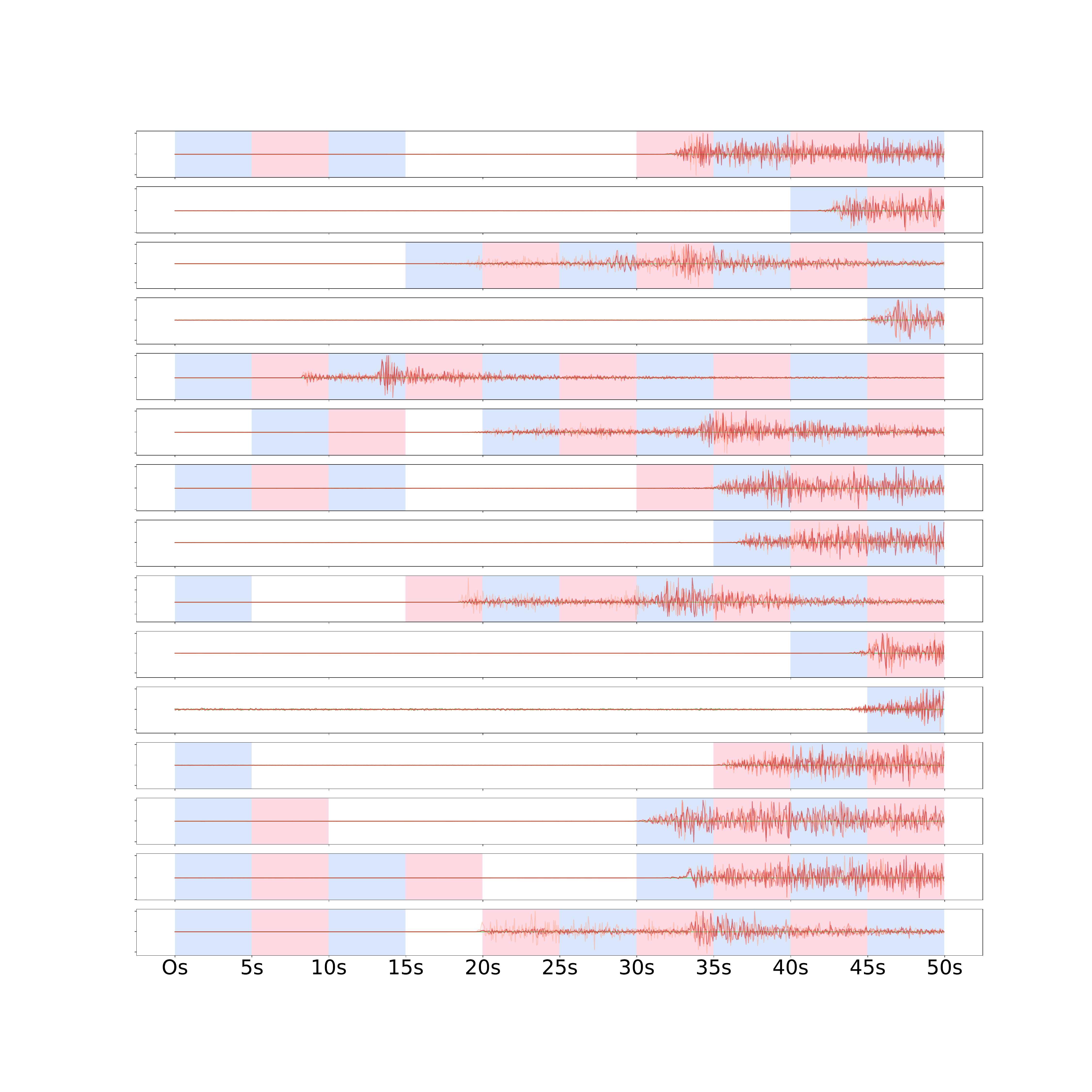}
\end{minipage}\\
\vspace{2em}
\begin{minipage}{0.25\textwidth} 
\centering
\textbf{\footnotesize LSST}\\
\vspace{0.5em}
\includegraphics[width=\linewidth,trim=160pt 80pt 140pt 70pt,clip]{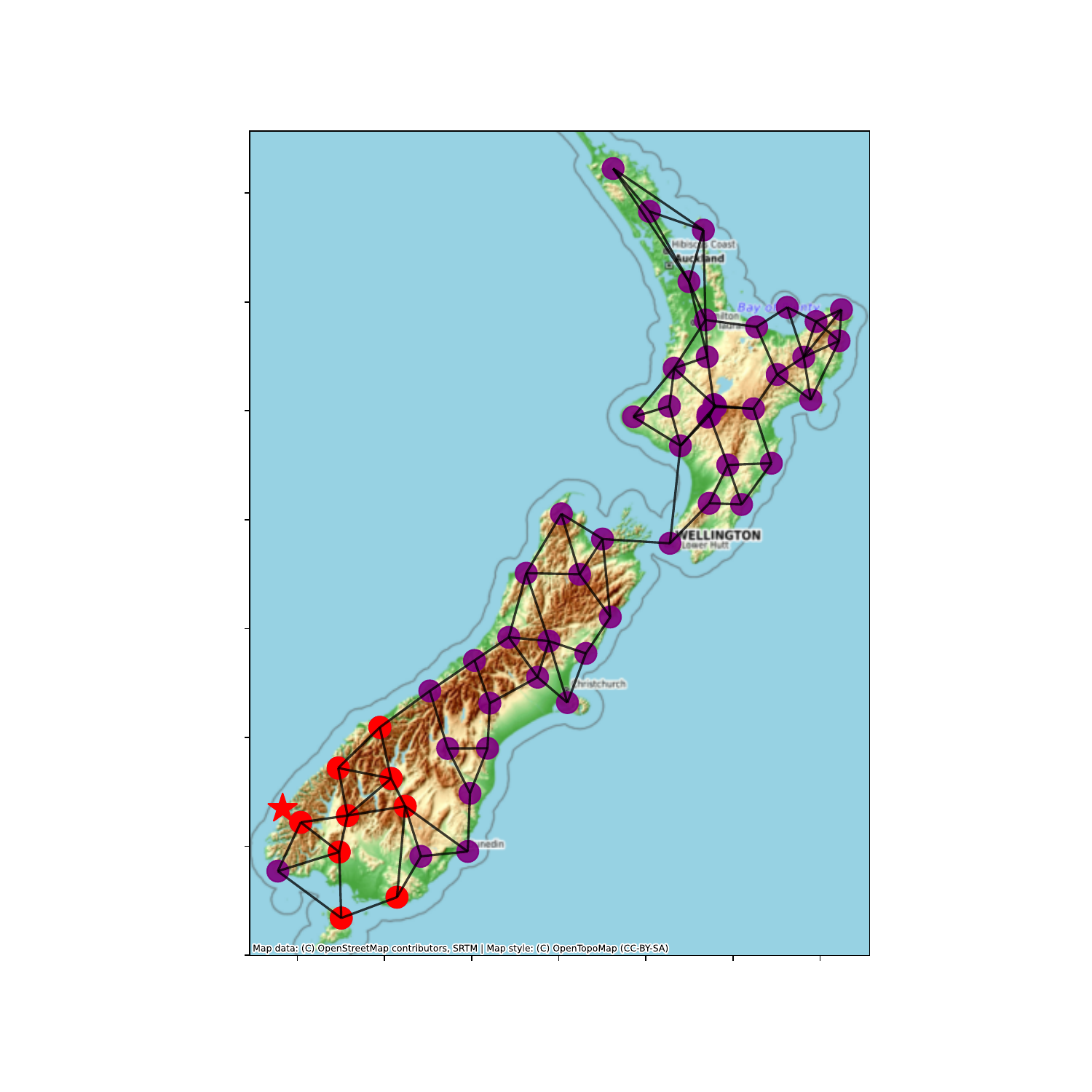}\\
\includegraphics[width=\linewidth,trim=0pt 550pt 285pt 0pt,clip]{figures/seisms_legend.pdf}
\end{minipage}
\hspace{1em}
\begin{minipage}{0.50\textwidth}
  \includegraphics[width=\linewidth,trim=250pt 110pt 210pt 150pt,clip]{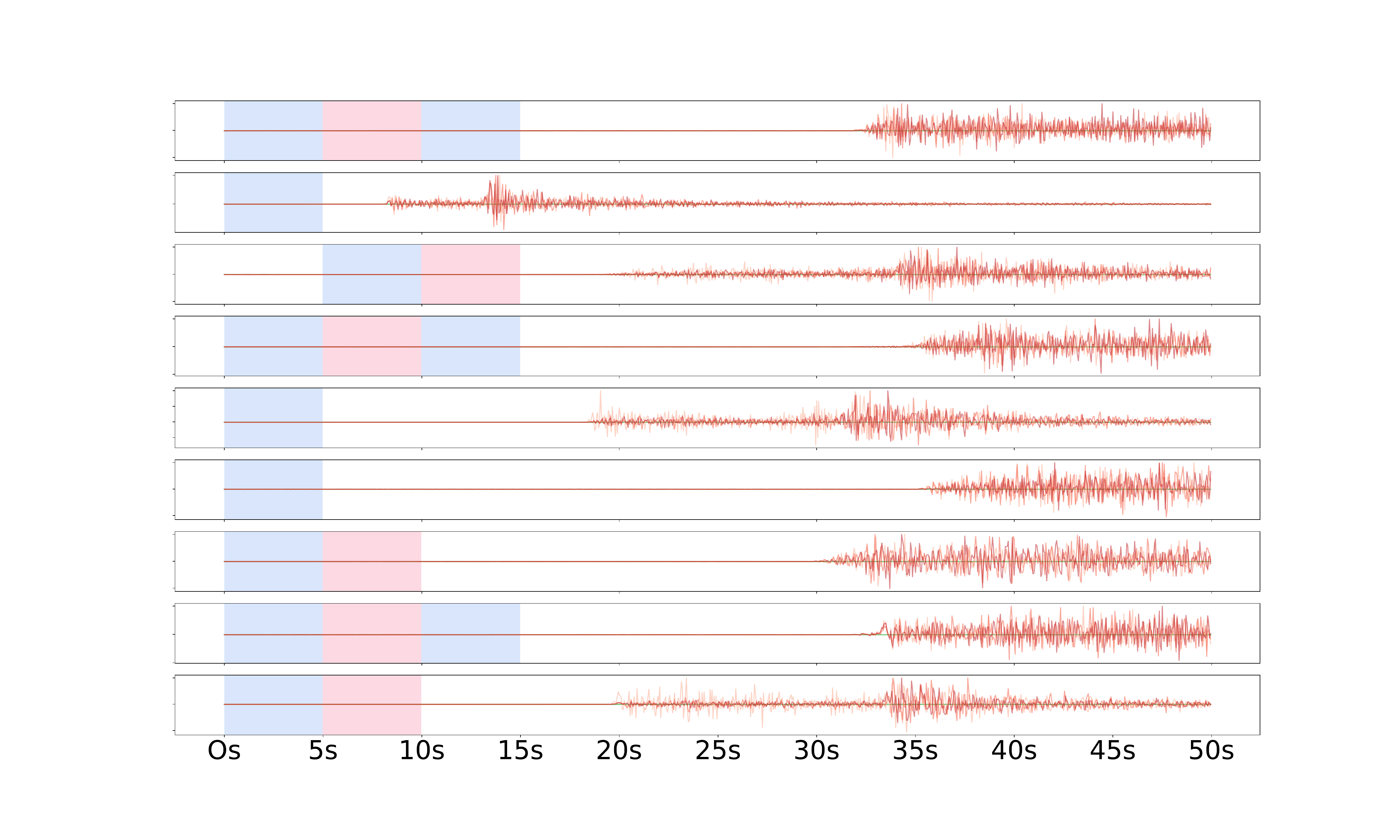}
\end{minipage}
\end{figure}


\begin{figure} 
\caption{\textbf{Seism\,A 
in New Zealand (3 of 4).}}\label{fig:NEW_ZEALAND_2021_3}
\vspace{1em}
\centering
\vspace{-3mm}
\begin{minipage}{0.25\textwidth} 
\centering
\textbf{\footnotesize MMD-MEDIAN}\\
\vspace{0.5em}
\includegraphics[width=\linewidth,trim=160pt 80pt 140pt 70pt,clip]{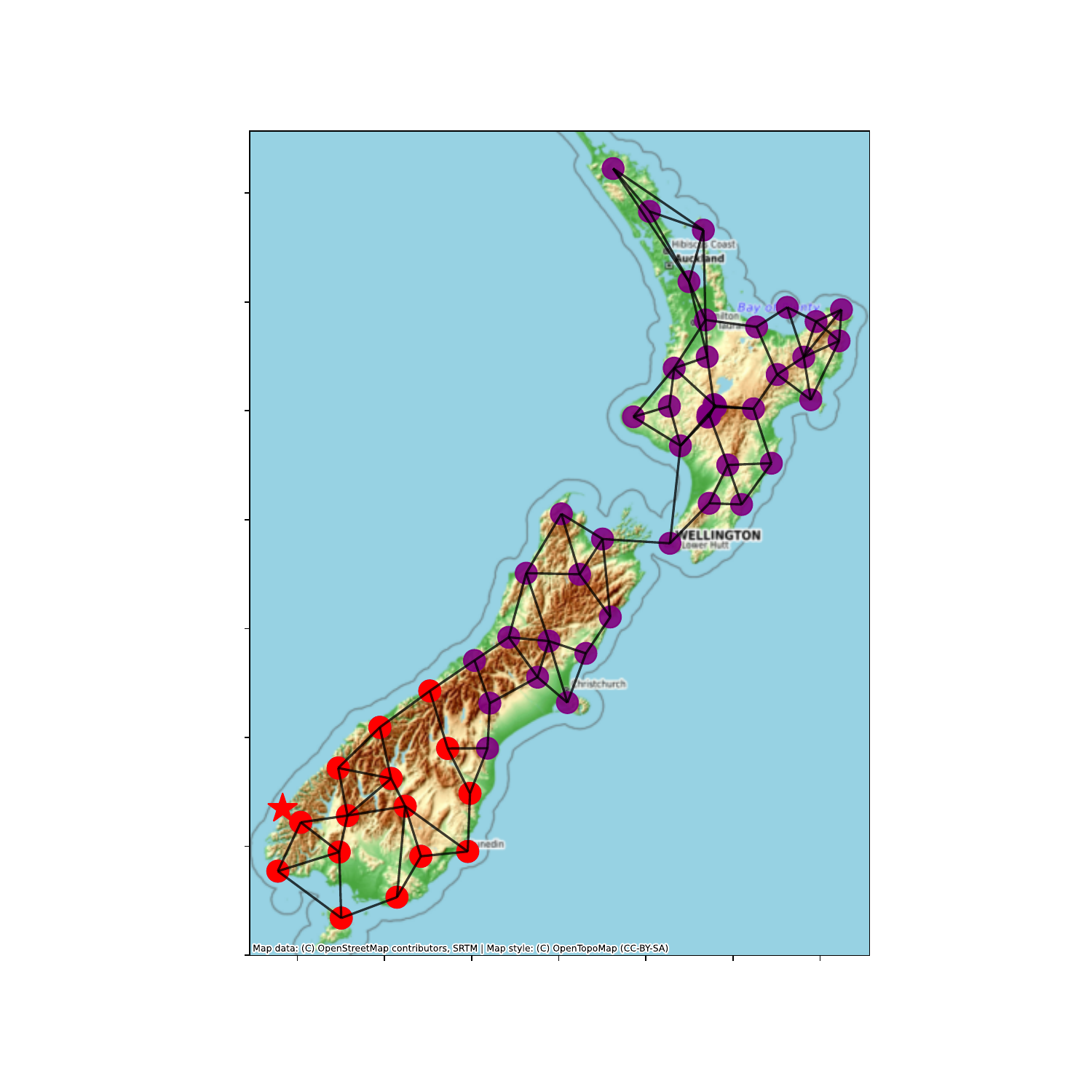}\\
\includegraphics[width=\linewidth,trim=0pt 550pt 285pt 0pt,clip]{figures/seisms_legend.pdf}
\end{minipage}
\hspace{1em}
\begin{minipage}{0.50\textwidth}
 \includegraphics[width=\linewidth,trim=250pt 150pt 210pt 230pt,clip]{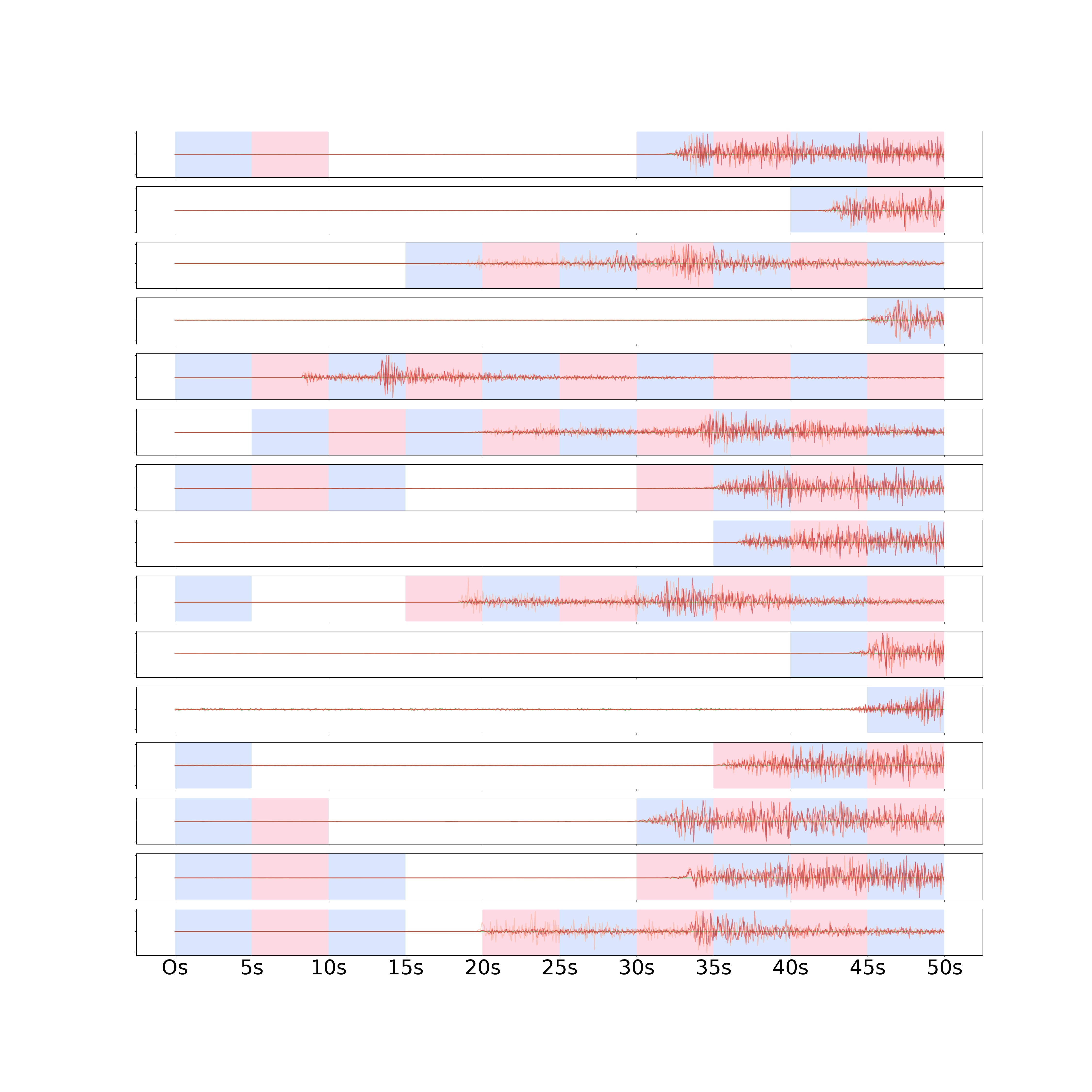}
\end{minipage}\\
\vspace{2em}
\begin{minipage}{0.25\textwidth} 
\centering
\textbf{\footnotesize MMD-MAX}\\
\vspace{0.5em}
\includegraphics[width=\linewidth,trim=160pt 80pt 140pt 70pt,clip]{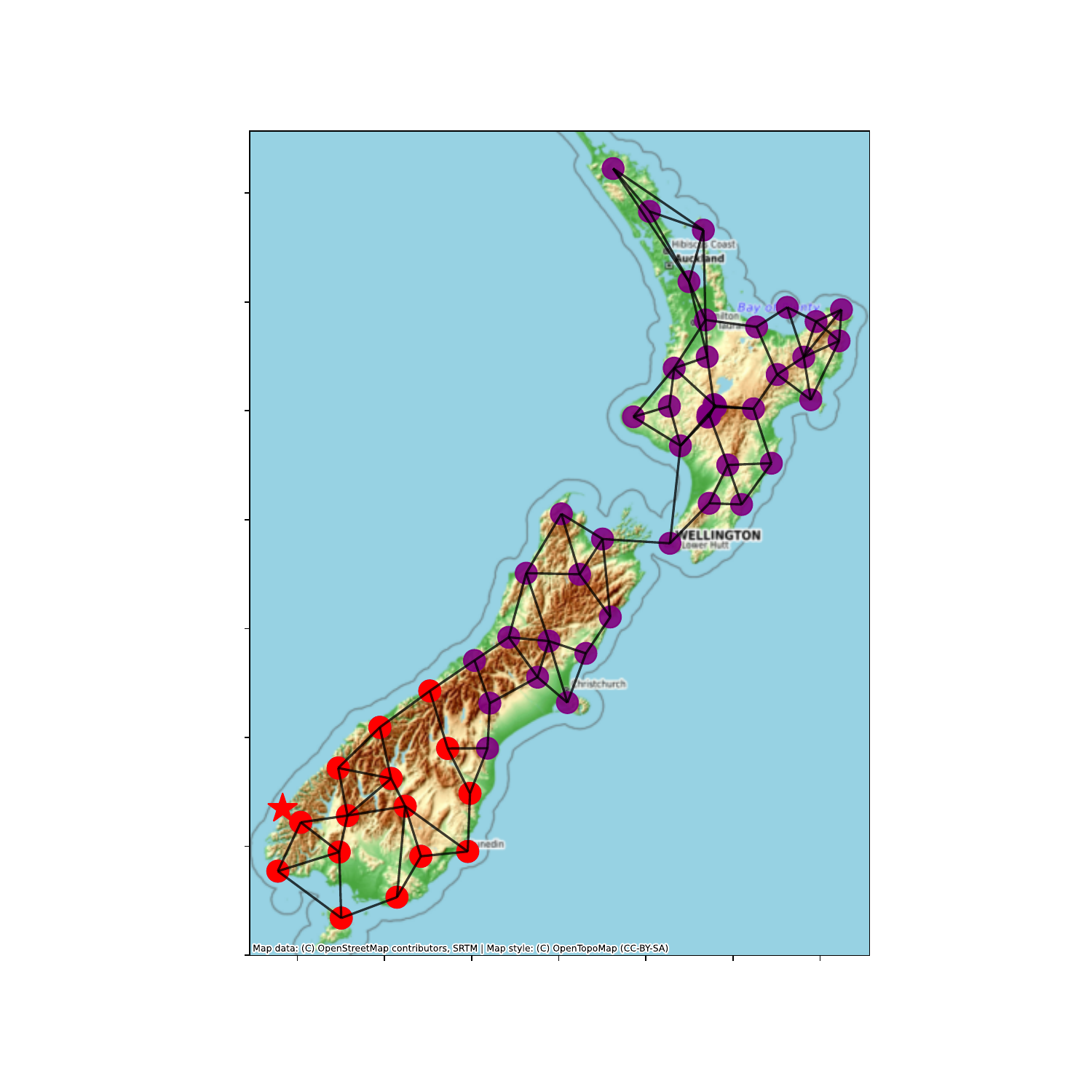}\\
\includegraphics[width=\linewidth,trim=0pt 550pt 285pt 0pt,clip]{figures/seisms_legend.pdf}
\end{minipage}
\hspace{1em}
\begin{minipage}{0.50\textwidth}  \includegraphics[width=\linewidth,trim=250pt 150pt 210pt 230pt,clip]{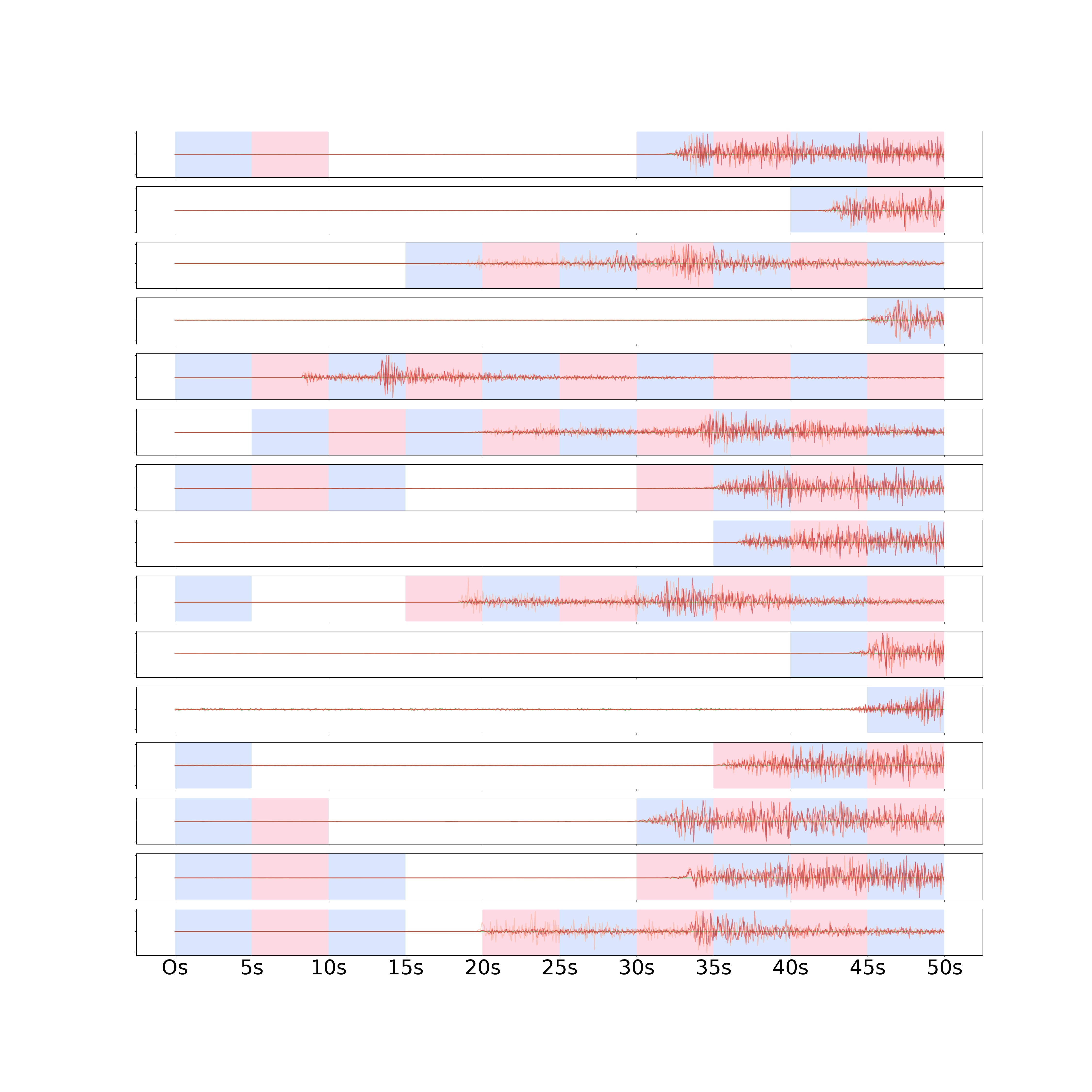}
\end{minipage}
\end{figure}


\begin{figure} 
\caption{\textbf{Seism\,A 
in New Zealand (4 of 4).}}\label{fig:NEW_ZEALAND_2021_4}
\vspace{1em}
\centering
\vspace{-3mm}
\begin{minipage}{0.25\textwidth} 
\centering
\textbf{\footnotesize KLIEP}\\
\vspace{0.5em}
\includegraphics[width=\linewidth,trim=160pt 80pt 140pt 70pt,clip]{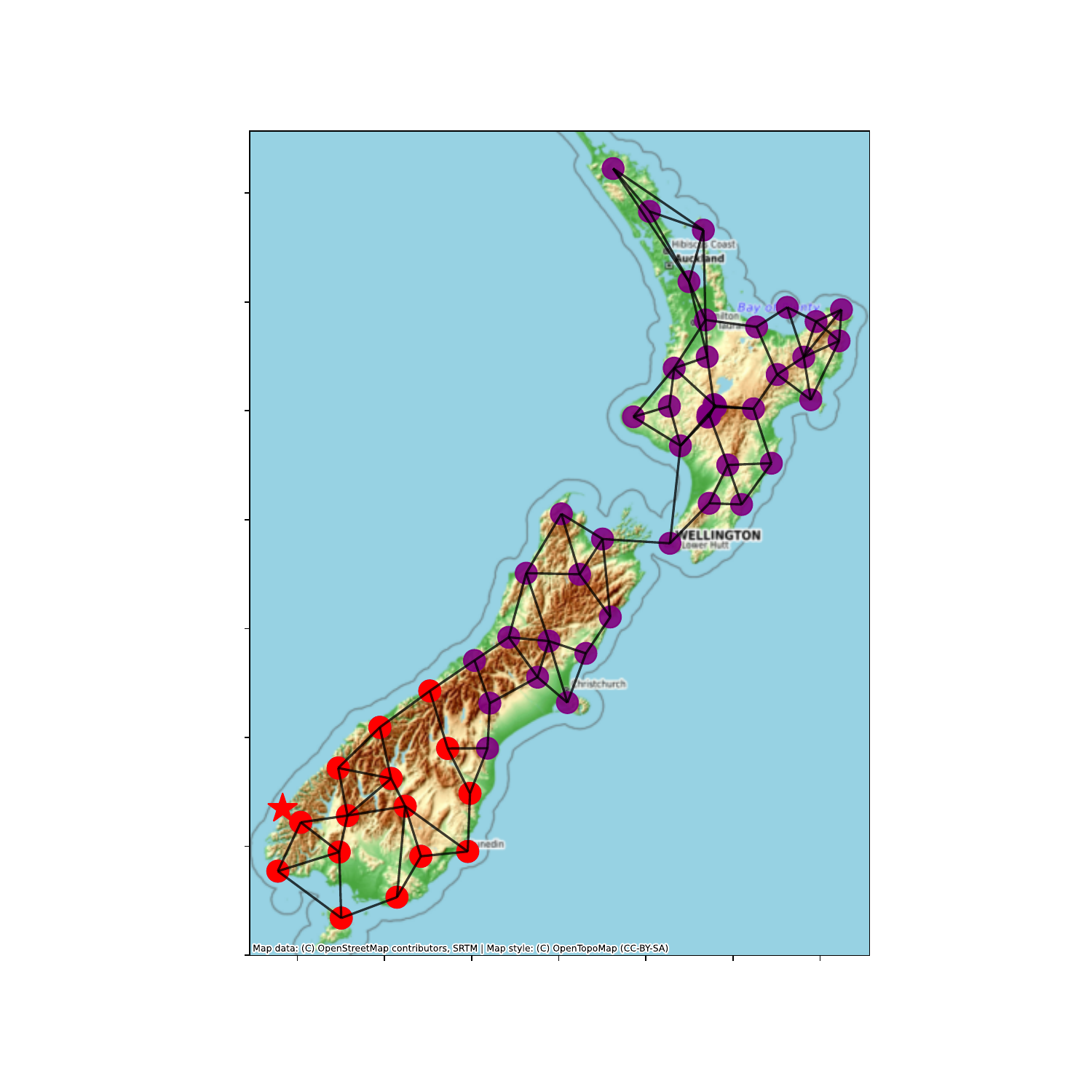}\\
\includegraphics[width=\linewidth,trim=0pt 550pt 285pt 0pt,clip]{figures/seisms_legend.pdf}
\end{minipage}
\hspace{1em}
\begin{minipage}{0.50\textwidth}
 \includegraphics[width=\linewidth,trim=250pt 150pt 210pt 230pt,clip]{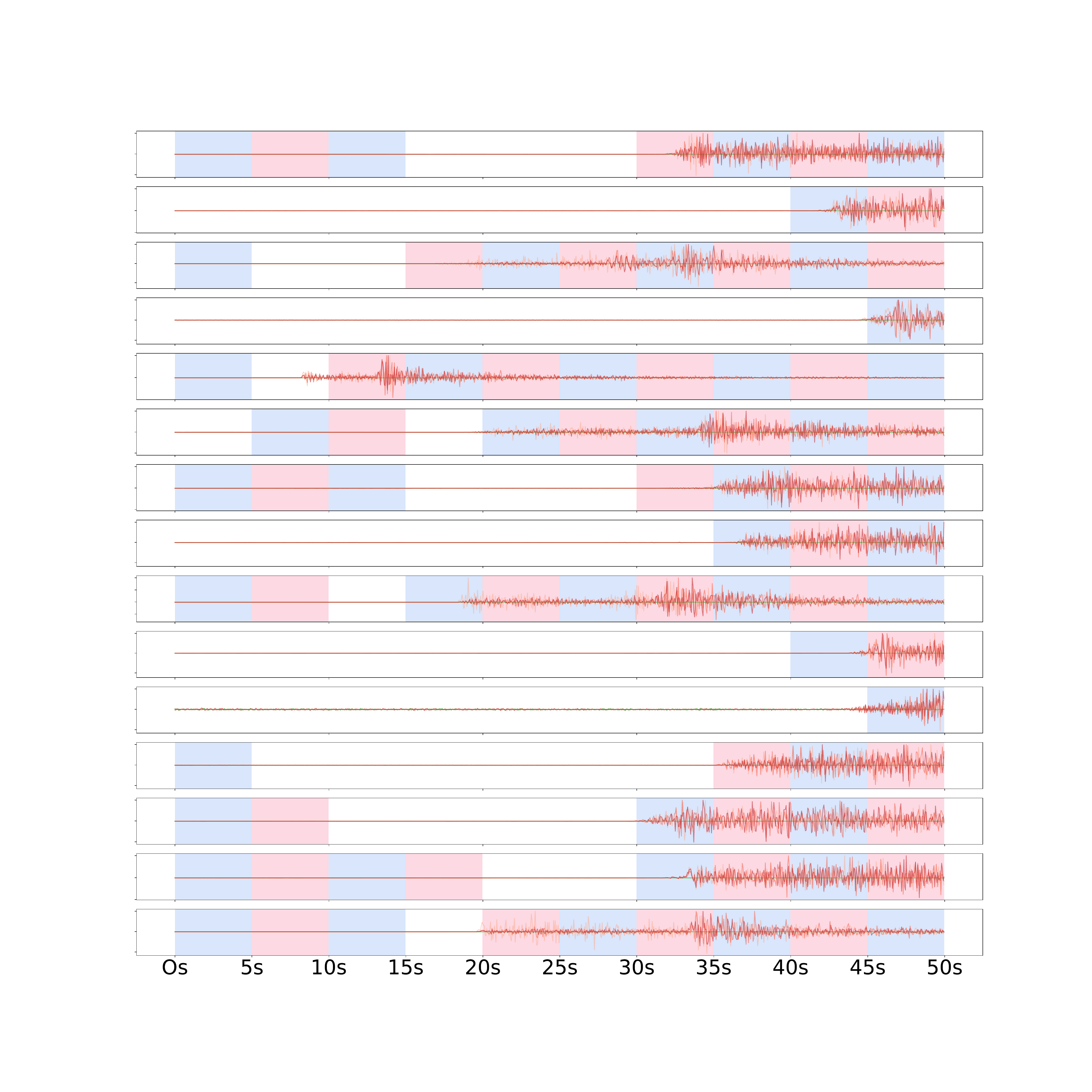}
\end{minipage}
\end{figure}



\begin{figure*} 
\caption{\textbf{Seism\,B 
in New Zealand (1 of 5).}}\label{fig:NEW_ZEALAND_2023_1}
\vspace{1em}
\centering
\vspace{-3mm}
\begin{minipage}{0.25\textwidth} 
\centering
\textbf{\footnotesize \CTST $\alpha=0.1$}\\
\vspace{0.5em}
\includegraphics[width=\linewidth,trim=160pt 80pt 140pt 70pt,clip]{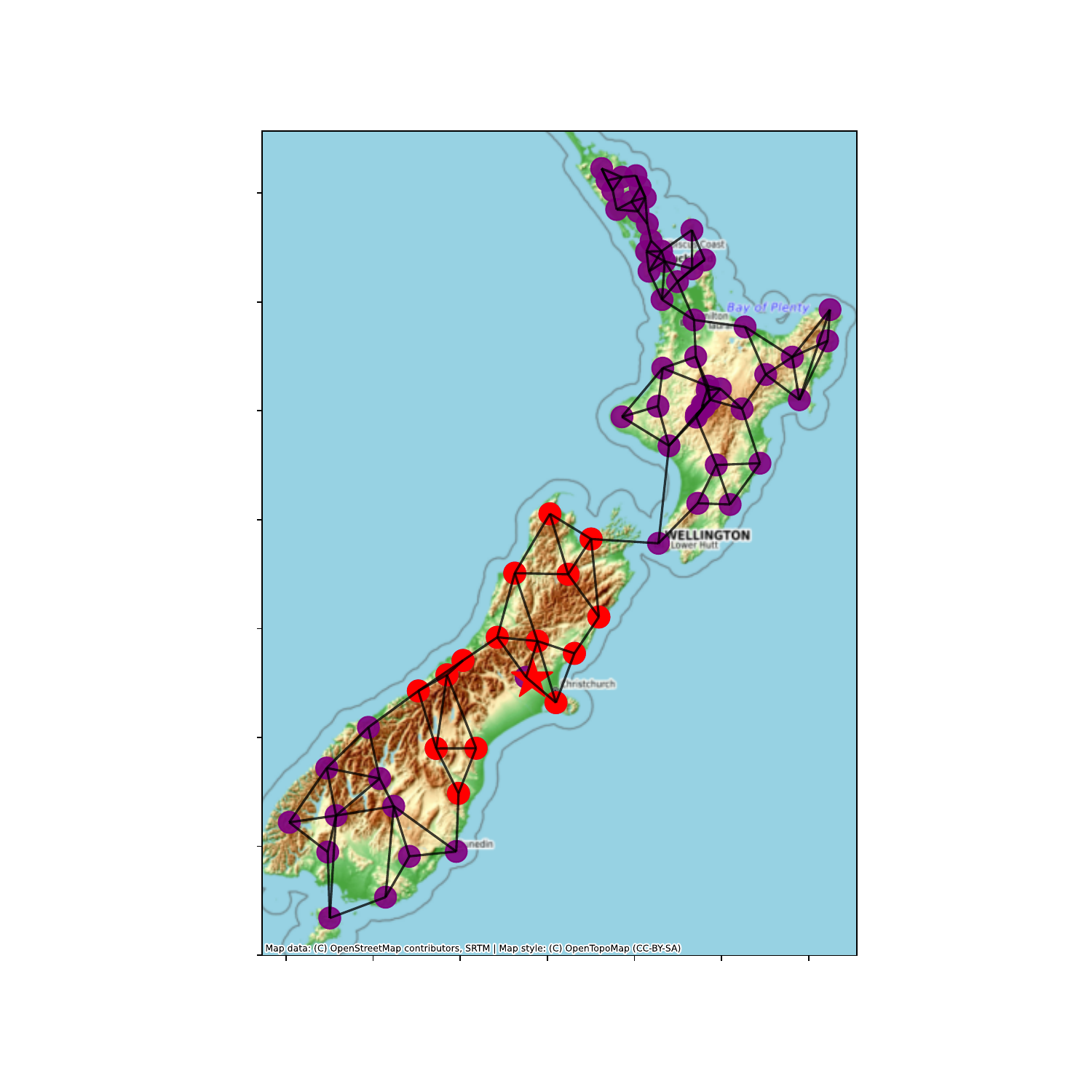}\\
\includegraphics[width=\linewidth,trim=0pt 550pt 285pt 0pt,clip]{figures/seisms_legend.pdf}
\end{minipage}
\hspace{1em}
\begin{minipage}{0.50\textwidth}
  \includegraphics[width=\linewidth,trim=250pt 150pt 210pt 230pt,clip]{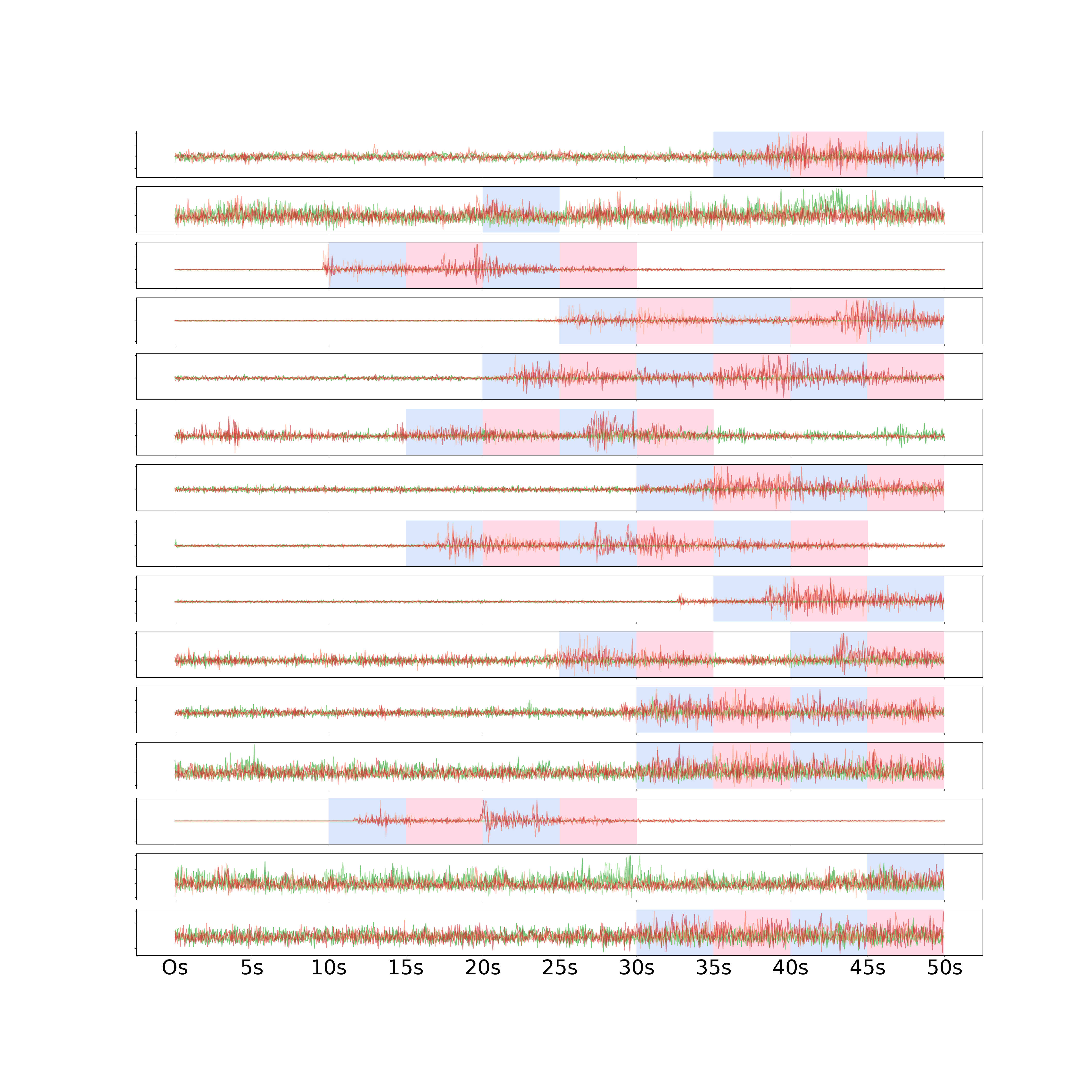}
\end{minipage}\\
\vspace{2em}
\begin{minipage}{0.25\textwidth} 
\centering
\textbf{\footnotesize POOL $\alpha=0.1$}\\
\vspace{0.5em}
\includegraphics[width=\linewidth,trim=160pt 80pt 140pt 70pt,clip]{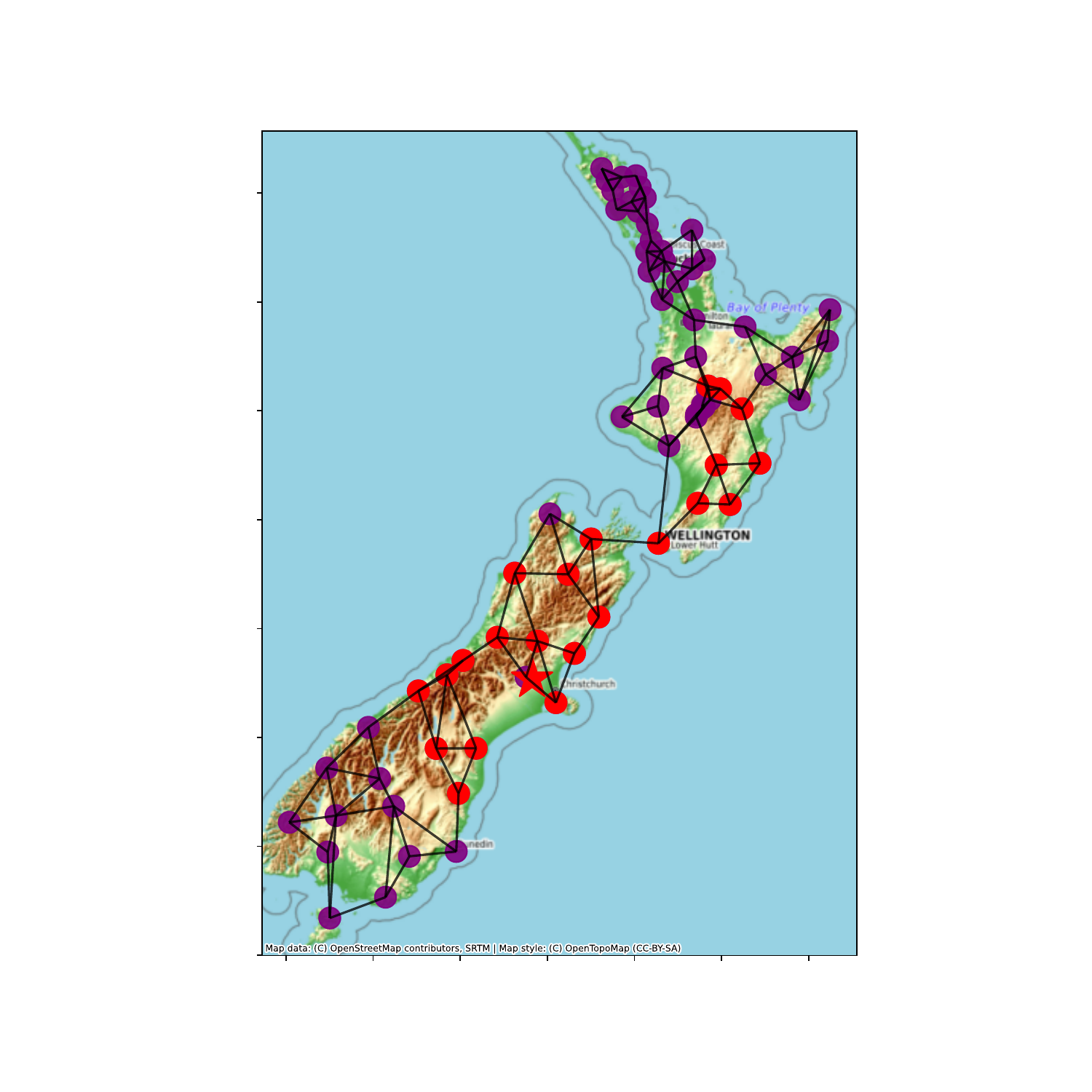}\\
\includegraphics[width=\linewidth,trim=0pt 550pt 285pt 0pt,clip]{figures/seisms_legend.pdf}
\end{minipage}
\hspace{1em}
\begin{minipage}{0.50\textwidth}
  \includegraphics[width=\linewidth,trim=250pt 150pt 210pt 340pt,clip]{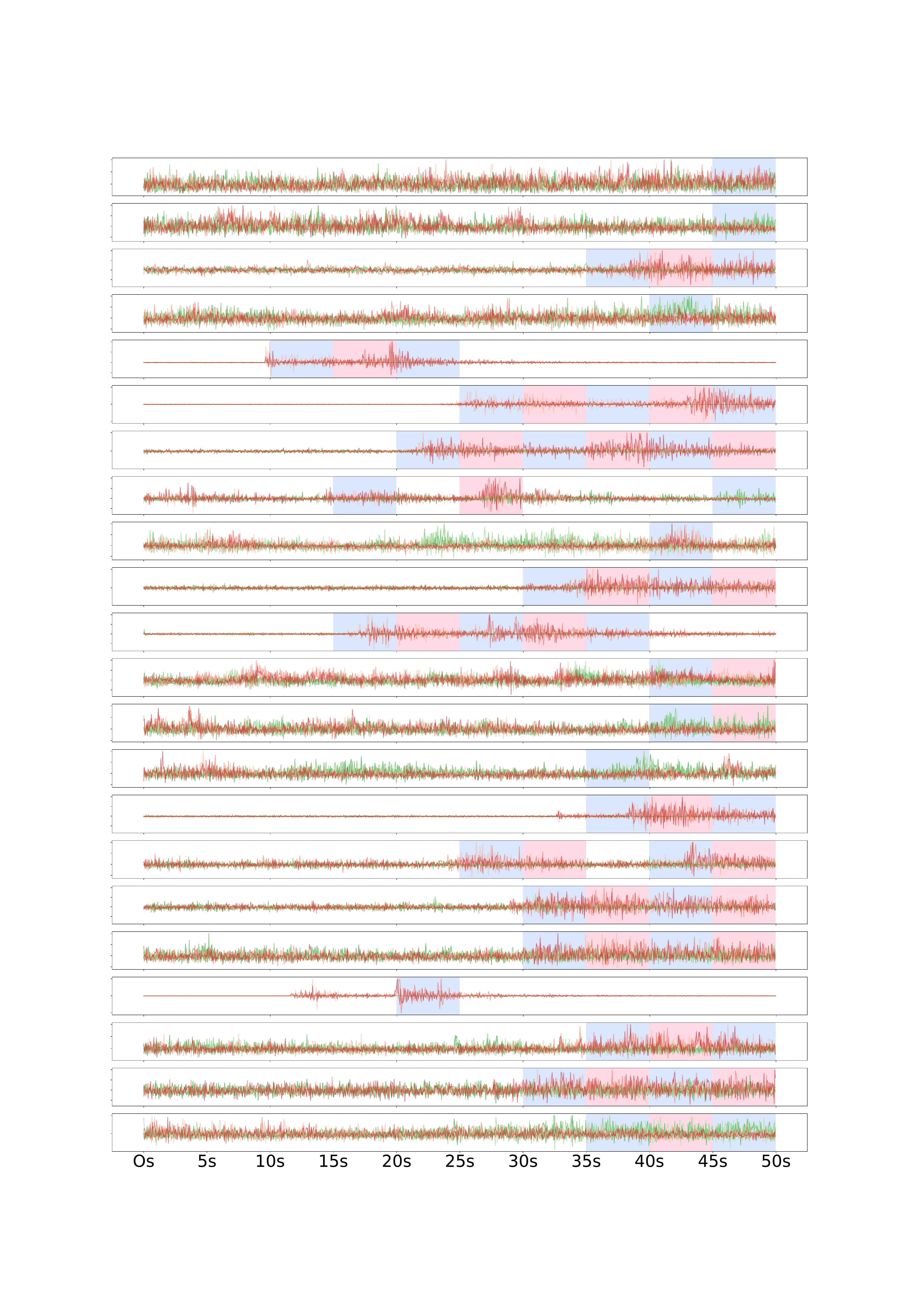}
\end{minipage}
\end{figure*}

\begin{figure*} 
\caption{\textbf{Seism\,B 
in New Zealand (2 of 5).}}\label{fig:NEW_ZEALAND_2023_2}
\vspace{1em}
\centering
\vspace{-3mm}
\begin{minipage}{0.25\textwidth} 
\centering
\textbf{\footnotesize RULSIF $\alpha=0.1$}\\
\vspace{0.5em}
\includegraphics[width=\linewidth,trim=160pt 80pt 140pt 70pt,clip]{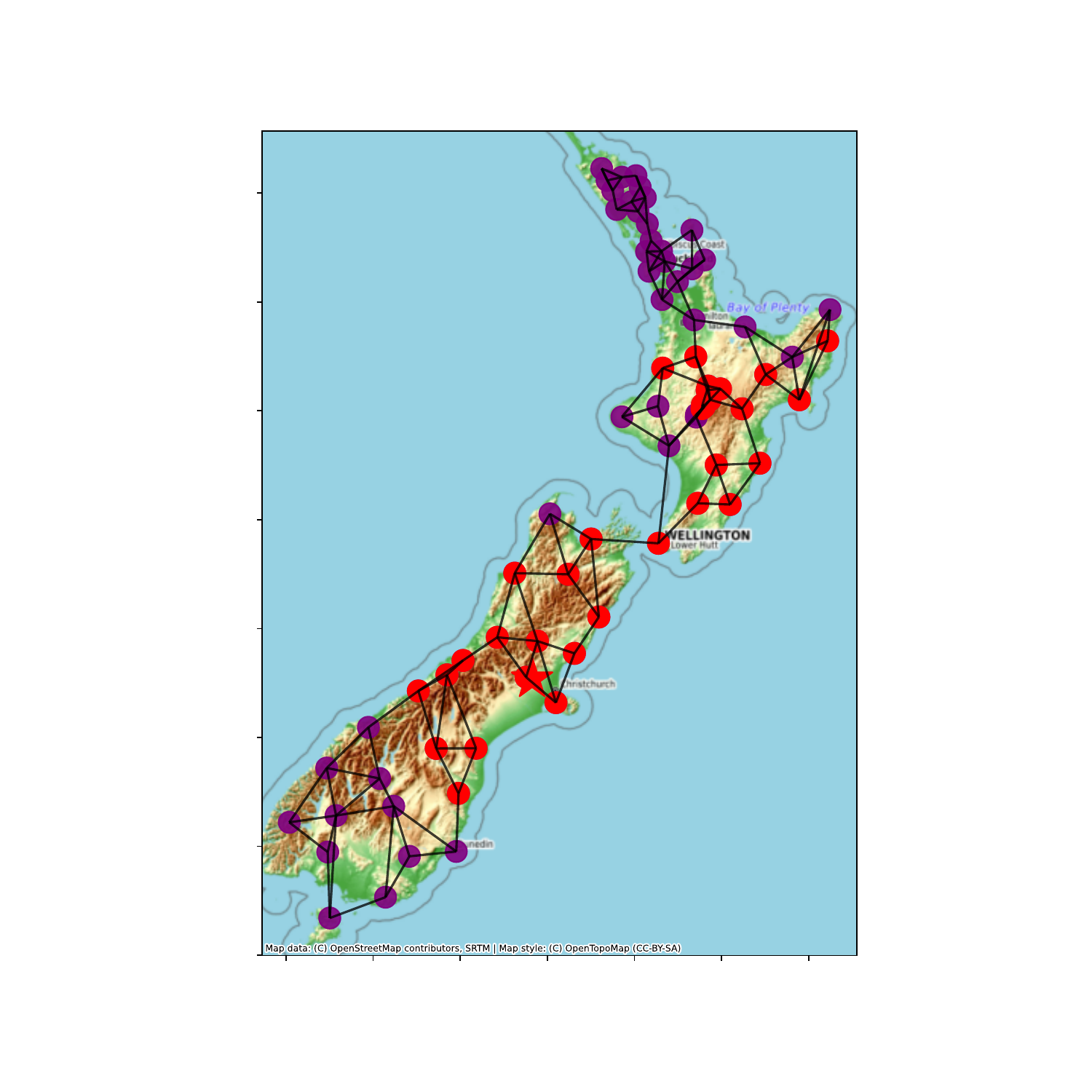}\\
\includegraphics[width=\linewidth,trim=0pt 550pt 285pt 0pt,clip]{figures/seisms_legend.pdf}
\end{minipage}
\hspace{1em}
\begin{minipage}{0.5\textwidth} \includegraphics[width=\linewidth,trim=250pt 550pt 210pt 540pt,clip]{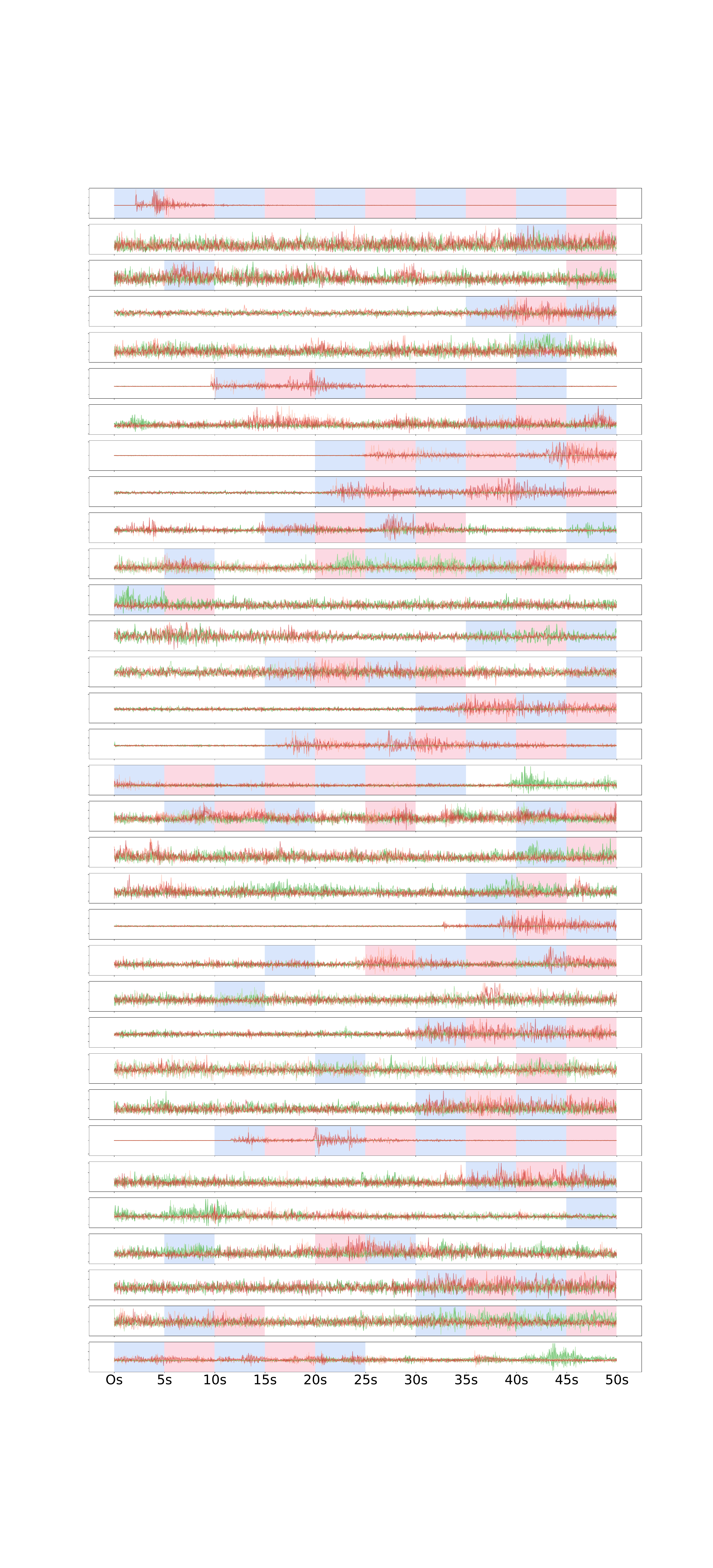}
\end{minipage}
\end{figure*}

\begin{figure*} 
\caption{\textbf{Seism\,B 
in New Zealand (3 of 5).}}\label{fig:NEW_ZEALAND_2023_3}
\vspace{1em}
\centering
\vspace{-3mm}
\begin{minipage}{0.25\textwidth} 
\centering
\textbf{\footnotesize LSST}\\
\vspace{0.5em}
\includegraphics[width=\linewidth,trim=160pt 80pt 140pt 70pt,clip]{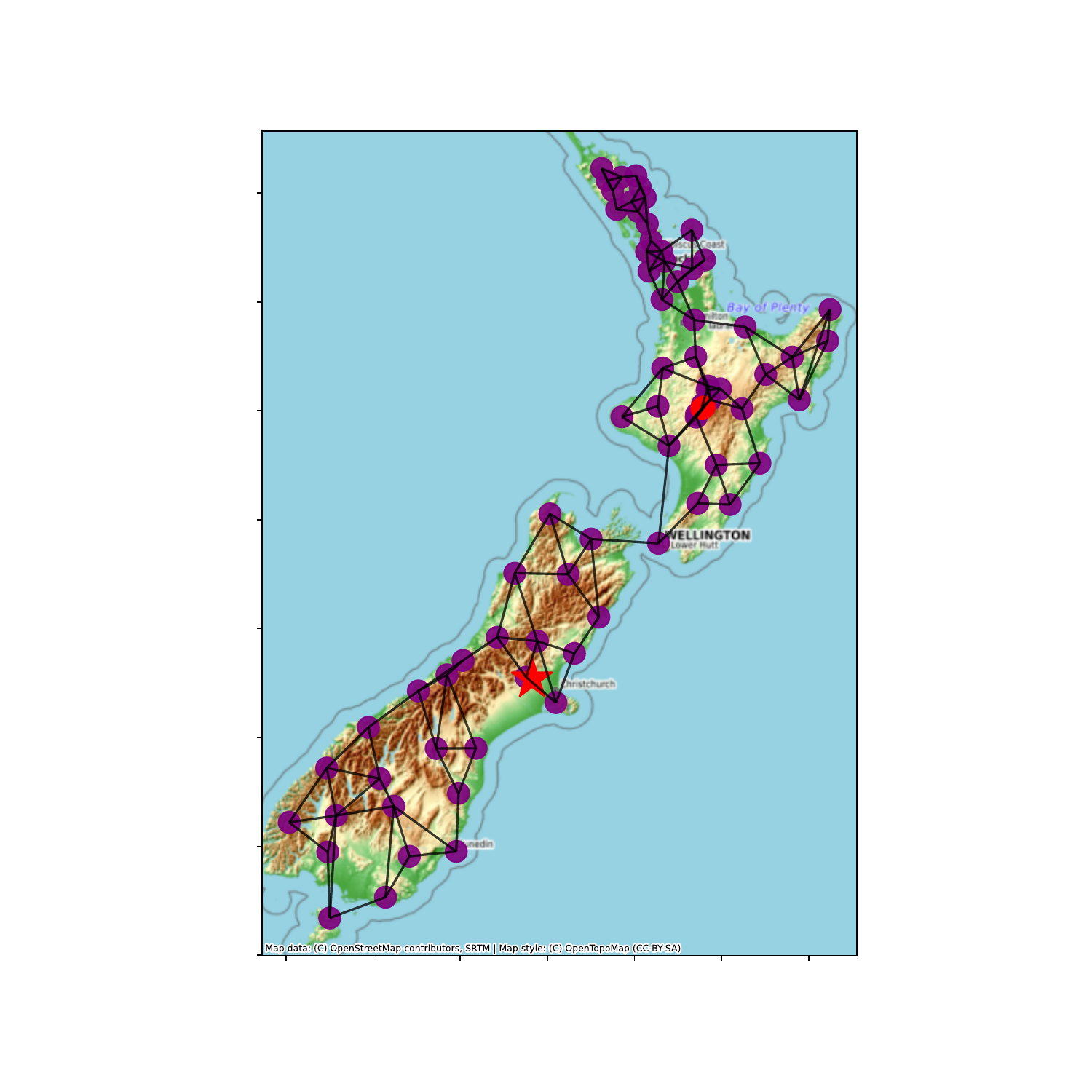}\\
\includegraphics[width=\linewidth,trim=0pt 550pt 285pt 0pt,clip]{figures/seisms_legend.pdf}
\end{minipage}
\hspace{1em}
\begin{minipage}{0.5\textwidth}
  \includegraphics[width=\linewidth,trim=250pt 260pt 210pt 800pt,clip]{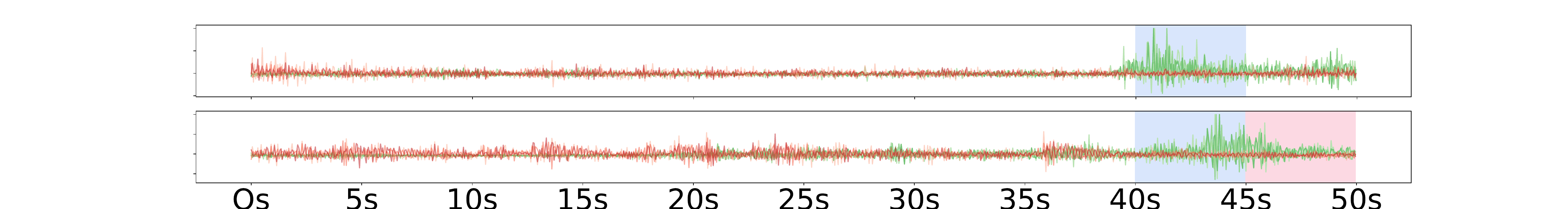}
\end{minipage}\\
\begin{minipage}{0.25\textwidth} 
\centering
\textbf{\footnotesize MMD-MEDIAN}\\
\vspace{0.5em}
\includegraphics[width=\linewidth,trim=160pt 80pt 140pt 70pt,clip]{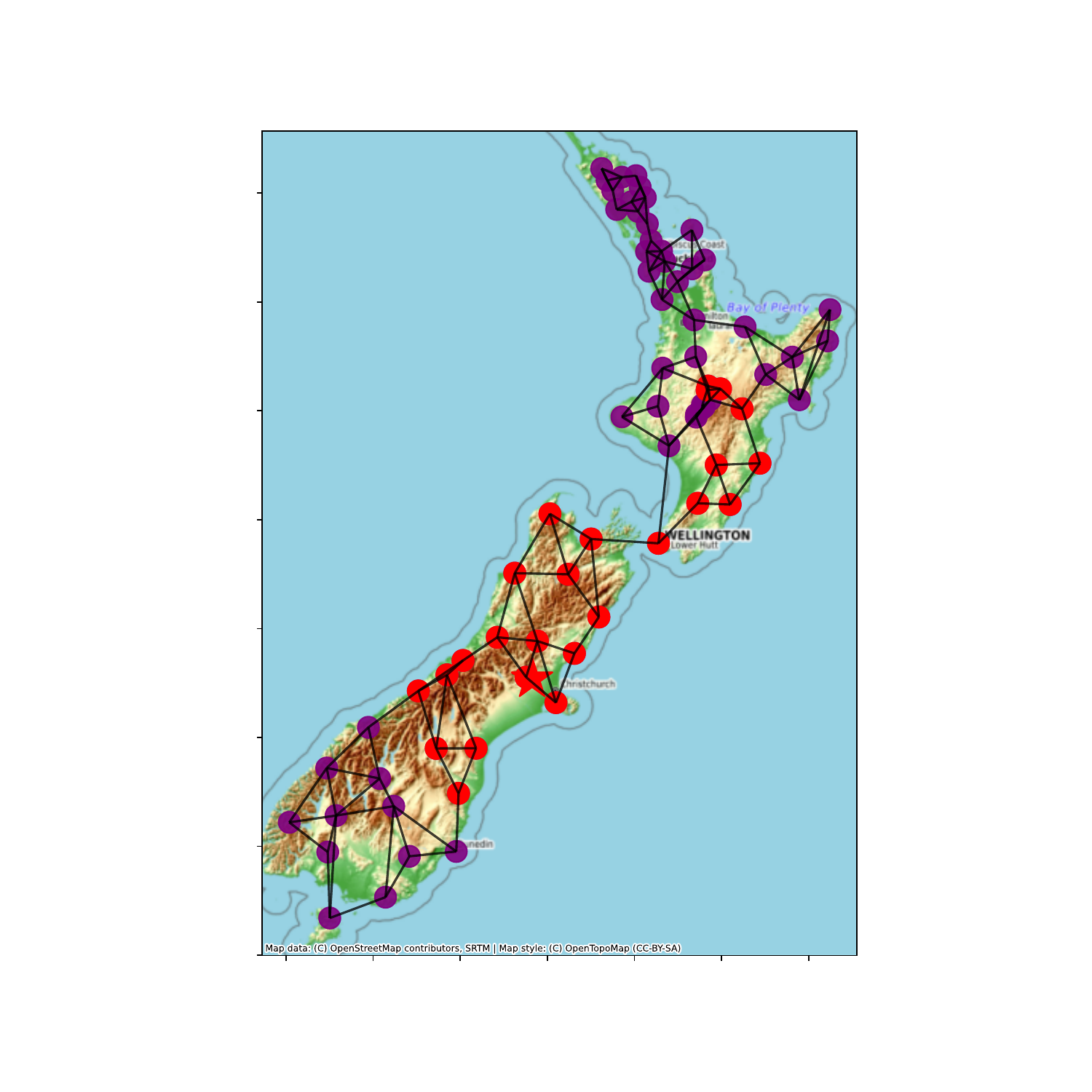}\\
\includegraphics[width=\linewidth,trim=0pt 550pt 285pt 0pt,clip]{figures/seisms_legend.pdf}
\end{minipage}
\hspace{1em}
\begin{minipage}{0.50\textwidth}
  \includegraphics[width=\linewidth,trim=250pt 400pt 210pt 340pt,clip]{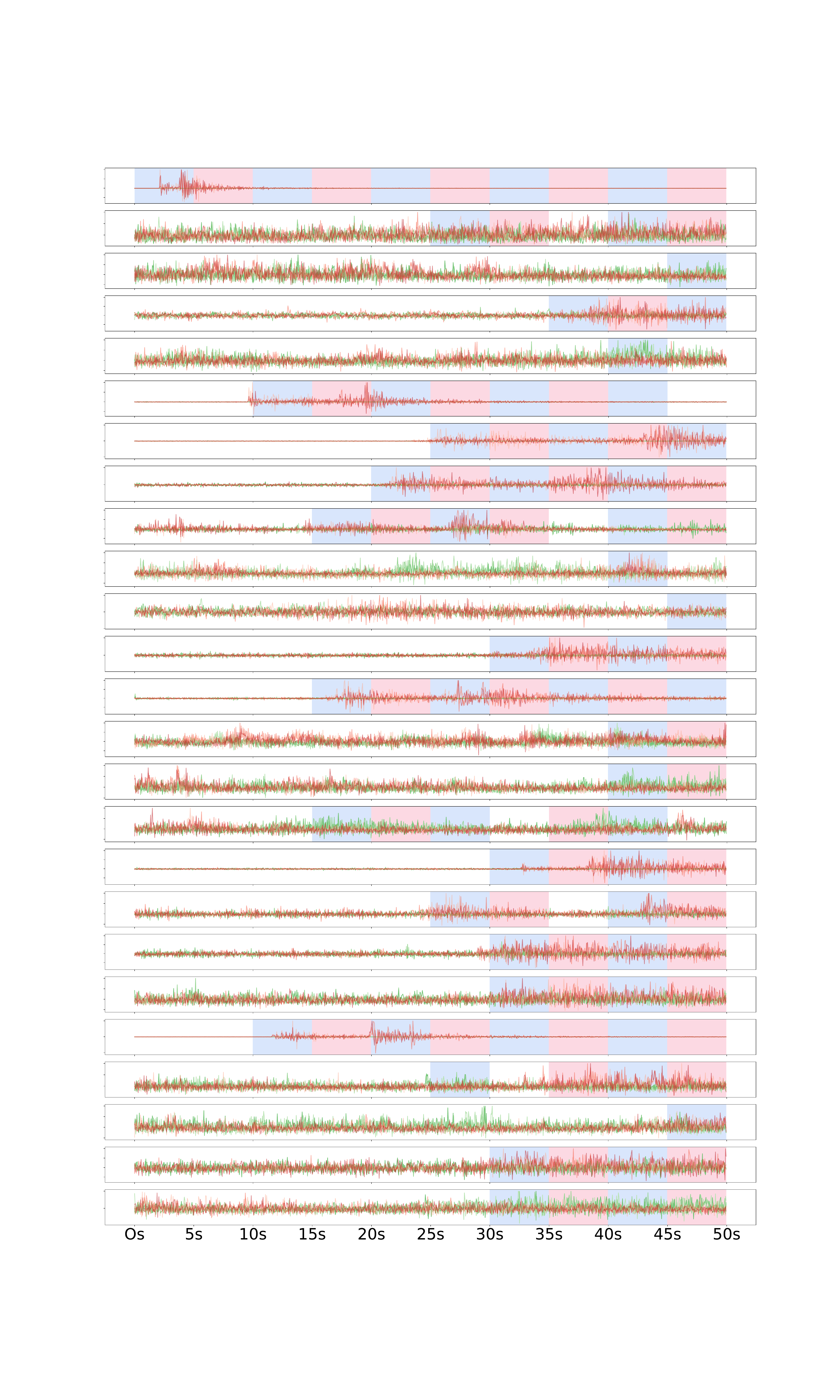}
\end{minipage}
\end{figure*}

\begin{figure*} 
\caption{\textbf{Seism\,B 
in New Zealand (4 of 5).}}\label{fig:NEW_ZEALAND_2023_4}
\vspace{1em}
\centering
\vspace{-3mm}
\begin{minipage}{0.25\textwidth} 
\centering
\textbf{\footnotesize MMD-MAX}\\
\vspace{0.5em}
\includegraphics[width=\linewidth,trim=160pt 80pt 140pt 70pt,clip]{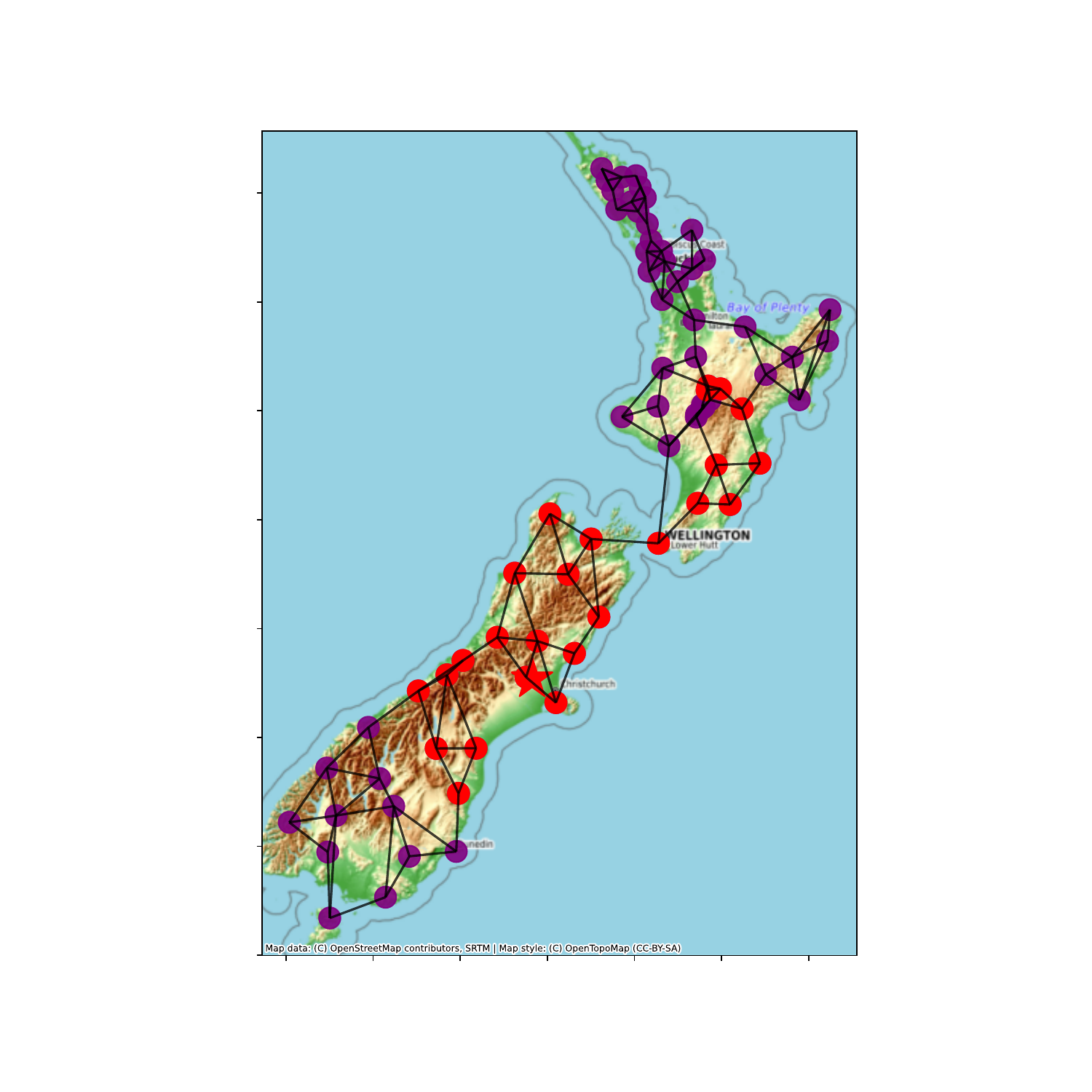}\\
\includegraphics[width=\linewidth,trim=0pt 550pt 285pt 0pt,clip]{figures/seisms_legend.pdf}
\end{minipage}
\hspace{1em}
\begin{minipage}{0.50\textwidth}
  \includegraphics[width=\linewidth,trim=250pt 150pt 210pt 400pt,clip]{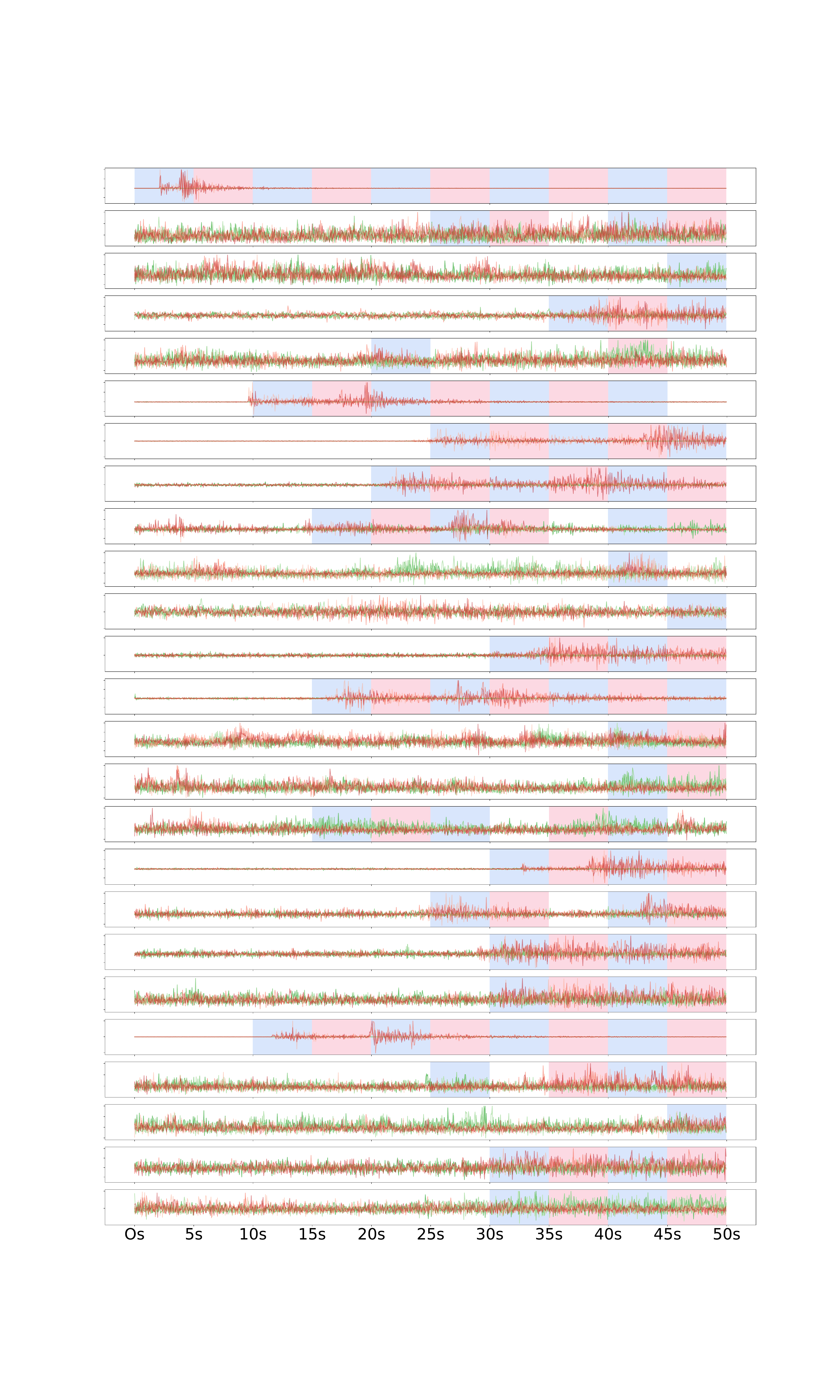}
\end{minipage}%
\end{figure*}

\begin{figure*} 
\caption{\textbf{Seism\,B 
in New Zealand (5 of 5).}}\label{fig:NEW_ZEALAND_2023_5}
\vspace{1em}
\centering
\vspace{-3mm}
\begin{minipage}{0.25\textwidth} 
\centering
\textbf{\footnotesize KLIEP}\\
\vspace{0.5em}
\includegraphics[width=\linewidth,trim=160pt 80pt 140pt 70pt,clip]{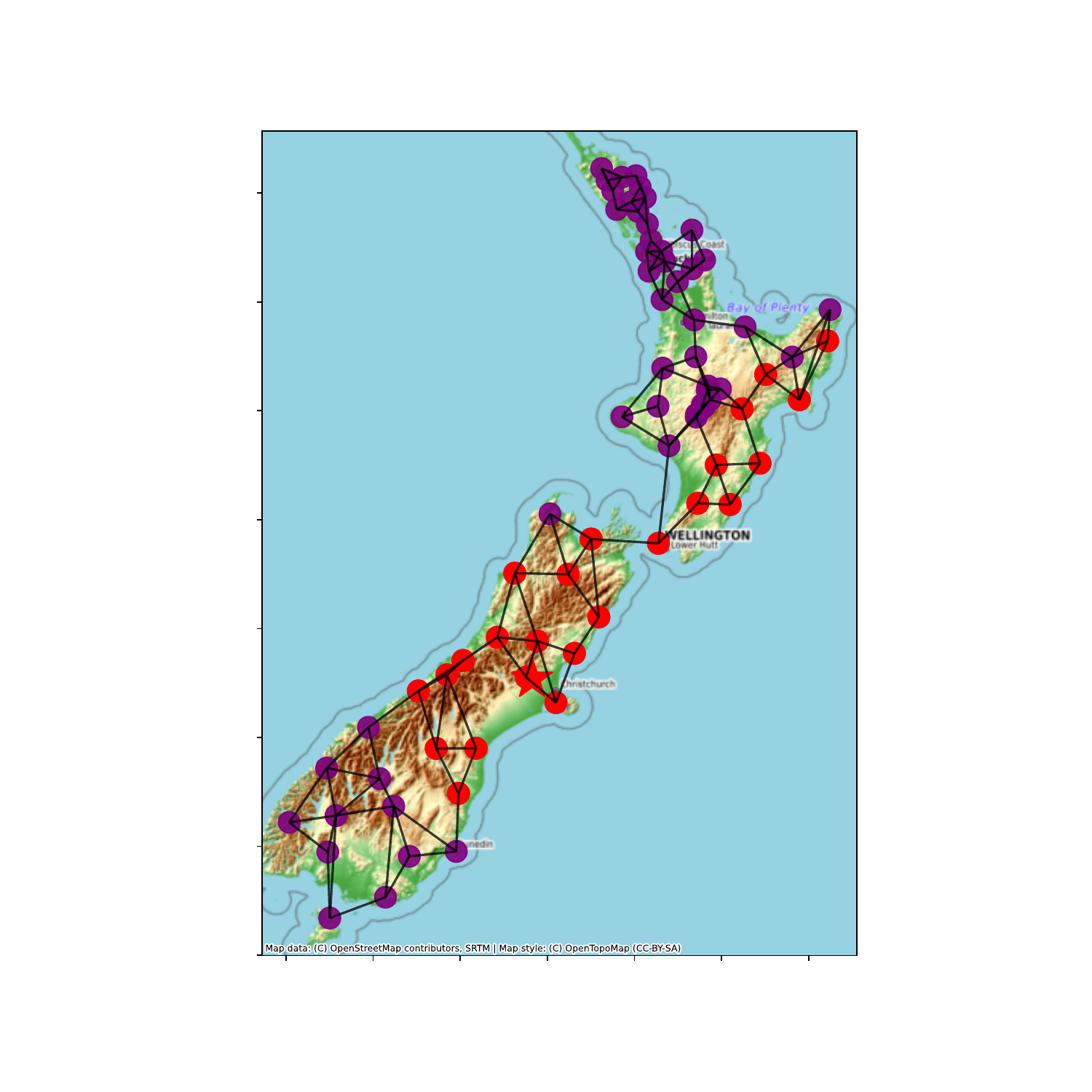}\\
\includegraphics[width=\linewidth,trim=0pt 550pt 285pt 0pt,clip]{figures/seisms_legend.pdf}
\end{minipage}
\hspace{1em}
\begin{minipage}{0.50\textwidth}
  \includegraphics[width=\linewidth,trim=250pt 150pt 210pt 400pt,clip]{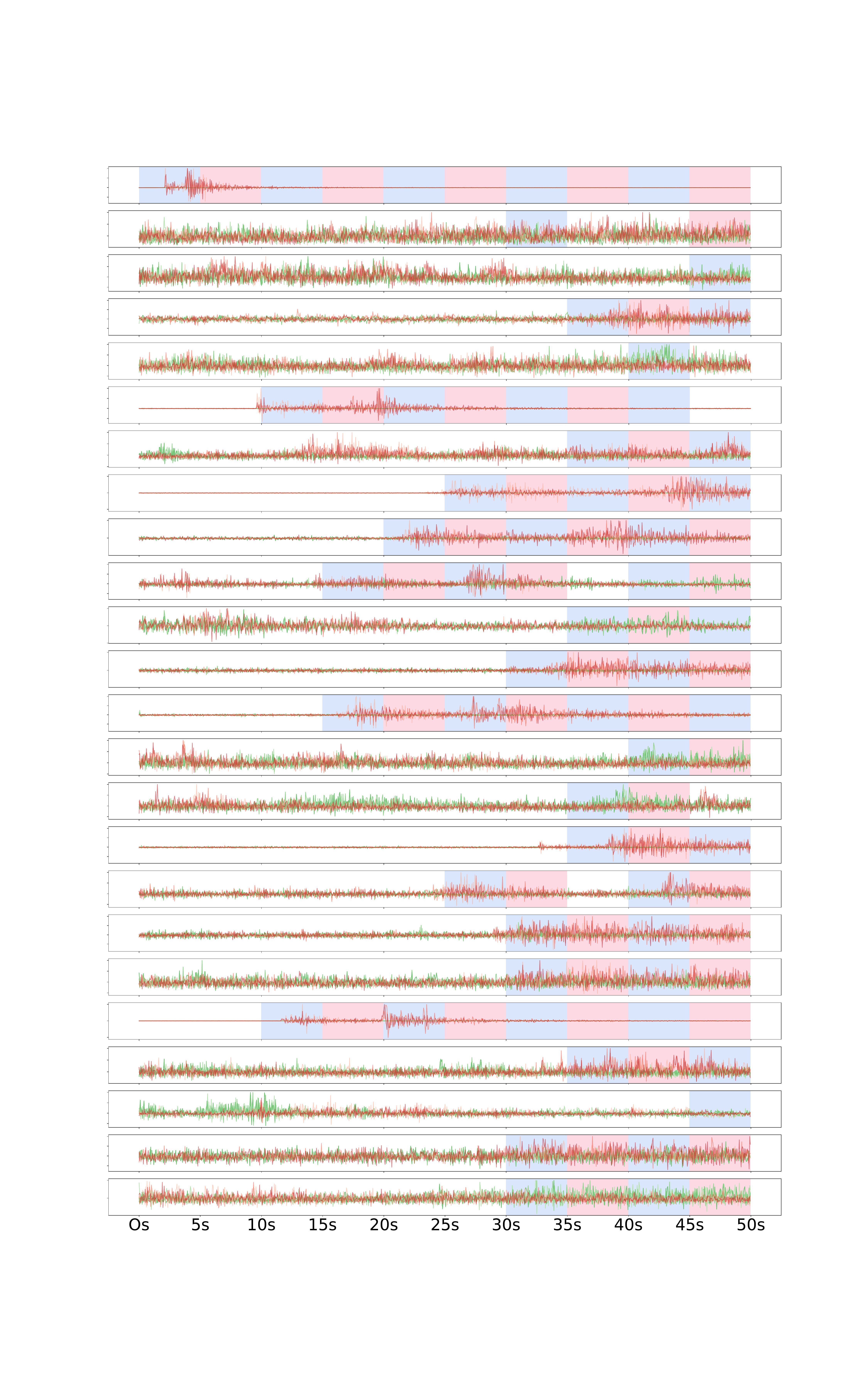}
\end{minipage}%
\end{figure*}

\end{document}